\documentclass[nohyperref]{article}
\usepackage[margin=1in]{geometry}
\usepackage{microtype}
\usepackage{mathpazo}
\usepackage{amsmath,amssymb,amsthm,amsfonts,bm}
\usepackage{mathtools}
\usepackage{thmtools}
\usepackage{thm-restate}
\usepackage{xspace}
\usepackage{graphicx}
\usepackage[dvipsnames]{xcolor}
\usepackage{booktabs}
\usepackage[linktocpage,colorlinks=true,linkcolor=blue,citecolor=blue,urlcolor=blue]{hyperref}
\usepackage{todonotes}
\usepackage[normalem]{ulem}
\usepackage{subcaption}
\usepackage{url}
\usepackage{comment}
\usepackage{algorithm}
\usepackage{algpseudocode}
\usepackage[toc,page,header]{appendix}
\usepackage{wrapfig}
\usepackage{multirow}
\usepackage{multicol}
\usepackage{float}
\usepackage{titletoc}
\usepackage{cleveref}
\usepackage[T1]{fontenc}
\usepackage[backend=biber,style=alphabetic,natbib=true,maxbibnames=99,minalphanames=3]{biblatex}
\usepackage[section]{placeins}
\usepackage{xspace}
\usepackage{fancyvrb}
\usepackage{fvextra}     %
\usepackage[most]{tcolorbox}
\usepackage{enumitem}
\usepackage{array}
\usepackage{tabularx}
\usepackage{color,soul}
\usepackage[utf8]{inputenc}
\usepackage{relsize}

\setlist[itemize]{leftmargin=*,labelindent=1em}
\setlist[enumerate]{leftmargin=*,labelindent=1em}

\newenvironment{prompt}
  {\VerbatimEnvironment
   \begin{tcolorbox}[
     colback   = white,      %
     colframe  = black,      %
     boxrule   = 0.5mm,      %
     sharp corners,          %
     breakable,              %
     enhanced                %
   ]
   \begin{Verbatim}[breaklines=true, breakanywhere=true]}
  {\end{Verbatim}
   \end{tcolorbox}}

\newcommand{\hlc}[2][yellow]{{%
\colorlet{foo}{#1}%
\sethlcolor{foo}\hl{#2}}%
}

\makeatletter
\renewcommand{\paragraph}{%
  \@startsection{paragraph}{4}%
  {\z@}{1.5ex \@plus 1ex \@minus .2ex}{-1em}%
  {\normalfont\normalsize\bfseries}%
}
\makeatother

\theoremstyle{plain}
\newtheorem{theorem}{Theorem}[section]

\theoremstyle{definition}
\newtheorem{definition}[theorem]{Definition}

\theoremstyle{remark}

\newcommand*{\lasso}{\textsc{Lasso}\xspace}
\newcommand*{\plm}{p_\text{LM}}

\newcommand*{\cc}{\textsc{ContextCite}\xspace}
\newcommand*{\ours}{AT2\xspace}

\newcommand*{\ablate}{\textsc{Ablate}\xspace}
\newcommand*{\attn}{\text{Attn}\xspace}
\newcommand*{\aattn}{\text{Attn}\xspace}
\newcommand{\myhl}[4]{%
  \definecolor{customhl}{RGB}{#1,#2,#3}%
  \sethlcolor{customhl}%
  \hl{#4}%
}

\usepackage[normalem]{ulem}
\usepackage{xcolor}

\newcommand{\skipthis}[1]{\ignorespaces}

\newcommand{\scrcmd}{\textsuperscript}
\newcommand{\scr}[1]{\scrcmd{#1}}

\title{Learning to Attribute with Attention}
\author{
  Benjamin Cohen-Wang,\
  Yung-Sung Chuang,\
  Aleksander M\k{a}dry \\
  MIT \\
  \texttt{\{bencw,yungsung,madry\}@mit.edu}
}
\date{}

\begin{document}
\makeatletter
\let\c@table\c@figure
\let\c@lstlisting\c@figure
\makeatother
\maketitle

\begin{abstract}
Given a sequence of tokens generated by a language model, we may want to identify the preceding tokens that \emph{influence} the model to generate this sequence.
Performing such \emph{token attribution} is expensive; a common approach is to ablate preceding tokens and directly measure their effects.
To reduce the cost of token attribution, we revisit attention weights as a heuristic for how a language model uses previous tokens.
Na\"ive approaches to attribute model behavior with attention (e.g., averaging attention weights across attention heads to estimate a token's influence) have been found to be unreliable.
To attain faithful attributions, we propose treating the attention weights of different attention heads as \emph{features}.
This way, we can \emph{learn} how to effectively leverage attention weights for attribution (using signal from ablations).
Our resulting method, Attribution with Attention (\ours), reliably performs on par with approaches that involve many ablations, while being significantly more efficient.
To showcase the utility of \ours, we use it to prune less important parts of a provided context in a question answering setting, improving answer quality.
We provide code for \ours at \url{https://github.com/MadryLab/AT2}.

\end{abstract}

\section{Introduction}
\label{sec:intro}
When a language model generates content, it is guided by various prompts, provided context, and other preceding information.
Pinpointing the specific tokens among these that \emph{influence} the language model to generate a particular sequence can be valuable for understanding and debugging model behavior.
Indeed, such attributions have been used to provide citations for generated statements~\citep{cohenwang2024contextcite,Qi2024ModelIA}, identify biases~\citep{vig2020causal}, and assess the faithfulness of model-provided explanations~\citep{deyoung2019eraser,lanham2023measuring,madsen2024self}.

When we say that a model is \emph{influenced} by a particular set of tokens, we mean that \emph{removing} these tokens would substantially affect its generation\footnote{This definition follows from prior work on attributing model behavior~\citep{lundberg2017unified,ilyas2022datamodels,shah2024decomposing,cohenwang2024contextcite} and measuring the faithfulness of interpretability methods~\citep{arras2017relevant,deyoung2019eraser,madsen2021evaluating}}.
With this perspective, methods for (accurately) identifying influential tokens are often expensive.
For example, the gradient of a model's output with respect to a preceding token approximates the token's influence but involves significant additional computation and memory to obtain~\citep{yin2022interpreting,sarti2023quantifying}.
Ablating prior tokens is a way to directly measure their influence~\citep{ribeiro2016should,lundberg2017unified,cohenwang2024contextcite} but requires an inference pass for each ablation;
these ablations must be repeated for every example that we would like to attribute.
Seeking to reduce the cost of attribution, we ask:

\begin{center}
    \emph{Can we learn to (efficiently) predict the effect of ablating prior tokens on a model's generation?}
\end{center}

To answer this question, we consider the attention mechanism~\citep{bahdanau2014neural}, a key component of transformer language models~\citep{vaswani2017attention}.
Examining attention weights is a widely used strategy for interpreting model behavior \citep{lee2017interactive,ding2017visualizing,abnar2020quantifying,vig2020causal}.
Intuitively, attention weights reflect how a language model utilizes information from preceding tokens when predicting the next one.
However, prior work has illustrated that attention weights are often unreliable as explanations of model behavior ~\citep{jain2019attention,serrano2019attention,vashishth2019attention}.
In particular, when attributing a model's generation to provided context, methods leveraging attention were shown to be accurate for attributing certain models but highly inaccurate for attributing others \citep{cohenwang2024contextcite}.

To make effective use of attention weights for attribution, we are guided by the observation that different attention heads have different roles and utilities~\citep{zheng2024attention}.
For example,~\citet{wu2024retrieval} identify ``retrieval heads'' responsible for directly copying information from a context.
\citet{chuang2024lookback} illustrate that certain heads are more useful than others for detecting hallucinations.
These findings suggest that we may need to account for these differences in roles and utilities when considering attention weights as a signal for attribution.

\subsection{Our contributions}

We present an attribution method, Attribution with Attention (\ours), that treats the attention weights from different heads as \emph{features} (\Cref{sec:method}).
Specifically, we assign a (learnable) coefficient to each attention head signifying the extent to which we should rely on it to estimate influences.
As a source of ground-truth for influence, we \emph{ablate} random subsets of tokens and measure the corresponding effect on the model's generation.
Learning these coefficients requires performing several ablations for each example in a training dataset;
however, to then attribute any \emph{new} generation, we can then just combine attention weights according to these coefficients.

We evaluate the effectiveness of \ours in two settings: attributing a model's generation to provided context and to its intermediate thoughts~(\Cref{sec:evaluations}).
We find that \ours yields reliable attributions comparable to approaches that involve performing ablations for every example, while being several times more efficient.
In particular, \ours is effective even when na\"ive approaches for attributing with attention (e.g., averaging attention weights across layers and heads) are quite inaccurate.
To assess the downstream utility of \ours, we use it to prune unimportant pieces of context in a context-based question answering setting~(\Cref{sec:applications}).
Doing so improves answer quality across models on HotpotQA~\citep{yang2018hotpotqa}.

\section{Problem statement}
\label{sec:problem_statement}
We begin by formalizing the task of \emph{token attribution}: identifying the tokens that \emph{influence} a model to generate a particular sequence (\Cref{sec:token_attribution}).
Next, we present metrics for evaluating the quality of attribution methods (\Cref{sec:evaluation_metrics}).
Finally, we describe two examples of specific token attribution tasks that we will focus on in this work: \emph{context attribution} and \emph{thought attribution} (\Cref{sec:examples}).

\subsection{Token attribution}
\label{sec:token_attribution}

\paragraph{Setup.}

Let $\plm$ be an autoregressive language model, with $\plm(t_i\mid t_1,\hdots,t_{i-1})$ denoting the probability that the next token is $t_i$ given preceding tokens $t_1,\hdots,t_{i-1}$.
Suppose that we use $\plm$ to generate a sequence of tokens $Y=y_1,\hdots,y_{|Y|}$ from an input sequence $X=x_1,\hdots,x_{|X|}$.
Concretely, we generate the $i$'th token $y_i$ as follows:
\[y_i\sim \plm(\cdot\mid x_1,\hdots,x_{|X|},y_1,\hdots,y_{i-1}).\]
We let $\plm(Y\mid X)$ denote the probability of generating the entire sequence $Y$ given $X$.
The input sequence $X$ might be a user-provided prompt, a series of messages between a user and the language model itself, or a piece of writing for the language model to extend.

\paragraph{Attributing generation to sources.}

Token attribution involves understanding the extent to which different parts of the input sequence $X$ contribute to the generated sequence $Y$.
We refer to these ``parts of $X$'' as \emph{sources}, where each source is just a subsequence of $X$.
The sources that we consider depend on the specific attribution task.
They might be sentences, paragraphs, or individual tokens;
they might span the entirety of $X$ or just a particular subset of interest.
If, for example, the input sequence $X$ consists of a context and a query, we might be specifically interested in understanding how the model uses just the context.

We denote by $\smash{S = s_1,\hdots,s_{|S|}}$ the set of sources associated with our attribution example, with each source $\smash{s_i\in\text{Subseq}(X)}$.
Our goal is to assign a score $\tau_i$ to each source $s_i$ that reflects the \emph{influence} of $s_i$ when generating $Y$.
We formalize this goal by defining a \emph{token attribution method}:

\begin{definition}[Token attribution method.]
    Suppose that we are given a language model $\plm$, an input sequence $X$, a sequence $Y$ generated from $X$ by $\plm$, and a set of sources $S$ over $X$.
    A \emph{token attribution method} is a function $\tau(\plm,X,S,Y)\in\mathbb{R}^{|S|}$ that assigns an attribution score to each source signifying its influence.
\end{definition}

\subsection{Measuring the quality of attributions}
\label{sec:evaluation_metrics}

We have established that attribution scores should reflect the extent to which each source $s_i$ \emph{influences} the model to generate the sequence $Y$---how do we formalize this?
Following prior work~\citep{lundberg2017unified,arras2017relevant,deyoung2019eraser,ilyas2022datamodels,madsen2021evaluating,cohenwang2024contextcite}, we adopt the perspective that if a source $s_i$ influences the model to generate $Y$, then \emph{removing} $s_i$ should decrease the probability of generating $Y$.
This perspective offers a natural methodology for evaluating the quality of attribution scores.
Intuitively, when higher-scoring sources are removed, the probability of $Y$ should decrease more.

To operationalize this intuition, we first introduce the notion of a \emph{source ablation}.
Let $v\in\{0,1\}^{|S|}$ be an \emph{ablation vector} indicating a subset of the sources to ablate.
We denote by $\ablate(X,S,v)$ a modified input sequence in which we remove every source $s_i$ for which $v_i=0$.
Our first metric, the \emph{top-$k$ drop}~\citep{cohenwang2024contextcite}, measures the decrease in the log-probability of generating $Y$ when we ablate the $k$ highest-scoring sources:

\begin{definition}[\emph{Top-$k$ drop}]
    \label{def:top_k_drop}
    Suppose that we are given an attribution method $\tau$.
    Let $v_{\text{top-}k}(\tau)$ be an ablation vector that excludes the $k$ highest-scoring sources according to $\tau$.
    Then the \emph{top-$k$ drop} is the decrease in the log-probability of generating $Y$ when these top-$k$ sources are ablated:
    \begin{align*}
        \label{eq:top_k_drop}
        \text{Top-}k\text{-drop}(\tau)\coloneqq\underbrace{\log p_\text{LM}(Y\mid X)}_{\text{original log-probability}}-\underbrace{\log p_\text{LM}(Y\mid \ablate(X,S,v_{\text{top-}k}(\tau)))}_{\text{log-probability with top-}k\text{ sources ablated}}.
    \end{align*}
\end{definition}

The top-$k$ drop evaluates the ability of an attribution method to identify the $k$ most important sources.
We would also like to consider whether the attribution scores of an \emph{arbitrary} set of sources reflect the effect of ablating them.
In particular, we treat the sum of the attribution scores of a set of sources as a prediction for the effect of ablating them.
If the sum of the attribution scores of one set of sources is higher than that of another, then ablating the first set should have a larger effect on the probability of generating $Y$.
More generally, if we perform a number of random ablations, their predicted effects should correlate with their actual effects.
To measure this, \citet{park2023trak} introduced the \emph{linear datamodeling score} (LDS):

\begin{definition}[\emph{Linear datamodeling score}]
    \label{def:lds}
    Let $\tau$ be an attribution method.
    Next, let $v^{(1)},\hdots,v^{(m)}$ be $m$ randomly sampled ablation vectors and let $f(v^{(1)}),\hdots,f(v^{(m)})$ be the corresponding probabilities of generating the sequence $Y$.
    That is, $f(v^{(j)})=\plm(Y\mid \ablate(X,S,v^{(j)}))$.
    Finally, let $\smash{\hat{f}_\tau(v)=\langle\tau(\plm,X,S,Y), v\rangle}$ be the sum of the scores (according to $\tau$) of sources that are included by ablation vector $v$.
    This sum represents the ``predicted effect'' of ablating according to $v$.
    Letting $\rho$ denote the Spearman rank correlation coefficient \citep{spearman}, the \emph{linear datamodeling score} (LDS) of $\tau$ is
    \begin{equation*}
        \label{eq:lds}
        \text{LDS}(\tau)\coloneqq\rho(\underbrace{\{f(v^{(1)}), \hdots, f(v^{(m)})\}}_{\text{actual probabilities under ablations}},\;\; \underbrace{\{\hat{f}_\tau(v^{(1)}),\hdots,\hat{f}_\tau(v^{(m)})\}}_{\text{``predicted effects'' of ablations}}),
    \end{equation*}
\end{definition}

\subsection{Examples of token attribution tasks}
\label{sec:examples}

We now describe two examples of token attribution tasks that will be of interest to us.
In~\Cref{sec:evaluations}, we will be evaluating the effectiveness of attribution methods on these tasks.

\paragraph{Context attribution.}

The goal of \emph{context attribution}~\citep{cohenwang2024contextcite} is to understand how a language model uses different pieces of information from a provided context when responding to a query.
In this case, the input sequence $X$ consists of the provided context and the query.
The generated sequence $Y$ is the response (or any span from it\footnote{We may want to attribute a particular span, in which case $Y$ would be this span and $X$ would include everything leading up to it.}).
The sources might be the sentences, words, or tokens of the context.
Sources with high attribution scores can be interpreted as ``citations'' for the model's response.
These citations can help verify the correctness of the response, particularly when the context is long and difficult to read through entirely.

\paragraph{Thought attribution.}

As a second task, we propose \emph{thought attribution}.
Certain recent language models have been designed to generate ``thoughts'' before responding to a query~\citep{jaech2024openai,guo2025deepseek}.
These thoughts can significantly improve response quality.
When models generate intermediate thoughts, the final response is often a highly distilled version of them.
To understand how a model arrives at a particular conclusion presented in the response, it would thus be helpful to attribute this conclusion to its thoughts.
In this setting, the input sequence $X$ consists of the original query and the model-generated thoughts.
The generated sequence $Y$ to be attributed is the final response (or any span from it).
The sources we consider are parts of the thoughts, e.g., sentences.

\section{\ours: Token attribution with attention}
\label{sec:method}
In this section, we describe Attribution with Attention (\ours), our method for token attribution.
We first outline \emph{attribution via surrogate modeling}, a general framework for attribution that has been applied extensively in prior work~(\Cref{sec:background}).
The key idea of this framework is to identify an easy-to-understand proxy that models the effects of ablating sources.
By studying this proxy, we can attribute the model's generation to individual sources.
Building on this framework, \ours learns a surrogate model that uses attention weights as features, making attribution for unseen examples highly efficient~(\Cref{sec:approximate_weights}).

\subsection{Background: attribution via surrogate modeling}
\label{sec:background}

To motivate \ours, we first describe \emph{surrogate modeling}~\citep{designofexperiments}, a technique widely used for attributing model behavior~\citep{ribeiro2016should,ilyas2022datamodels,shah2024decomposing,cohenwang2024contextcite}.
This technique aims to identify a simple proxy \emph{surrogate model} that approximates a model's behavior while being easy to understand.
This surrogate model can then be used to shed light on the model's behavior, and, in this case, to attribute behavior to individual sources.

To make this concrete, consider the token attribution setting: we would like to attribute a generation $Y$ to a set of sources $S$, each of which is a subsequence of the input $X$.
Let $f(v)$ be the probability of generating $Y$ when ablating the sources according to $v$, that is,
\[f(v)\coloneqq\plm(Y\mid\ablate(X,S,v)).\]
We would like to understand how $f$ varies as a function of $v$;
are there specific sources that signficantly affect $f$ when ablated?
To answer this question, suppose that we could approximate $f$ with a \emph{linear} surrogate model of the following form:
\[\smash{\hat{f}_w(v)\coloneqq\langle w,v\rangle}.\]
To understand $f$, we could then simply examine the coefficients $w$ of $\hat{f}_w$.
Each coefficient $w_i$ would signify the effect of ablating the $i$'th source on the probability of generating $Y$---this value could be directly interpreted as an attribution score.
While the assumption of a \emph{linear} model may seem strong, prior work has empirically shown that this modeling choice is often effective\footnote{Specifically, a linear surrogate model can attain a high LDS (\Cref{def:lds}), meaning that it faithfully predicts the effects of ablations.}.
Indeed, this type of surrogate model has been successfully applied to attribute model behavior to training examples~\citep{ilyas2022datamodels,park2023trak}, features~\citep{ribeiro2016should,lundberg2017unified}, model internals~\citep{shah2024decomposing}, and in-context information~\citep{cohenwang2024contextcite}.

\paragraph{Learning a linear surrogate model.}

With a linear surrogate model of the form $\hat{f}_w$, our remaining goal is to find parameters $w$ to approximate $f$ as well as possible.
We do so by sampling random ablations and optimizing the parameters $w$ to minimize the empirical difference between $\hat{f}_w$ and $f$, i.e., optimizing $\hat{f}_w$ to predict the effects of ablations.
Given an example $(\plm,X,S,Y)$, this process consists of the following steps:

\begin{enumerate}
    \item Sample $m$ ablation vectors $v^{(1)},\hdots,v^{(m)}\in\{0,1\}^{|S|}$.
    \item For each ablation $v^{(j)}$, compute $f(v^{(j)})$, the corresponding probability of generating $Y$.
    \item Find parameters $\hat{w}$ that minimize a loss $\mathcal{L}$ (e.g., MSE) between $f$ and $\hat{f}_w$:
    \[\hat{w}=\arg\min_w\mathcal{L}(\{f(v^{(j)})\}_j, \{\hat{f}_w(v^{(j)})\}_j).\]
\end{enumerate}

Before proceeding, we highlight a subtle yet important property of this type of surrogate model: it is \emph{example-specific}.
It approximates a model's behavior for a certain example and can only provide attribution scores for the particular sources associated with this example.
As a result, to perform attribution for any given example, we would need to learn a new surrogate model from scratch for it (which involves performing several ablations to learn from).

\subsection{Learning a generalizable surrogate model using attention}
\label{sec:approximate_weights}

\begin{figure*}[t]
    \input{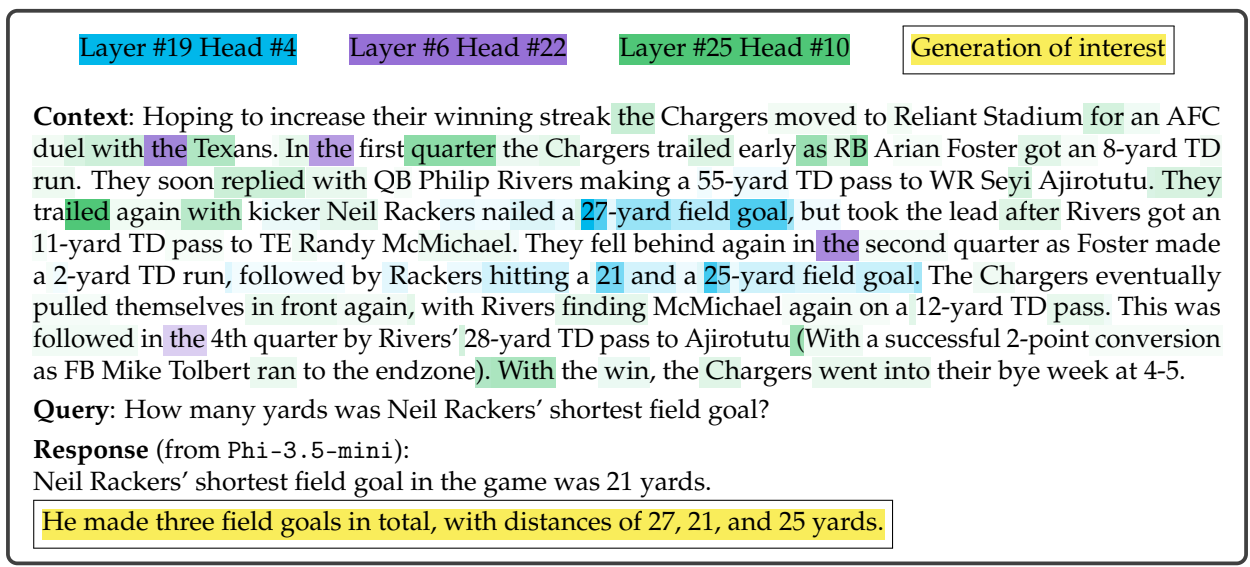}
    \caption{
        \textbf{Attention heads vary in their usefulness for attribution.}
        When visualizing attention weights of three individual heads, we observe that certain heads appear to be more useful for attribution than others.
        In particular, \hla{layer \#19, head \#4} assigns high attention weights to ``27'', ``21'' and ``25'' which are the field goal distances mentioned in the generation of interest.
        Meanwhile, \hlb{layer \#6, head \#22} and \hlc{layer \#25, head \#10} assign high attention weights to other, seemingly unrelated parts of the context.
        This example is from DROP~\citep{dua2019drop} with a generation from \texttt{Phi-3.5-mini}~\citep{abdin2024phi}.
        Attention weights are averaged across the generation of interest and normalized by dividing by the maximum weight for each head.
    }
    \label{fig:attention_heads}
\end{figure*}

In \Cref{sec:background}, we reviewed \emph{surrogate modeling} as a method for attributing model behavior.
Using a surrogate model to model the effects of ablations is an effective approach across attribution settings.
However, the standard instantiation of this approach---learning an \emph{example-specific} surrogate model for each attribution---may be prohibitively costly.
The design of \ours is motivated by the following question: 
\begin{center}
    \emph{Can we learn to model the effects of ablations in a way that extends beyond a specific example?}
\end{center}
If so, we would just need to learn the surrogate model's parameters once.
Attributing a \emph{new} example would be cheap (assuming the surrogate model is cheap to evaluate).

\paragraph{Learning a generalizable surrogate model.}

Recall that the example-specific surrogate model requires learning a coefficient $w_i$ for each source, which is interpreted as an attribution score for this source.
Our approach to identify a surrogate model is as follows:
instead of directly learning attribution scores $w$ for particular sources, we learn a parameterized function $w_\theta(X,S,Y)$ that estimates scores \emph{across} examples and sources (for a particular language model $\plm$).
This score-estimating function would use information about the language model $\plm$, the input sequence $X$, the sources $S$, and the generated sequence $Y$ as its features.
Its learnable parameters, $\theta$, would be optimized to predict the effects of ablations (just as we optimize attribution scores $w$ directly for example-specific surrogate models).
The resulting surrogate model for predicting the probability of generating $Y$ would then be
\[\hat{f}_\theta(v,X,S,Y)\coloneqq\langle w_\theta(X,S,Y),v\rangle.\]
In order to be useful, such a score-estimating function $\smash{w_\theta}$ needs to be both (1) cheap to compute and (2) effective at modeling the effects of ablations.
To satisfy these criteria, the first step is designing features for $\smash{w_\theta}$, that is, identifying information that is useful for attribution and cheap to extract from $\plm$, $X$, $S$, and $Y$.
The second step is learning parameters $\theta$ to make use of these features to predict the effects of ablations.

\paragraph{Designing features for $w_\theta$.}

To obtain features for $w_\theta$ that are both cheap to compute and provide signal for attribution, we consider using artifacts from the model's generative process (these would require \emph{no} additional computation).
In particular, for transformer models~\citep{vaswani2017attention}, attention weights are a natural candidate for the features of $\smash{w_\theta}$.
Prior work suggests that specific attention heads are responsible for particular behaviors~\citep{zheng2024attention} and have distinct utilities~\citep{chuang2024lookback}.
We illustrate this in \Cref{fig:attention_heads}: when we visualize the attention weights of a few attention heads, we observe that some appear to be more useful for attribution than others.

Motivated by this observation, we use the attention weights of individual heads as features for attribution.
This way, we can learn to rely on specific heads that more accurately reflect the model's use of preceding tokens.
To formalize this, we begin by introducing notation for describing attention weights.
Recall that the transformer architecture consists of $L$ layers with $H$ heads in each layer.
We write $\attn(X,Y,i,j)\in \mathbb{R}^{L\times H}$ to denote the attention weights (across layers and heads) assigned to the $i$-th token of $X$ when generating the $j$-th token of $Y$.
We are interested in attributing $Y$ to individual sources $\smash{s_1,\hdots,s_{|S|}\in\text{Subseq}(X)}$.
Thus, we aggregate attention weights over $Y$ and a source $s$ as follows (overloading the ``\attn'' notation):
\[\aattn(X,Y,s)\coloneqq\frac{1}{|Y|}\sum_{j=1}^{|Y|}\sum_{i\in s}\attn(X,Y,i,j).\]
We are now ready to express $\smash{w_\theta}$ using these aggregated attention weights as features.
The $i$-th element of $\smash{w_\theta}$, i.e., the attribution score for the $i$-th source $s_i$ is given by
\[w_\theta(X,S,Y)_i=\sum_{\ell=1}^L\sum_{h=1}^H\theta_{\ell h}\aattn(X,Y,s_i)_{\ell h}.\]
Here, $\theta\in\mathbb{R}^{L\times H}$ are coefficients specifying the extent to which we rely on each attention head.

\paragraph{Learning the parameters $\theta$.}

The next step is to learn parameters $\theta$ such that $w_\theta$ predicts effective attribution scores.
To do so, we apply the same methodology as in \Cref{sec:background}: we perform a small number of random ablations and learn $\theta$ to predict the effects of these ablations (via the surrogate model $\smash{\hat{f}_\theta}$).
However, instead of learning from a single example, we learn from a \emph{dataset} of examples (each consisting of an input sequence $X$, a set of sources $S$ and a generated sequence $Y$).
In doing so, we hope to identify parameters $\theta$ that generalize to unseen examples.
We summarize the resulting method, \ours, in~\Cref{alg:ours}.

\begin{algorithm}
\caption{Attribution with Attention (\ours)}
\label{alg:ours}
\begin{algorithmic}[1]
    \State {\bfseries Input:} Transformer language model $\smash{\plm}$, training dataset of $n$ examples $\smash{\{X^{(i)},S^{(i)},Y^{(i)}\}_{i=1}^n}$, number of ablations $m$ to perform per example, loss function $\smash{\mathcal{L}}$ measuring surrogate model quality.
    
    \State {\bfseries Output:} Learned attribution method $\hat{\tau}_\text{\ours}$
    \State $w_\theta(X,S,Y)_i\coloneqq\sum_{\ell,h}\theta_{\ell h}\aattn(X,Y,s_i)_{\ell h}$
    \hfill$\triangleright$ Score-estimating function with attention weights as features
    \State $\hat{f}_\theta(v,X,S,Y)\coloneqq \langle w_\theta(X,S,Y), v\rangle$
    \hfill$\triangleright$ Surrogate model (linear in the ablation vector $v$)
    \For{$i \in \{1,\ldots,n\}$}
        \For{$j \in \{1,\ldots,m\}$}
            \State Sample ablation $v^{(j)}$ from $\{0,1\}^{|S^{(i)}|}$
        \EndFor
        \State $f^{(i)}(v) \coloneqq p_\text{LM}(Y^{(i)}\mid\ablate(X^{(i)},S^{(i)},v))$
        \hfill$\triangleright$ Probability of $Y^{(i)}$ when ablating $X^{(i)}$ according to $v$
        \State $\hat{f}_\theta^{(i)}(v)\coloneqq \hat{f}_\theta(v,X^{(i)},S^{(i)},Y^{(i)})$
        \hfill$\triangleright$ Surrogate model prediction when ablating $X^{(i)}$ according to $v$
        \State $\mathcal{L}^{(i)}(\theta)\coloneqq \mathcal{L}(\{\hat{f}_\theta^{(i)}(v^{(j)})\}_j,\{f(v^{(j)})\}_j)$
        \hfill$\triangleright$ Loss for $i$'th example (across $m$ ablations)
    \EndFor
    \State $\hat{\theta}\gets\arg\min_\theta\frac{1}{n}\sum_{i=1}^n\mathcal{L}^{(i)}(\theta)$
    \hfill$\triangleright$ Optimize parameters $\theta$ to minimize loss across examples
    \State $\hat{\tau}_\text{\ours}(\plm,X,S,Y)\coloneqq w_{\hat{\theta}}(X,S,Y)$
    \hfill$\triangleright$ Interpret score-estimating function as an attribution method
    \State \textbf{return} $\hat{\tau}_\text{\ours}$
\end{algorithmic}
\end{algorithm}

\paragraph{Implementation details.}

The following are a few design choices for \ours that we found to be effective.
We sample ablation vectors $v$ uniformly from $\smash{\{0,1\}^{|S|}}$, that is, each source is ablated independently with probability $1/2$.
To ablate a source, we use an attention mask to ignore the source's tokens across layers and heads.
For the loss function used to measure how well the surrogate model $\smash{\hat{f}_\theta}$ approximates $f$, we consider the negative Pearson correlation~\citep{pearson1895vii}.
Instead of directly computing the correlation between surrogate model predictions and probabilities of generating $Y$, we apply a \emph{logit} transformation to these probabilities;
this transformation maps probabilities to values in $(-\infty,\infty)$ and is more natural for measuring linear correlation\footnote{This transformation has been used in prior work for linear surrogate models when predicting probabilities~\citep{park2023trak,cohenwang2024contextcite}.}.
Lastly, to minimize the loss, we sample batches of examples and optimize with Adam~\citep{kingma2015adam} (see~\Cref{sec:ours_details} for additional details and hyperparameters used in our experiments).

\paragraph{Visualizing the attention weights of heads used and ignored by \ours.}

\begin{figure*}[t]
    \input{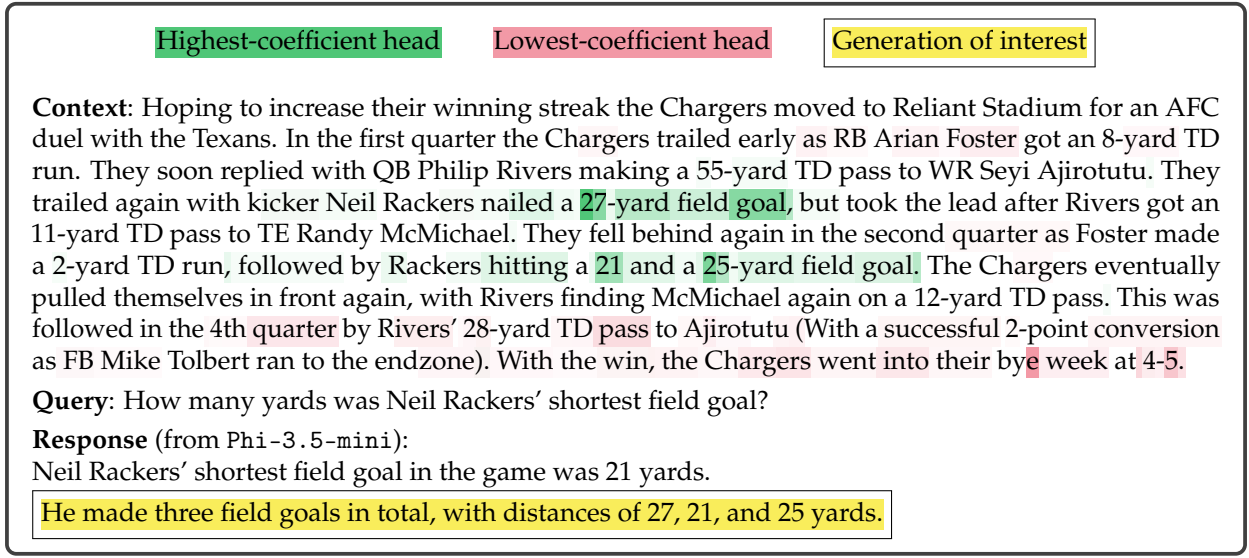}
    \caption{
        \textbf{\ours identifies attention heads that appear useful for attribution.}
        The attention head with the \hla{highest coefficient} (as learned by \ours) attends to tokens that seem very relevant to the generation of interest.
        On the other hand, the head with the \hlb{lowest-magnitude coefficient} attends to other, seemingly unrelated parts of the context.
        To learn these coefficients, \ours is trained on \emph{Dolly 15k}~\citep{DatabricksBlog2023DollyV2} (see \Cref{sec:evaluations} for details).
        The details are otherwise identical to \Cref{fig:attention_heads}.
        See \Cref{sec:coefficients} for the coefficients themselves.
    }
    \label{fig:learned_attention_heads}
\end{figure*}

Previously, in \Cref{fig:attention_heads}, we observed that certain attention heads appear to be more useful for attribution than others.
In \Cref{fig:learned_attention_heads}, we find that the attention heads that \ours learns to rely on align with this intuition: the highest-coefficient head attends to tokens that seem very relevant to the generation of interest, while the lowest-coefficient head attends to other, seemingly unrelated tokens.

\section{Evaluating token attribution methods}
\label{sec:evaluations}
To evaluate the effectiveness of \ours, we consider two token attribution tasks: \emph{context attribution} (\Cref{sec:context_attribution}) and \emph{thought attribution} (\Cref{sec:thought_attribution}).
See~\Cref{sec:implementation_details,sec:dataset_details,sec:model_details,sec:regression_details,sec:ours_details} for additional details and~\Cref{sec:additional_evaluations} for more fine-grained evaluations.

\paragraph{Baseline attribution methods.}

In addition to \ours, we evaluate the following attribution methods:

\begin{enumerate}
    \item \emph{Example-specific surrogate modeling (ESM)}:
    We attribute using an example-specific linear surrogate model as described in~\Cref{sec:background}.
    With a sufficient number of ablations, an example-specific surrogate model is an oracle of sorts for \ours (which learns a surrogate model to generalize across examples).
    We follow the implementation of~\citet{cohenwang2024contextcite} (see~\Cref{sec:regression_details}) using $32,64,128$ and $256$ ablations.

    \item \emph{Average attention}:
    As a baseline for leveraging attention weights, we consider simply averaging the attention weights over attention heads (\ours learns a coefficient for each head).
    This strategy has been used to explain model behavior~\citep{kim2019and,sarti2023quantifying}.
    In the context attribution setting of~\citet{cohenwang2024contextcite}, average attention performed the best among attention-based methods and yielded effective attributions (i.e., close in performance to surrogate modeling) for some models but not others.

    \item \emph{Gradient $\ell_1$-norm}: 
    The gradient with respect to a preceding token is another measure of its influence~\citep{simonyan2013deep,li2015visualizing,smilkov2017smoothgrad}.
    As another baseline, we compute the gradient of the log-probability of the generated sequence with respect to the embeddings of preceding tokens.
    To compute a score for each source, we sum the $\ell_1$-norm of the gradients for the source's tokens.
\end{enumerate}

\paragraph{Experiment setup.}

For each of the context and thought attribution settings, we consider several datasets.
For each dataset, we sample $400$ examples from the validation split for evaluation.
Each example consists of a prompt to which we generate a response (and intermediate thoughts where applicable) using a language model fine-tuned to follow instructions.
When the response consists of multiple sentences (according to an off-the-shelf sentence tokenizer~\citep{bird2009natural}), we consider each sentence as a separate attribution target.
We evaluate each method by measuring its top-$5$ log-probability drop (\Cref{def:top_k_drop}) and LDS (\Cref{def:lds}), averaged across examples and attribution targets.

\paragraph{Training \ours.}

A key ingredient of \ours is the training dataset used to learn the attention head coefficients $\theta$.
In some cases, we may have access to a dataset matching the distribution of examples we would like to attribute.
In others, we may only have access to, e.g., a generic instruction following dataset.
To evaluate \ours, we consider both settings.
We write \ours (task-specific) to denote training using examples from the task of interest, and \ours (general) to denote training using examples from a generic dataset.
We train \ours using $2,000$ examples sampled from the training split of each dataset, performing $32$ ablations for each example.
Hence, the cost of training \ours in this way is roughly equivalent to the cost of $64,000$ forward passes of the language model.
We always train \ours using individual tokens as sources (we find that the resulting parameters transfer well to less fine-grained sources, e.g., sentences) and sentences as attribution targets.

\begin{figure}[t]
    \centering
    \input{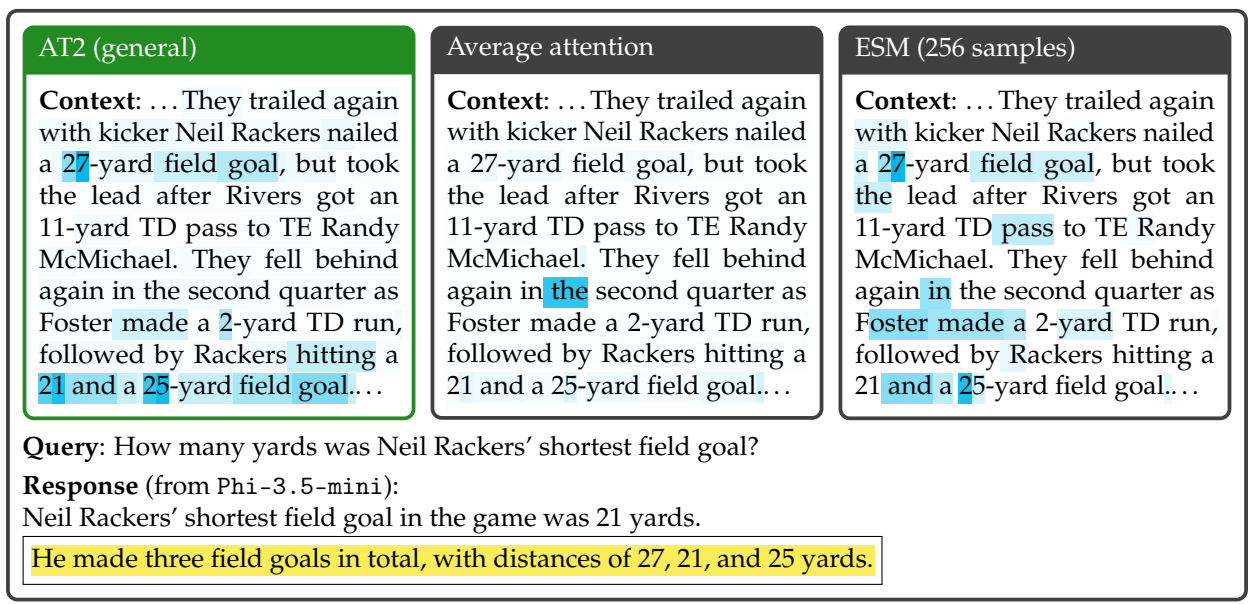}
    \caption{
        \textbf{Qualitative comparison of token attributions.}
        We visualize the attribution scores (blue) of different methods for a particular generated statement (yellow) in a context attribution setting for \texttt{Phi-3.5-mini} (with individual tokens as sources).
        \ours (trained on a generic dataset) and ESM (with $256$ ablations) yield similar attributions with high scores for tokens related to the generated statement of interest, while average attention assigns the highest score to a seemingly arbitrary token.
        See~\Cref{sec:attribution_examples} for additional examples of attributions.
    }
    \label{fig:method_comparison}
\end{figure}

\begin{figure*}[p]
    \centering
    \begin{subfigure}[b]{\textwidth}
        \includegraphics[width=\textwidth]{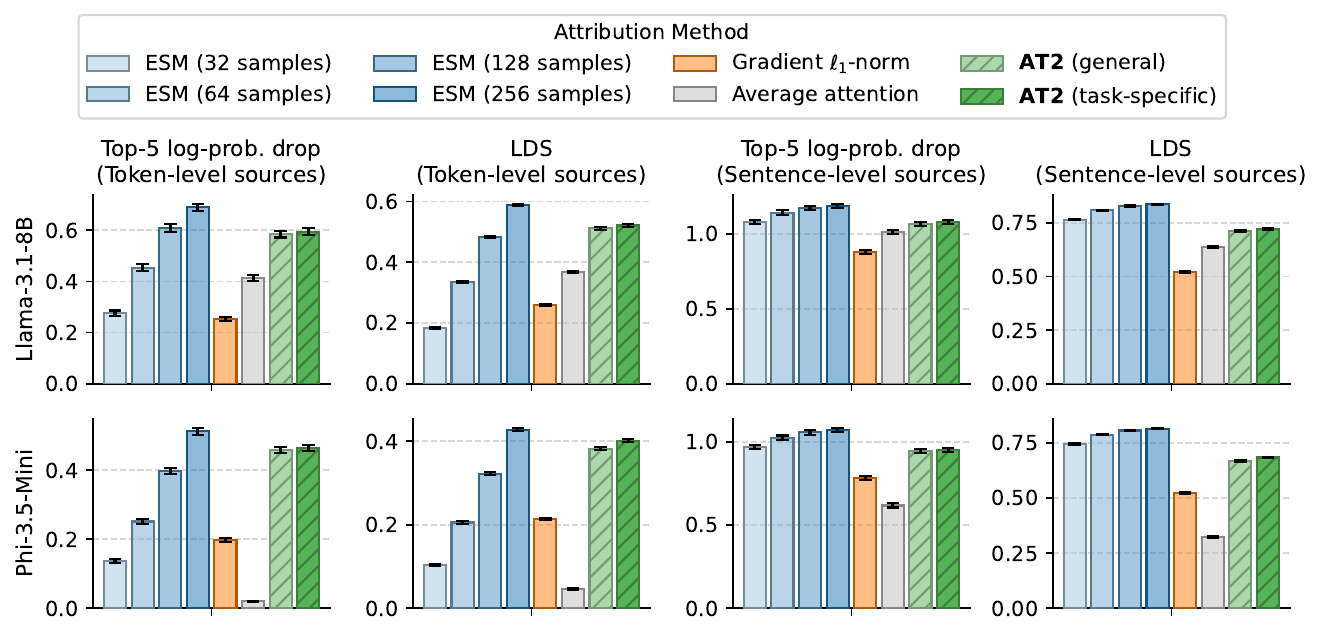}
        \caption{
            \textbf{Evaluating context attributions.}
            We report the log-probability drop (\Cref{def:top_k_drop}) and LDS (\Cref{def:lds}) for different attribution methods applied to \texttt{Llama-3.1-8B} and \texttt{Phi-3.5-Mini}.
            We consider \emph{individual tokens} as sources (left) and \emph{sentences} as sources (right).
            Metrics are averaged across different context attribution tasks (see~\Cref{sec:context_attribution}).
        }
        \label{fig:context_attribution}
    \end{subfigure}\hfill
    \begin{subfigure}[b]{\textwidth}
        \includegraphics[width=\textwidth]{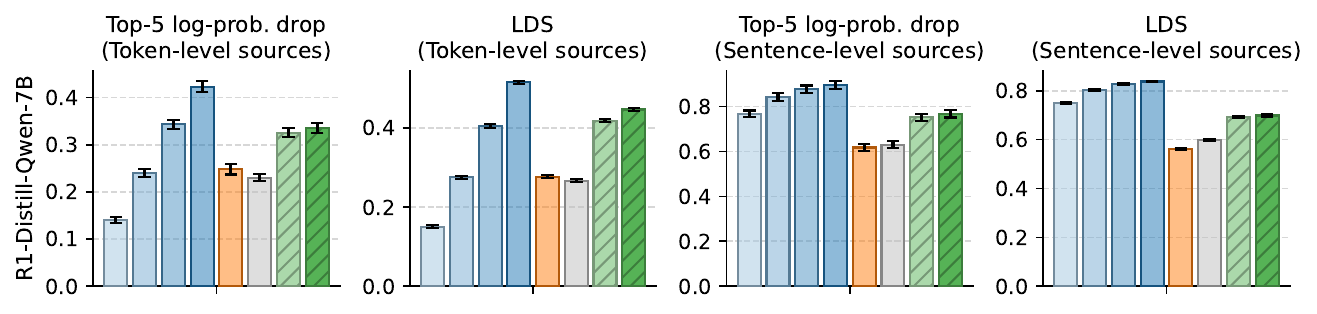}
        \caption{
            \textbf{Evaluating thought attributions.}
            We report the log-probability drop (\Cref{def:top_k_drop}) and LDS (\Cref{def:lds}) for different attribution methods applied to \texttt{DeepSeek-R1-Qwen-7B}.
            We consider \emph{individual tokens} as sources (left) and \emph{sentences} as sources (right).
            Metrics are averaged across different thought attribution tasks (see~\Cref{sec:thought_attribution}).
        }
        \label{fig:thought_attribution}
    \end{subfigure}\hfill
    \begin{subfigure}[b]{\textwidth}
        \includegraphics[width=0.85\textwidth]{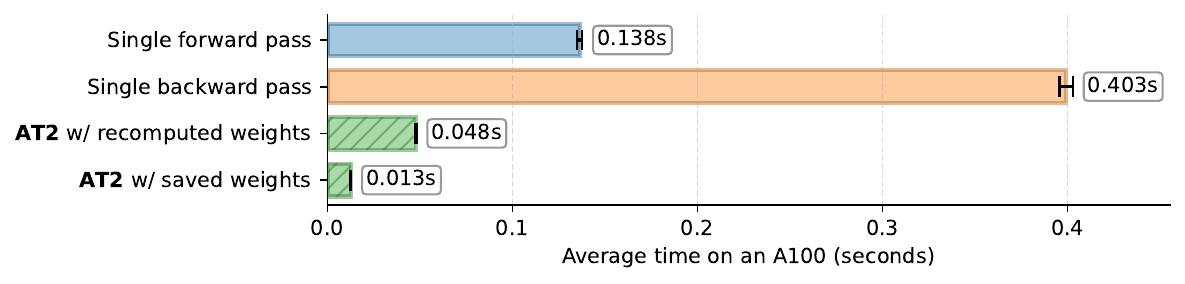}
        \caption{
            \textbf{Comparing efficiency.}
            We report the average running time for a single forward pass, a single backward pass, and for \ours for \texttt{Llama-3.1-8B} on Hotpot QA (on an A100 GPU).
            We consider \ours with saved attention weights and with recomputed attention weights (for certain attention implementations, weights cannot be easily saved).
        }
        \label{fig:efficiency_comparison}
    \end{subfigure}\hfill
    \caption{
        \textbf{Evaluating token attributions.}
        We report performance metrics for different attribution methods in context attribution (\Cref{fig:context_attribution}) and thought attribution (\Cref{fig:thought_attribution}) settings.
        \ours performs similarly when trained on examples from the task of interest (task-specific) or when trained on examples from a generic task (general).
        It consistently outperforms the gradient and average attention baselines and performs comparably to example-specific surrogate modeling (ESM) with a substantial number of ablations.
        In~\Cref{fig:efficiency_comparison}, we compare the running time of \ours to a single forward pass (ESM uses $\geq32$) and a single backward pass (which the gradient method requires).
        See~\Cref{sec:additional_evaluations} for additional evaluations.
    }
    \label{fig:attribution_evaluation}
\end{figure*}

\subsection{Context attribution}
\label{sec:context_attribution}

Context attribution (see \Cref{sec:examples}) is the task of understanding how language models use different parts of a provided context when generating a response.
In their work,~\citet{cohenwang2024contextcite} focus on \emph{sentences} as sources;
we consider both sentences and individual tokens from the context as sources.

\paragraph{Models.}

We attribute the variants of \texttt{Llama-3.1-8B}~\citep{dubey2024llama} and \texttt{Phi-3.5-Mini}~\citep{abdin2024phi} fine-tuned to follow instructions.

\paragraph{Datasets.}

For evaluation, we consider \emph{CNN DailyMail}~\citep{nallapati2016abstractive}, a news article summarization dataset, \emph{Hotpot QA}~\citep{yang2018hotpotqa}, a multi-hop question answering dataset, \emph{Natural Questions}~\citep{kwiatkowski2019natural}, a long-context question answering dataset, and \emph{MS MARCO}~\citep{nguyen2016ms}, a multi-document question answering dataset.
As a generic context attribution dataset for training \ours, we consider \emph{Dolly 15k}~\citep{DatabricksBlog2023DollyV2}, an instruction following dataset (filtered to only examples with a context).

\subsection{Thought attribution}
\label{sec:thought_attribution}

Thought attribution (see \Cref{sec:examples}) is the task of understanding how language models use intermediate thoughts when generating a response.
In this evaluation, we attribute to both intermediate thoughts and context (when present).
As in the context attribution evaluation, we consider both sentences and individual tokens as sources.

\paragraph{Models.}

We attribute a distilled reasoning model, \texttt{DeepSeek-R1-Qwen-7B}~\citep{guo2025deepseek}.

\paragraph{Datasets.}

For evaluation, we consider \emph{DROP} \citep{dua2019drop}, a context-based question answering dataset focused on reasoning, and \emph{Global Opinions QA}~\citep{durmus2023towards}, a dataset of questions of opinions on global issues.
As a generic thought attribution dataset for training \ours, we consider \emph{AGIEval}~\citep{zhong2023agieval}, a dataset of reasoning problems spanning law, logic and math.

\subsection{Results}

We provide a qualitative comparison of the attributions of different methods in~\Cref{fig:method_comparison}.
In~\Cref{fig:context_attribution,fig:thought_attribution}, we find that \ours consistently outperforms the gradient baseline, which requires substantial additional computation, as well as the average attention baseline (a na\"ive application of attention for attribution).
\ours performs comparably to example-specific surrogate modeling with a substantial number of ablations, even when average attention performs quite poorly (as is the case for \texttt{Phi-3.5-mini}).
We observe that the learned parameters $\theta$ (signifying reliance on each attention head) are fairly robust to the distribution of training examples: \ours performs similarly when trained using examples from the task of interest (task-specific), or when trained using examples from a generic dataset (general).
Furthermore, despite being trained only with individual tokens as sources, \ours performs well when considering sentences as sources.

\paragraph{Attribution efficiency.}

In~\Cref{fig:efficiency_comparison}, we compare the running time of our implementation of \ours to a single forward pass and a single backward pass on a single A100 GPU.
We consider two settings for \ours: one in which we have access to saved attention weights and one in which we recompute the relevant attention weights from hidden states.
The latter setting arises when using attention implementations that do not store the entire attention matrix (e.g., FlashAttention~\citep{dao2022flashattention}).
Even when recomputing attention weights, \ours is more than $2\times$ faster than one forward pass and $8\times$ faster than one backward pass.
This is substantially faster than methods involving ablations (which require $\geq32$ forward passes for similar performance) or gradient-based methods (which require at least one backward pass).

\begin{figure}[t]
    \centering
    \includegraphics[width=0.95\textwidth]{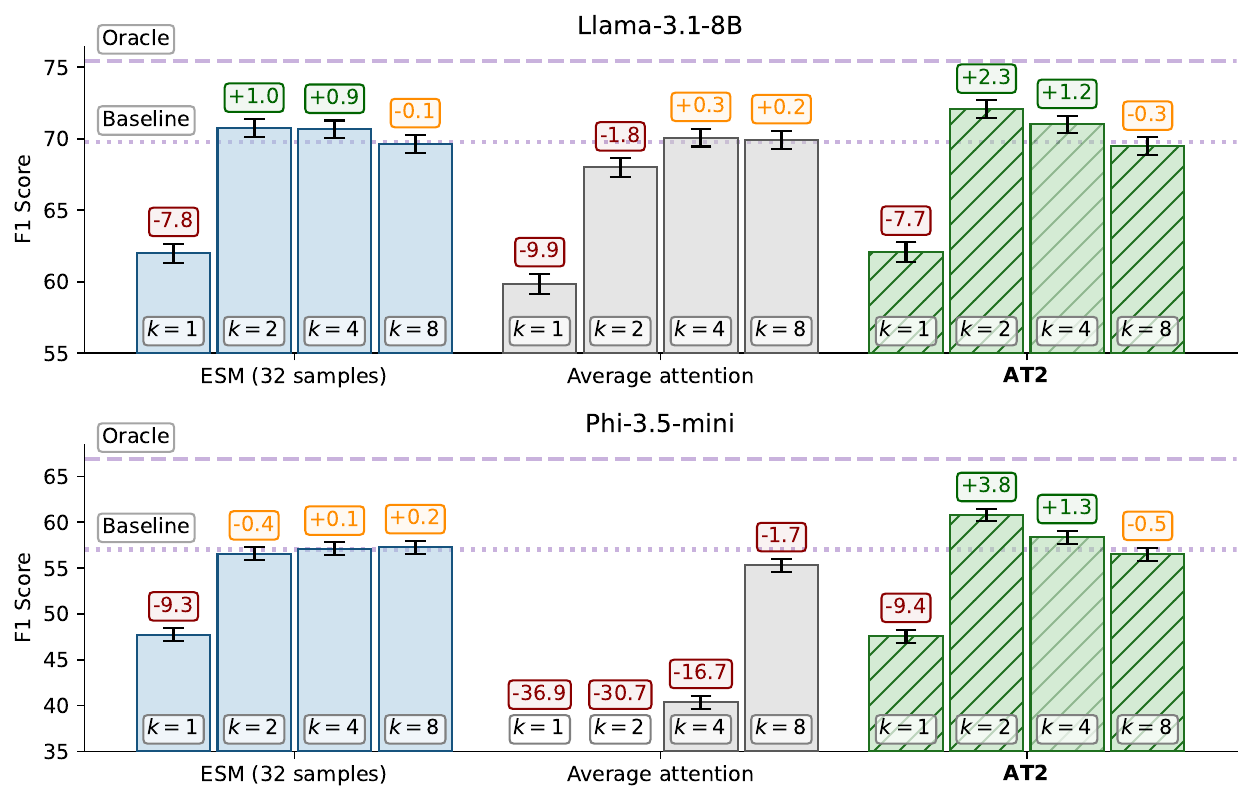}
    \caption{
        \textbf{Improving response quality by pruning the context.}
        Using attributions from \ours to prune away less important parts of the context improves response quality on HotpotQA. 
        We compare \ours to example-specific surrogate modeling (ESM) with $32$ ablations and average attention for different numbers of retained passages $k$, finding that \ours improves performance more while being less costly than a single inference pass.
        ``Baseline'' denotes the performance of the model without context pruning, while ``Oracle'' denotes the performance of the model when provided only with the ground-truth sources.
        The F1 score is averaged over $4,000$ examples from the HotpotQA validation set.
    }
    \label{fig:pruning}
\end{figure}

\section{Applying \ours to improve response quality}
\label{sec:applications}
Beyond directly measuring the attribution performance of \ours, we examine its utility in a downstream application of context attribution: pruning the context to improve response quality.
Specifically, ~\citet{cohenwang2024contextcite} find that in a context-based question answering setting, pruning away parts of the context that are unused by the model (according to their attribution method) can improve performance.
This aligns with previous observations that language models often struggle to answer questions when the relevant information is within long contexts~\citep{peysakhovich2023attention,liu2024lost}.
The attribution method proposed by~\citet{cohenwang2024contextcite}, an example-specific surrogate modeling method, requires several inference passes (at least $32$) to identify the most important parts of the context.
Hence, improving performance comes at a substantially increased inference cost.

In~\Cref{fig:pruning}, we investigate whether attributions from \ours can be used to prune the context to improve response quality on Hotpot QA.
Hotpot QA is a multi-hop question answering dataset in which the answer involves information from two different passages among several provided passages (most of which are ``distractors'' that are irrelevant to the question).
In this setting, if we provide the model with just the two relevant passages (``Oracle'' in \Cref{fig:pruning}), it performs substantially better than when provided with the entire context (``Baseline'' in \Cref{fig:pruning}).
We find that pruning according to \ours attributions improves response quality and is more effective than using attributions from example-specific surrogate modeling with $32$ ablations and average attention.
When using \ours, the cost of these performance gains (besides re-generating the answer) is substantially less than a single additional forward pass.
We provide additional details in~\Cref{sec:pruning_details}.

\section{Related work}
\label{sec:related_work}

\paragraph{Attributing generation to preceding tokens.}

Attributing model generation to preceding tokens is valuable in a variety of settings.
For example, \emph{context attribution} seeks to pinpoint the parts of a provided context that a language model uses when making a particular statement~\citep{cohenwang2024contextcite,Qi2024ModelIA,liu2024attribot}.
Such attributions can also be used to detect hallucinations~\citep{chuang2024lookback}, identify biases~\citep{vig2020causal}, and assess the faithfulness of model-provided explanations~\citep{deyoung2019eraser,lanham2023measuring,madsen2024self}.
Methods for attribution include performing ablations~\citep{deyoung2019eraser,lanham2023measuring,cohenwang2024contextcite}, computing gradients~\citep{yin2022interpreting,enguehard2023sequential}, examining attention weights~\citep{vig2020causal}, and using embedding similarities~\citep{phukan2024peering}.

When attributing model behavior, we adopt the perspective that if a source is influential, then \emph{removing} it should significantly affect the model's behavior.
We use ablations as a source of ground-truth for both training \ours and evaluating the quality of attributions.
This definition of using ablations to quantify influence is common across a variety of attribution settings~\citep{lundberg2017unified,ilyas2022datamodels,shah2024decomposing,cohenwang2024contextcite}.
It has also been used to assess the faithfulness of interpretability methods~\citep{arras2017relevant,deyoung2019eraser,madsen2021evaluating}.

While we focus on identifying tokens that \emph{influence} a model's generation, attribution can also refer to identifying sources that \emph{support} a generated statement.
\citet{worledge2023unifying} refer to this type of attribution as \emph{corroborative attribution}.
In the context attribution setting, attributions are often corroborative and are described as ``citations'' for generated claims~\citep{menick2022teaching,gao2022rarr,rashkin2023measuring,gao2023enabling}.

\paragraph{Attribution via surrogate modeling.}

A general methodology for attributing model behavior is to learn a \emph{surrogate model} that approximates model behavior while being simple enough to interpret.
Learning a surrogate model for attribution involves ablating sources (e.g., features, training data, parameters) and measuring the corresponding effect on model behavior~\citep{ribeiro2016why,lundberg2017unified,ilyas2022datamodels,shah2024decomposing,cohenwang2024contextcite}.
The surrogate model then sheds light on the importance of different sources.
Surrogate models are generally \emph{example-specific}: they approximate model behavior within the narrow setting of a particular example.
Our work seeks to learn a surrogate model that transfers \emph{across examples}.

\paragraph{Efficient attribution and explanation.}

By learning attention head coefficients once and then performing attribution across examples, we are \emph{amortizing} the cost of attribution.
While the cost of training \ours is high, the cost of attributing an unseen example is low.
Prior work has similarly reduced the cost of explanation by training an explainer model (similar to our surrogate model) to mimic the behavior of an existing expensive explanation method~\citep{schwarzenberg2021efficient,situ2021learning}.
For example, \citet{jethani2021fastshap} approximate Shapley values~\citep{shapley1953value} in this manner, which normally require several ablations to approximate well.
\citet{covert2024stochastic} propose a general framework for amortizing the cost of attribution by learning from noisy attributions.
While these methods use a separate learned neural network over the input to attribute or explain model behavior, we use just a linear model over attention weights.

\paragraph{Using attention to explain model behavior.}
Visualizing attention weights is a common strategy for interpreting model behavior \citep{lee2017interactive,ding2017visualizing,abnar2020quantifying,vig2020causal}.
However, prior work has cast doubt on the reliability of attention weights as explanations \citep{jain2019attention,wiegreffe2019attention,serrano2019attention}.
For example,~\citet{jain2019attention} find that attention weights are not well-correlated with other measures of importance such as gradients and can frequently be manipulated without substantially changing a model's output.
\citet{serrano2019attention} observe that attention weights are predictive of the effect of zeroing out the weight, but that gradients are better predictors of this effect.

In this work, we are interested in whether attention weights across layers and heads can be used to predict the effect of ablating a source.
For transformer models, a typical approach is to average attention weights across layers and heads~\citep{kim2019and,sarti2023quantifying}.
In a context attribution setting (one of the token attribution settings we consider), this strategy has been found to be effective for attributing some models but yields very inaccurate attributions for other models~\citep{cohenwang2024contextcite}.
Our work is motivated by the observation that specific attention heads have been found to perform specific tasks \citep{wu2024retrieval,zheng2024attention,chuang2024lookback}.
We seek to mitigate the reliability shortcomings of using attention weights by \emph{learning} the extent to which different attention heads are useful for attribution.
Our observations suggest that, through this approach, the signal from attention weights \emph{can} faithfully attribute model behavior (at least with respect to representing the effects of ablating sources).

\section{Conclusion}
\label{sec:conclusion}
We propose a token attribution method, Attribution with Attention (\ours), which models the effects of source ablations by treating the attention weights of different attention heads as features.
By learning to combine these weights, \ours attains comparable attribution quality to previous methods that require significantly more computation.
\ours can be applied to a variety of tasks, including context and thought attribution.

\section{Acknowledgements}
\label{sec:acknowledgements}
The authors would like to thank Harshay Shah and Eric Liu for helpful feedback and discussions.
Work supported in part by Open Philanthropy.

\clearpage
\printbibliography
\clearpage

\appendix
\appendixpage
\startcontents[sections]
\printcontents[sections]{l}{1}{\setcounter{tocdepth}{2}}
\pagebreak

\section{Experiment details}

\label{sec:appendix_details}
\subsection{General}
\label{sec:implementation_details}

We run all experiments on a cluster of A100 GPUs.
When we split a piece of text into sentences (e.g., when we split a context into individual sentences to consider as sources), we first split on newlines and then use the off-the-shelf sentence tokenizer from the \texttt{nltk} library~\citep{bird2009natural}.
Error bars throughout the paper represent standard errors.

\subsection{Models}
\label{sec:model_details}
The language models we consider in this work are \texttt{Llama-3.1-8B} \citep{dubey2024llama}, \texttt{Phi-3.5-mini} \citep{abdin2024phi} and \texttt{DeepSeek-R1-Qwen-7B}~\citep{yang2024qwen2,guo2025deepseek}.
We use instruction-tuned variants of these models (where applicable).
We use the implementations of language models from HuggingFace's \texttt{transformers} library \citep{wolf2020transformers}.
Specifically, we use the following models:
\begin{itemize}
    \item \texttt{Llama-3.1-8B}: \texttt{meta-llama/Meta-Llama-3.1-8B-Instruct} with \texttt{float16} format
    \item \texttt{Phi-3.5-mini}: \texttt{microsoft/Phi-3.5-mini-instruct} with \texttt{float16} format
    \item \texttt{DeepSeek-R1-Qwen-7B}: \texttt{deepseek-ai/DeepSeek-R1-Distill-Qwen-7B} with \texttt{bfloat16} format
\end{itemize}
When generating responses with these models, we use their standard chat templates.
For the thinking model \texttt{DeepSeek-R1-Qwen-7B}, we force the model to produce thoughts before generating its response by starting its response with \texttt{<think>}.

\subsection{Datasets}
\label{sec:dataset_details}

We consider a variety of datasets to train and evaluate \ours.
In this section, we provide details about these datasets and the prompts we use for each.
When generating responses, we use greedy decoding (i.e., we always pick the token with the highest probability) rather than sampling from the model's distribution.
We sample up to $512$ tokens for context attribution tasks and $4,096$ tokens for thought attribution tasks (because in this case, the model's generation includes both the intermediate thoughts and the final response).
If the language model does not produce a final response within this limit (i.e., if it is still generating thoughts), we exclude the example from training and from our evaluations.

\subsubsection{Context attribution datasets}

\paragraph{CNN DailyMail.} CNN DailyMail~\citep{nallapati2016abstractive} is a news summarization dataset.
The contexts consists of a news article and the query asks the language model to briefly summarize the articles in up to three sentences.
We use the following prompt template:
\begin{prompt}
Context: {article}

Query: Please summarize the article in up to three sentences.
\end{prompt}

\paragraph{Hotpot QA.} Hotpot QA~\citep{yang2018hotpotqa} is a \emph{multi-hop} question-answering dataset in which the context consists of multiple short documents.
Answering the question requires combining information from a subset of these documents---the rest are ``distractors'' containing information that is only seemingly relevant.
We use the following prompt template:
\begin{prompt}
Passages:

{passage_1}

{passage_2}
...
{passage_n}

Query: {query}
\end{prompt}

\paragraph{MS MARCO.} MS MARCO~\citep{nguyen2016ms} is a question-answering dataset in which the question is a Bing search query and the context is a passage from a retrieved web page that can be used to answer the question.
We use the following prompt template:
\begin{prompt}
Passages:

{passage_1}

{passage_2}
...
{passage_n}

Query: {query}
\end{prompt}

\paragraph{Natural Questions.} Natural Questions~\citep{kwiatkowski2019natural} is a question-answering dataset in which the questions are Google search queries and the context is a Wikipedia article.
The context is provided as raw HTML;
we include only paragraphs (text within \texttt{<p>} tags) and headers and provide these as context joined by newlines.
We filter the dataset to include only examples where the question can be answered just using the article.
We also only include examples where the context is at most 20,000 characters.
We use the following prompt template:
\begin{prompt}
Context: {context}

Query: {query}
\end{prompt}

\paragraph{Dolly 15K.} We use Dolly 15K ~\citep{DatabricksBlog2023DollyV2} as a generic context attribution dataset for training \ours (to evaluate generalization to the above datasets).
Dolly 15K is an instruction fine-tuning dataset that includes prompts with a context and query.
We use only the examples that include a context, which fall into the following categories: summarization, information extraction, and question answering.
We further filter the dataset to include only examples where the prompt is at most 10,000 characters.
We use the following prompt template:
\begin{prompt}
Context: {context}

Query: {query}
\end{prompt}

\subsubsection{Thought attribution datasets}

\paragraph{DROP.} Discrete Reasoning over Paragraphs (DROP)~\citep{dua2019drop} is a context-based question-answering dataset focused on reasoning.
For example, it might include a paragraph about a football game and ask about the total number of points scored by a particular player.
We use the following prompt template:
\begin{prompt}
Context: {context}

Query: {query}
\end{prompt}

\paragraph{Global Opinions QA.} Global Opinions QA~\citep{durmus2023towards} is a dataset of opinion questions on global issues.
For example, it asks about the extent to which gun ownership should be restricted.
The dataset is crowdsourced from a variety of sources and includes a variety of topics and opinions.
Although this dataset does not directly involve reasoning, we find it to be suitable for thought attribution because the language model extensively considers different information and plans how to respond before responding.
We present each question to the language model directly as a prompt.
This dataset only includes one split, so we randomly select $2,000$ examples for training from this split and use the rest for evaluation.

\paragraph{AGIEval.} AGIEval~\citep{zhong2023agieval} is a dataset of multiple-choice reasoning problems spanning english, law, logic and math.
We use this dataset as a generic dataset for training \ours for thought attribution.
When the example includes a passage, we use the following prompt template:
\begin{prompt}
Passage: {context}

Question: {query}
{option_1}
{option_2}
...
\end{prompt}
When the example does not include a passage, we use the following prompt template:
\begin{prompt}
{query}
{option_1}
{option_2}
...
\end{prompt}

\subsection{Example-specific surrogate modeling}
\label{sec:regression_details}

Our implementation of example-specific surrogate modeling (see~\Cref{alg:regression}) follows \cc, which applies the surrogate modeling technique to the context attribution setting~\citep{cohenwang2024contextcite}.
We summarize this approach in~\Cref{alg:regression}.
We highlight the following design choices:
\begin{enumerate}
    \item Instead of the surrogate model $\hat{f}_w$ predicting probabilities directly, we predict \emph{logit-scaled} probabilities.
    The logit transform (denoted by $\sigma^{-1}$) is the inverse of the sigmoid function and maps probabilities in $(0,1)$ to real values in $(-\infty,\infty)$.
    We apply this transform because logit-scaled probabilities are a more natural target for a $\hat{f}_w$, a linear model, than probabilities.

    \item To optimize the surrogate model, we use \lasso~\citep{tibshirani1994regression}.
    \lasso minimizes the MSE loss with a regularization term to encourage sparsity.
    This is effective because a language model often only relies on a small subset of sources when generating a particular sequence (as illustrated empirically in a context attribution setting by~\citet{cohenwang2024contextcite}).
    With \lasso, we can learn an effective surrogate model from a number of ablations smaller than the number of sources.
    We use the implementation of \lasso from \texttt{sklearn}~\citep{pedregosa2011scikit} and set the regularization parameter $\lambda$ to $0.01$.

    \item We sample ablation vectors uniformly from $\{0,1\}^{|S|}$.
\end{enumerate}

We apply this approach to the problem of token attribution in~\Cref{alg:regression} (adapted from \cc~\citep{cohenwang2024contextcite}).

\begin{algorithm}[H]
\caption{Example-specific surrogate modeling (ESM)}
\label{alg:regression}
\begin{algorithmic}[1]
    \State {\bfseries Input:} Autoregressive language model $p_\text{LM}$, sequence of input tokens $X$, set of sources $S$, sequence of generated tokens $Y$, number of ablations $n$, and regularization parameter $\lambda$
    \State {\bfseries Output:} Attribution scores $\hat{w} \in \mathbb{R}^{|S|}$
    \State $f(v) \coloneqq p_\text{LM}(Y\mid\ablate(X,S,v))$
    \State $g(v) \coloneqq \sigma^{-1}(f(v))$
    \For{$i \in \{1,\ldots,t\}$}
        \State Sample an ablation $v^{(i)}$ from $\{0,1\}^{|S|}$
        \State Compute $g(v^{(i)})$
    \EndFor
    \State $\hat{w},\hat{b}\gets\lasso(\{(v^{(i)},g(v^{(i)}))\}_{i=1}^n,\lambda)$
    \State \textbf{return} $\hat{w}$
\end{algorithmic}
\end{algorithm}

\subsection{\ours}
\label{sec:ours_details}

For each of the training examples, we consider each of the sentences in the model's generation as a potential attribution target $Y$.
We initialize the attention head coefficients $\theta$ to be uniform (matching the average attention baseline) and train for $1,000$ steps.
At each step, we sample a batch of $512$ examples and select a random sentence from the model's generation to consider as the attribution target $Y$.
We use the Adam optimizer~\citep{kingma2015adam} with a learning rate of $0.001$ and a cosine learning rate schedule.
To perform ablations, we set attention scores to zero for ablated tokens.

\subsection{Applying \ours to improve response quality}
\label{sec:pruning_details}

In \Cref{sec:applications}, we show that if we use \ours to prune a context and include only the highest-scoring sources, we can improve the quality of the model's response on Hotpot QA.
For this experiment, we use the version of \ours trained on the Hotpot QA dataset and consider entire passages as the sources.
To be able to evaluate the correctness of the model's answer, we modify the prompt template such that the model produces a short answer that can be matched against the ground truth.
We use the following modified prompt template:
\begin{prompt}
Passages:

{passage_1}

{passage_2}
...
{passage_n}

Query: {query}

Please respond in complete sentences and then end with just a single word or phrase in the format "Final answer: <answer>". If the question is a yes or no question, the answer should be "yes" or "no".
\end{prompt}
To evaluate correctness, we compare just the answer provided after \texttt{Final answer:} to the ground truth.
If the model does not produce an answer after \texttt{Final answer:}, we consider the answer to be the empty string.

\section{Additional results}

\label{sec:appendix_results}
\subsection{Visualization of learned coefficients}
\label{sec:coefficients}
\begin{figure*}[t]
    \centering
    \includegraphics[width=\textwidth]{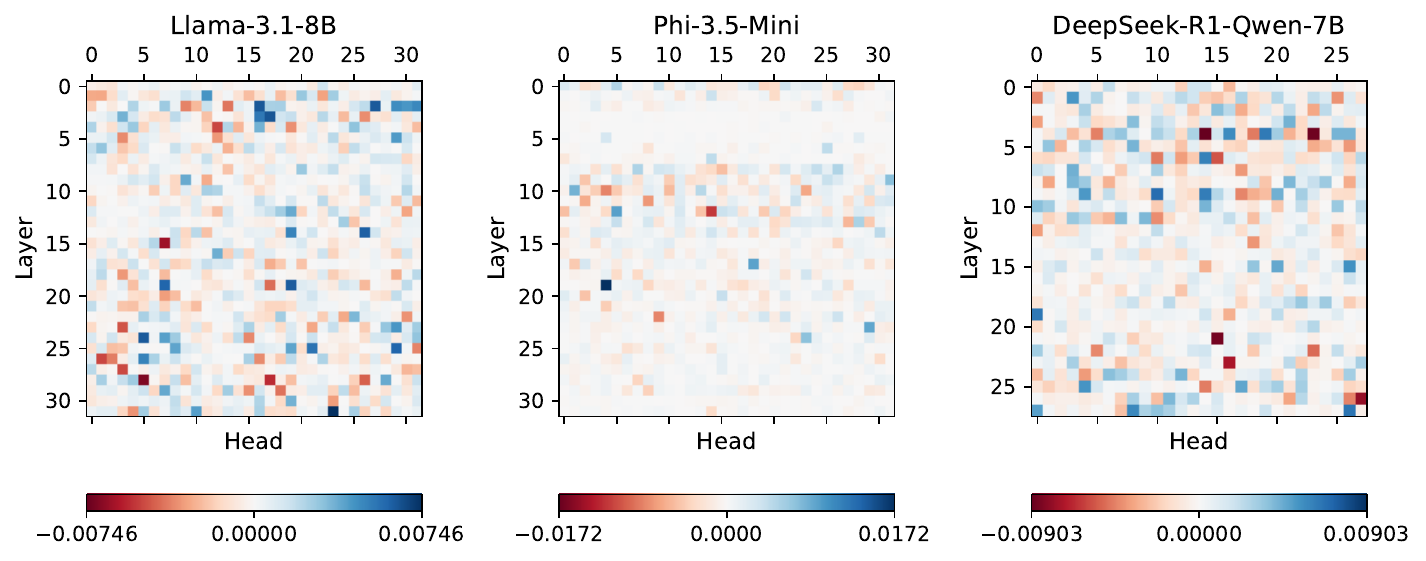}
    \caption{
        \textbf{Visualizing learned attention head coefficients.}
        We visualize the learned coefficients of \ours for different models trained on their corresponding generic datasets (Dolly 15K for \texttt{Llama-3.1-8B} and \texttt{Phi-3.5-mini}, and AGIEval for \texttt{DeepSeek-R1-Qwen-7B}).
        While the coefficients for \texttt{Phi-3.5-mini} seems to have the highest magnitude for middle layers, the high-magnitude coefficients for \texttt{Llama-3.1-8B} and \texttt{DeepSeek-R1-Qwen-7B} are more uniformly distributed.
    }
    \label{fig:coefficients}
\end{figure*}

In~\Cref{fig:coefficients}, we visualize the learned coefficients of different models trained on their corresponding generic datasets.
We do not observe a consistent pattern for high-magnitude coefficients---for \texttt{Llama-3.1-8B} and \texttt{DeepSeek-R1-Qwen-7B}, the high-magnitude coefficients are more uniformly distributed across layers, while for \texttt{Phi-3.5-mini}, the high-magnitude coefficients are well-concentrated in the middle layers.

\subsection{Examples of attributions}
\label{sec:attribution_examples}
To provide a more extensive qualitative evaluation of different attribution methods, we visualize the attribution scores of random examples from different datasets for different models and different sources types.
To visualize attributions, we highlight sources according to their attribution scores (normalizing by the maximum attribution score for each example).
We only show sources for which the attribution score is above $0.2$ of the maximum attribution score, or also surrounding sources in the case of token-level sources.
We show context attribution examples in~\Cref{fig:cnn_dailymail_llama_3_1_8b_sentence,fig:cnn_dailymail_phi_3_5_mini_sentence,fig:cnn_dailymail_llama_3_1_8b_token,fig:cnn_dailymail_phi_3_5_mini_token,fig:hotpot_qa_passages_llama_3_1_8b_sentence,fig:hotpot_qa_passages_phi_3_5_mini_sentence,fig:hotpot_qa_passages_llama_3_1_8b_token,fig:hotpot_qa_passages_phi_3_5_mini_token} and thought attribution examples in~\Cref{fig:global_opinions_r1_qwen_7b_dtype=bfloat16_token_0,fig:global_opinions_r1_qwen_7b_dtype=bfloat16_token_1,fig:global_opinions_r1_qwen_7b_dtype=bfloat16_sentence_0,fig:global_opinions_r1_qwen_7b_dtype=bfloat16_sentence_1}.

\begin{figure*}[t]
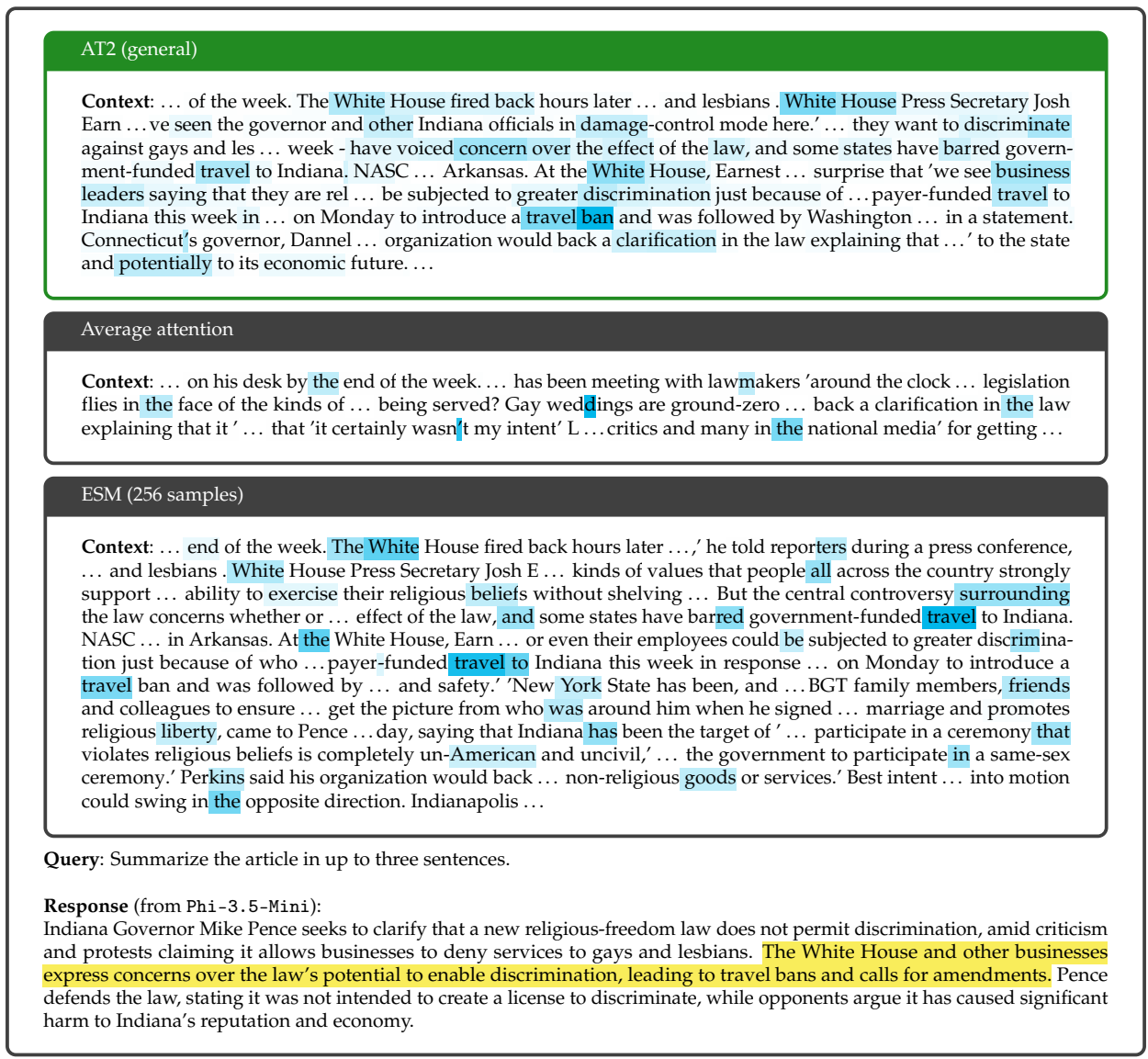

\centering\definecolor{forestgreen}{RGB}{34, 139, 34}
\relsize{-1}
\begin{tcolorbox}[colback=white]
\begin{tcolorbox}[colback=white, colframe=forestgreen, title=\ours (general)]
\textbf{Context}:
 \dots\myhl{253}{254}{254}{ of}\myhl{253}{254}{254}{ the}\myhl{254}{254}{254}{ week}\myhl{252}{254}{254}{.}\myhl{249}{253}{254}{ The}\myhl{178}{233}{248}{ White}\myhl{213}{243}{251}{ House}\myhl{219}{245}{252}{ fired}\myhl{217}{244}{252}{ back}\myhl{250}{253}{254}{ hours}\myhl{249}{253}{254}{ later} \dots\myhl{252}{254}{254}{ and}\myhl{254}{254}{254}{ les}\myhl{254}{254}{254}{bi}\myhl{254}{254}{254}{ans}\myhl{253}{254}{254}{ .}\myhl{121}{217}{244}{ White}\myhl{165}{230}{247}{ House}\myhl{233}{248}{253}{ Press}\myhl{232}{248}{253}{ Secretary}\myhl{250}{253}{254}{ Josh}\myhl{253}{254}{254}{ E}\myhl{253}{254}{254}{arn} \dots\myhl{251}{254}{254}{ve}\myhl{235}{249}{253}{ seen}\myhl{253}{254}{254}{ the}\myhl{249}{253}{254}{ governor}\myhl{250}{253}{254}{ and}\myhl{200}{239}{250}{ other}\myhl{253}{254}{254}{ Indiana}\myhl{249}{253}{254}{ officials}\myhl{252}{254}{254}{ in}\myhl{197}{238}{250}{ damage}\myhl{249}{253}{254}{-}\myhl{252}{254}{254}{control}\myhl{255}{255}{255}{ mode}\myhl{251}{254}{254}{ here}\myhl{250}{253}{254}{.'} \dots\myhl{249}{253}{254}{ they}\myhl{245}{252}{254}{ want}\myhl{239}{250}{253}{ to}\myhl{220}{245}{252}{ disc}\myhl{225}{246}{252}{rim}\myhl{169}{230}{248}{inate}\myhl{250}{253}{254}{ against}\myhl{251}{254}{254}{ g}\myhl{252}{254}{254}{ays}\myhl{253}{254}{254}{ and}\myhl{254}{254}{254}{ les} \dots\myhl{255}{255}{255}{ week}\myhl{248}{253}{254}{ -}\myhl{229}{247}{252}{ have}\myhl{225}{246}{252}{ vo}\myhl{216}{244}{251}{iced}\myhl{160}{228}{247}{ concern}\myhl{190}{236}{249}{ over}\myhl{227}{247}{252}{ the}\myhl{212}{242}{251}{ effect}\myhl{247}{252}{254}{ of}\myhl{251}{253}{254}{ the}\myhl{212}{243}{251}{ law}\myhl{229}{247}{252}{,}\myhl{242}{251}{254}{ and}\myhl{245}{252}{254}{ some}\myhl{229}{247}{252}{ states}\myhl{248}{253}{254}{ have}\myhl{203}{240}{250}{ bar}\myhl{231}{248}{253}{red}\myhl{246}{252}{254}{ government}\myhl{253}{254}{254}{-}\myhl{240}{250}{253}{fund}\myhl{255}{255}{255}{ed}\myhl{173}{232}{248}{ travel}\myhl{238}{250}{253}{ to}\myhl{255}{255}{255}{ Indiana}\myhl{213}{243}{251}{.}\myhl{235}{249}{253}{ NAS}\myhl{252}{254}{254}{C} \dots\myhl{254}{254}{254}{ Ark}\myhl{255}{255}{255}{ansas}\myhl{255}{255}{255}{.}\myhl{252}{254}{254}{ At}\myhl{253}{254}{254}{ the}\myhl{143}{223}{246}{ White}\myhl{209}{242}{251}{ House}\myhl{232}{248}{253}{,}\myhl{248}{253}{254}{ E}\myhl{253}{254}{254}{arn}\myhl{251}{253}{254}{est} \dots\myhl{251}{254}{254}{ surprise}\myhl{252}{254}{254}{ that}\myhl{253}{254}{254}{ '}\myhl{251}{254}{254}{we}\myhl{250}{253}{254}{ see}\myhl{165}{230}{247}{ business}\myhl{203}{240}{250}{ leaders}\myhl{232}{248}{253}{ saying}\myhl{242}{251}{254}{ that}\myhl{250}{253}{254}{ they}\myhl{250}{253}{254}{ are}\myhl{247}{252}{254}{ rel} \dots\myhl{253}{254}{254}{ be}\myhl{251}{254}{254}{ subject}\myhl{248}{253}{254}{ed}\myhl{249}{253}{254}{ to}\myhl{231}{248}{253}{ greater}\myhl{189}{236}{249}{ disc}\myhl{216}{244}{251}{rim}\myhl{224}{246}{252}{ination}\myhl{243}{251}{254}{ just}\myhl{242}{251}{253}{ because}\myhl{249}{253}{254}{ of} \dots\myhl{255}{255}{255}{p}\myhl{255}{255}{255}{ayer}\myhl{255}{255}{255}{-}\myhl{253}{254}{254}{fund}\myhl{255}{255}{255}{ed}\myhl{182}{234}{249}{ travel}\myhl{242}{251}{254}{ to}\myhl{254}{254}{254}{ Indiana}\myhl{243}{251}{254}{ this}\myhl{246}{252}{254}{ week}\myhl{232}{248}{253}{ in} \dots\myhl{252}{254}{254}{ on}\myhl{251}{253}{254}{ Monday}\myhl{250}{253}{254}{ to}\myhl{246}{252}{254}{ introduce}\myhl{251}{254}{254}{ a}\myhl{124}{218}{244}{ travel}\myhl{0}{183}{234}{ ban}\myhl{235}{249}{253}{ and}\myhl{253}{254}{254}{ was}\myhl{250}{253}{254}{ followed}\myhl{251}{254}{254}{ by}\myhl{250}{253}{254}{ Washington} \dots\myhl{253}{254}{254}{ in}\myhl{253}{254}{254}{ a}\myhl{250}{253}{254}{ statement}\myhl{245}{252}{254}{.}\myhl{244}{252}{254}{ Connecticut}\myhl{176}{233}{248}{'}\myhl{254}{254}{254}{s}\myhl{252}{254}{254}{ governor}\myhl{254}{254}{254}{,}\myhl{253}{254}{254}{ D}\myhl{253}{254}{254}{annel} \dots\myhl{251}{253}{254}{ organization}\myhl{251}{253}{254}{ would}\myhl{246}{252}{254}{ back}\myhl{245}{252}{254}{ a}\myhl{207}{241}{251}{ clar}\myhl{202}{240}{250}{ification}\myhl{249}{253}{254}{ in}\myhl{255}{255}{255}{ the}\myhl{252}{254}{254}{ law}\myhl{246}{252}{254}{ explaining}\myhl{250}{253}{254}{ that} \dots\myhl{255}{255}{255}{'}\myhl{255}{255}{255}{ to}\myhl{254}{254}{254}{ the}\myhl{255}{255}{255}{ state}\myhl{251}{254}{254}{ and}\myhl{189}{236}{249}{ potentially}\myhl{244}{251}{254}{ to}\myhl{254}{254}{254}{ its}\myhl{237}{250}{253}{ economic}\myhl{253}{254}{254}{ future}\myhl{251}{254}{254}{.} \dots
\end{tcolorbox}
\begin{tcolorbox}[colback=white, , title=Average attention]
\textbf{Context}:
 \dots\myhl{254}{254}{254}{ on}\myhl{254}{254}{254}{ his}\myhl{254}{254}{254}{ des}\myhl{254}{254}{254}{k}\myhl{254}{254}{254}{ by}\myhl{191}{237}{249}{ the}\myhl{254}{254}{254}{ end}\myhl{254}{254}{254}{ of}\myhl{254}{254}{254}{ the}\myhl{254}{254}{254}{ week}\myhl{253}{254}{254}{.} \dots\myhl{254}{254}{254}{ has}\myhl{254}{254}{254}{ been}\myhl{254}{254}{254}{ meeting}\myhl{254}{254}{254}{ with}\myhl{254}{254}{254}{ law}\myhl{190}{236}{249}{m}\myhl{254}{254}{254}{akers}\myhl{254}{254}{254}{ '}\myhl{254}{254}{254}{around}\myhl{254}{254}{254}{ the}\myhl{254}{254}{254}{ clock} \dots\myhl{254}{254}{254}{ legisl}\myhl{254}{254}{254}{ation}\myhl{254}{254}{254}{ f}\myhl{254}{254}{254}{lies}\myhl{254}{254}{254}{ in}\myhl{186}{235}{249}{ the}\myhl{254}{254}{254}{ face}\myhl{254}{254}{254}{ of}\myhl{254}{254}{254}{ the}\myhl{254}{254}{254}{ kinds}\myhl{254}{254}{254}{ of} \dots\myhl{254}{254}{254}{ being}\myhl{254}{254}{254}{ served}\myhl{254}{254}{254}{?}\myhl{254}{254}{254}{ Gay}\myhl{254}{254}{254}{ wed}\myhl{0}{183}{234}{d}\myhl{254}{254}{254}{ings}\myhl{254}{254}{254}{ are}\myhl{254}{254}{254}{ ground}\myhl{254}{254}{254}{-}\myhl{254}{254}{254}{zero} \dots\myhl{254}{254}{254}{ back}\myhl{254}{254}{254}{ a}\myhl{253}{254}{254}{ clar}\myhl{253}{254}{254}{ification}\myhl{254}{254}{254}{ in}\myhl{173}{232}{248}{ the}\myhl{253}{254}{254}{ law}\myhl{254}{254}{254}{ explaining}\myhl{254}{254}{254}{ that}\myhl{254}{254}{254}{ it}\myhl{254}{254}{254}{ '} \dots\myhl{254}{254}{254}{ that}\myhl{254}{254}{254}{ '}\myhl{254}{254}{254}{it}\myhl{254}{254}{254}{ certainly}\myhl{254}{254}{254}{ wasn}\myhl{36}{193}{237}{'}\myhl{254}{254}{254}{t}\myhl{254}{254}{254}{ my}\myhl{254}{254}{254}{ intent}\myhl{254}{254}{254}{'}\myhl{254}{254}{254}{ L} \dots\myhl{254}{254}{254}{crit}\myhl{254}{254}{254}{ics}\myhl{254}{254}{254}{ and}\myhl{254}{254}{254}{ many}\myhl{254}{254}{254}{ in}\myhl{119}{217}{244}{ the}\myhl{254}{254}{254}{ national}\myhl{253}{254}{254}{ media}\myhl{254}{254}{254}{'}\myhl{254}{254}{254}{ for}\myhl{254}{254}{254}{ getting} \dots
\end{tcolorbox}
\begin{tcolorbox}[colback=white, , title=ESM (256 samples)]
\textbf{Context}:
 \dots\myhl{235}{249}{253}{ end}\myhl{255}{255}{255}{ of}\myhl{255}{255}{255}{ the}\myhl{255}{255}{255}{ week}\myhl{255}{255}{255}{.}\myhl{162}{228}{247}{ The}\myhl{92}{209}{241}{ White}\myhl{255}{255}{255}{ House}\myhl{255}{255}{255}{ fired}\myhl{255}{255}{255}{ back}\myhl{255}{255}{255}{ hours}\myhl{255}{255}{255}{ later} \dots\myhl{255}{255}{255}{,'}\myhl{255}{255}{255}{ he}\myhl{254}{254}{254}{ told}\myhl{255}{255}{255}{ rep}\myhl{255}{255}{255}{or}\myhl{187}{236}{249}{ters}\myhl{255}{255}{255}{ during}\myhl{255}{255}{255}{ a}\myhl{255}{255}{255}{ press}\myhl{255}{255}{255}{ conference}\myhl{255}{255}{255}{,} \dots\myhl{255}{255}{255}{ and}\myhl{255}{255}{255}{ les}\myhl{255}{255}{255}{bi}\myhl{255}{255}{255}{ans}\myhl{255}{255}{255}{ .}\myhl{176}{233}{248}{ White}\myhl{255}{255}{255}{ House}\myhl{255}{255}{255}{ Press}\myhl{255}{255}{255}{ Secretary}\myhl{255}{255}{255}{ Josh}\myhl{255}{255}{255}{ E} \dots\myhl{255}{255}{255}{ kinds}\myhl{255}{255}{255}{ of}\myhl{255}{255}{255}{ values}\myhl{255}{255}{255}{ that}\myhl{255}{255}{255}{ people}\myhl{151}{225}{246}{ all}\myhl{255}{255}{255}{ across}\myhl{255}{255}{255}{ the}\myhl{255}{255}{255}{ country}\myhl{255}{255}{255}{ strongly}\myhl{255}{255}{255}{ support} \dots\myhl{255}{255}{255}{ ability}\myhl{255}{255}{255}{ to}\myhl{227}{247}{252}{ exercise}\myhl{255}{255}{255}{ their}\myhl{255}{255}{255}{ religious}\myhl{192}{237}{249}{ belief}\myhl{255}{255}{255}{s}\myhl{255}{255}{255}{ without}\myhl{255}{255}{255}{ sh}\myhl{255}{255}{255}{el}\myhl{255}{255}{255}{ving} \dots\myhl{255}{255}{255}{ But}\myhl{255}{255}{255}{ the}\myhl{255}{255}{255}{ central}\myhl{255}{255}{255}{ controvers}\myhl{255}{255}{255}{y}\myhl{152}{226}{246}{ surrounding}\myhl{255}{255}{255}{ the}\myhl{255}{255}{255}{ law}\myhl{255}{255}{255}{ concerns}\myhl{255}{255}{255}{ whether}\myhl{255}{255}{255}{ or} \dots\myhl{255}{255}{255}{ effect}\myhl{255}{255}{255}{ of}\myhl{255}{255}{255}{ the}\myhl{255}{255}{255}{ law}\myhl{255}{255}{255}{,}\myhl{156}{227}{247}{ and}\myhl{255}{255}{255}{ some}\myhl{255}{255}{255}{ states}\myhl{255}{255}{255}{ have}\myhl{255}{255}{255}{ bar}\myhl{140}{223}{245}{red}\myhl{255}{255}{255}{ government}\myhl{255}{255}{255}{-}\myhl{255}{255}{255}{fund}\myhl{255}{255}{255}{ed}\myhl{0}{183}{234}{ travel}\myhl{255}{255}{255}{ to}\myhl{255}{255}{255}{ Indiana}\myhl{255}{255}{255}{.}\myhl{255}{255}{255}{ NAS}\myhl{255}{255}{255}{C} \dots\myhl{255}{255}{255}{ in}\myhl{255}{255}{255}{ Ark}\myhl{253}{254}{254}{ansas}\myhl{255}{255}{255}{.}\myhl{255}{255}{255}{ At}\myhl{92}{209}{242}{ the}\myhl{255}{255}{255}{ White}\myhl{255}{255}{255}{ House}\myhl{255}{255}{255}{,}\myhl{255}{255}{255}{ E}\myhl{255}{255}{255}{arn} \dots\myhl{255}{255}{255}{ or}\myhl{255}{255}{255}{ even}\myhl{255}{255}{255}{ their}\myhl{255}{255}{255}{ employees}\myhl{255}{255}{255}{ could}\myhl{197}{238}{250}{ be}\myhl{255}{255}{255}{ subject}\myhl{255}{255}{255}{ed}\myhl{255}{255}{255}{ to}\myhl{255}{255}{255}{ greater}\myhl{255}{255}{255}{ disc}\myhl{165}{229}{247}{rim}\myhl{255}{255}{255}{ination}\myhl{255}{255}{255}{ just}\myhl{255}{255}{255}{ because}\myhl{255}{255}{255}{ of}\myhl{255}{255}{255}{ who} \dots\myhl{255}{255}{255}{p}\myhl{255}{255}{255}{ayer}\myhl{220}{245}{252}{-}\myhl{255}{255}{255}{fund}\myhl{255}{255}{255}{ed}\myhl{23}{190}{236}{ travel}\myhl{100}{211}{242}{ to}\myhl{255}{255}{255}{ Indiana}\myhl{255}{255}{255}{ this}\myhl{255}{255}{255}{ week}\myhl{255}{255}{255}{ in}\myhl{255}{255}{255}{ response} \dots\myhl{255}{255}{255}{ on}\myhl{255}{255}{255}{ Monday}\myhl{255}{255}{255}{ to}\myhl{255}{255}{255}{ introduce}\myhl{255}{255}{255}{ a}\myhl{122}{217}{244}{ travel}\myhl{255}{255}{255}{ ban}\myhl{255}{255}{255}{ and}\myhl{255}{255}{255}{ was}\myhl{255}{255}{255}{ followed}\myhl{255}{255}{255}{ by} \dots\myhl{255}{255}{255}{ and}\myhl{255}{255}{255}{ safety}\myhl{255}{255}{255}{.'}\myhl{255}{255}{255}{ '}\myhl{255}{255}{255}{New}\myhl{196}{238}{250}{ York}\myhl{255}{255}{255}{ State}\myhl{255}{255}{255}{ has}\myhl{255}{255}{255}{ been}\myhl{255}{255}{255}{,}\myhl{255}{255}{255}{ and} \dots\myhl{255}{255}{255}{B}\myhl{255}{255}{255}{GT}\myhl{255}{255}{255}{ family}\myhl{255}{255}{255}{ members}\myhl{255}{255}{255}{,}\myhl{179}{233}{248}{ friends}\myhl{255}{255}{255}{ and}\myhl{255}{255}{255}{ colle}\myhl{255}{255}{255}{agues}\myhl{255}{255}{255}{ to}\myhl{255}{255}{255}{ ensure} \dots\myhl{255}{255}{255}{ get}\myhl{255}{255}{255}{ the}\myhl{255}{255}{255}{ picture}\myhl{255}{255}{255}{ from}\myhl{255}{255}{255}{ who}\myhl{185}{235}{249}{ was}\myhl{255}{255}{255}{ around}\myhl{255}{255}{255}{ him}\myhl{255}{255}{255}{ when}\myhl{255}{255}{255}{ he}\myhl{255}{255}{255}{ signed} \dots\myhl{255}{255}{255}{ marriage}\myhl{255}{255}{255}{ and}\myhl{249}{253}{254}{ prom}\myhl{255}{255}{255}{otes}\myhl{255}{255}{255}{ religious}\myhl{198}{239}{250}{ liberty}\myhl{255}{255}{255}{,}\myhl{255}{255}{255}{ came}\myhl{255}{255}{255}{ to}\myhl{255}{255}{255}{ P}\myhl{255}{255}{255}{ence} \dots\myhl{255}{255}{255}{day}\myhl{255}{255}{255}{,}\myhl{255}{255}{255}{ saying}\myhl{255}{255}{255}{ that}\myhl{255}{255}{255}{ Indiana}\myhl{148}{225}{246}{ has}\myhl{255}{255}{255}{ been}\myhl{255}{255}{255}{ the}\myhl{255}{255}{255}{ target}\myhl{255}{255}{255}{ of}\myhl{255}{255}{255}{ '} \dots\myhl{255}{255}{255}{ particip}\myhl{255}{255}{255}{ate}\myhl{255}{255}{255}{ in}\myhl{255}{255}{255}{ a}\myhl{255}{255}{255}{ ceremony}\myhl{161}{228}{247}{ that}\myhl{255}{255}{255}{ viol}\myhl{255}{255}{255}{ates}\myhl{255}{255}{255}{ religious}\myhl{255}{255}{255}{ belief}\myhl{255}{255}{255}{s}\myhl{255}{255}{255}{ is}\myhl{255}{255}{255}{ completely}\myhl{255}{255}{255}{ un}\myhl{255}{255}{255}{-}\myhl{203}{240}{250}{American}\myhl{255}{255}{255}{ and}\myhl{255}{255}{255}{ un}\myhl{255}{255}{255}{c}\myhl{255}{255}{255}{ivil}\myhl{255}{255}{255}{,'} \dots\myhl{255}{255}{255}{ the}\myhl{255}{255}{255}{ government}\myhl{255}{255}{255}{ to}\myhl{255}{255}{255}{ particip}\myhl{255}{255}{255}{ate}\myhl{159}{228}{247}{ in}\myhl{255}{255}{255}{ a}\myhl{255}{255}{255}{ same}\myhl{255}{255}{255}{-}\myhl{255}{255}{255}{sex}\myhl{255}{255}{255}{ ceremony}\myhl{255}{255}{255}{.'}\myhl{255}{255}{255}{ Per}\myhl{190}{236}{249}{kins}\myhl{255}{255}{255}{ said}\myhl{255}{255}{255}{ his}\myhl{255}{255}{255}{ organization}\myhl{255}{255}{255}{ would}\myhl{255}{255}{255}{ back} \dots\myhl{255}{255}{255}{ non}\myhl{255}{255}{255}{-}\myhl{255}{255}{255}{rel}\myhl{255}{255}{255}{ig}\myhl{255}{255}{255}{ious}\myhl{191}{237}{249}{ goods}\myhl{255}{255}{255}{ or}\myhl{255}{255}{255}{ services}\myhl{255}{255}{255}{.'}\myhl{255}{255}{255}{ Best}\myhl{255}{255}{255}{ intent} \dots\myhl{255}{255}{255}{ into}\myhl{255}{255}{255}{ motion}\myhl{255}{255}{255}{ could}\myhl{255}{255}{255}{ swing}\myhl{255}{255}{255}{ in}\myhl{115}{215}{243}{ the}\myhl{255}{255}{255}{ opposite}\myhl{255}{255}{255}{ direction}\myhl{255}{255}{255}{.}\myhl{255}{255}{255}{ Indian}\myhl{255}{255}{255}{apolis} \dots
\end{tcolorbox}
\textbf{Query}:
Summarize the article in up to three sentences.
\newline\newline\textbf{Response} (from \texttt{Phi-3.5-Mini}):
\newline
Indiana Governor Mike Pence seeks to clarify that a new religious-freedom law does not permit discrimination, amid criticism and protests claiming it allows businesses to deny services to gays and lesbians. \myhl{249}{237}{91}{The White House and other businesses express concerns over the law's potential to enable discrimination, leading to travel bans and calls for amendments.} Pence defends the law, stating it was not intended to create a license to discriminate, while opponents argue it has caused significant harm to Indiana's reputation and economy.
\end{tcolorbox}
\relsize{1}
\caption{Visualized attributions for a random example from CNN DailyMail for \texttt{Phi-3.5-Mini} with tokens as sources.}
\label{fig:cnn_dailymail_phi_3_5_mini_token}
\end{figure*}

\begin{figure*}[t]
\centering\definecolor{forestgreen}{RGB}{34, 139, 34}
\relsize{-1}
\begin{tcolorbox}[colback=white]
\begin{tcolorbox}[colback=white, colframe=forestgreen, title=\ours (general)]
\textbf{Context}:
 \dots\myhl{252}{254}{254}{ online}\myhl{244}{252}{254}{ gambling}\myhl{217}{244}{252}{ firm}\myhl{222}{245}{252}{ linked}\myhl{255}{255}{255}{ to}\myhl{115}{215}{243}{ Harry}\myhl{196}{238}{250}{ Red}\myhl{250}{253}{254}{kn}\myhl{215}{243}{251}{app}\myhl{215}{243}{251}{ and}\myhl{239}{250}{253}{ two}\myhl{255}{255}{255}{ Premier} \dots\myhl{255}{255}{255}{ online}\myhl{255}{255}{255}{ gambling}\myhl{255}{255}{255}{ firm}\myhl{253}{254}{254}{,}\myhl{204}{240}{250}{ which}\myhl{54}{198}{238}{ signed}\myhl{124}{218}{244}{ Harry}\myhl{207}{241}{251}{ Red}\myhl{238}{250}{253}{kn}\myhl{210}{242}{251}{app}\myhl{140}{222}{245}{ to}\myhl{25}{190}{236}{ star}\myhl{216}{244}{251}{ in}\myhl{210}{242}{251}{ its}\myhl{0}{183}{234}{ adverts}\myhl{210}{242}{251}{ (}\myhl{243}{251}{254}{pictured}\myhl{255}{255}{255}{ above}\myhl{244}{251}{254}{)}\myhl{238}{250}{253}{ has} \dots\myhl{255}{255}{255}{ fraud}\myhl{255}{255}{255}{ .}\myhl{234}{249}{253}{ The}\myhl{255}{255}{255}{ online}\myhl{226}{247}{252}{ firm}\myhl{165}{230}{247}{ announced}\myhl{38}{194}{237}{ deals}\myhl{217}{244}{251}{ with}\myhl{116}{216}{243}{ West}\myhl{102}{212}{242}{ Brom}\myhl{82}{206}{241}{wich}\myhl{144}{224}{246}{ Alb}\myhl{185}{235}{249}{ion}\myhl{205}{241}{251}{ and}\myhl{52}{198}{238}{ Leicester}\myhl{197}{238}{250}{ City}\myhl{180}{234}{249}{ last}\myhl{220}{245}{252}{ summer}\myhl{178}{233}{248}{.}\myhl{251}{253}{254}{ It}\myhl{249}{253}{254}{ also}\myhl{208}{242}{251}{ signed}\myhl{255}{255}{255}{ former}\myhl{255}{255}{255}{ Q} \dots\myhl{229}{247}{252}{app}\myhl{225}{246}{252}{ to}\myhl{255}{255}{255}{ appear}\myhl{240}{250}{253}{ in}\myhl{220}{245}{252}{ its}\myhl{202}{240}{250}{ adverts}\myhl{242}{251}{254}{,}\myhl{234}{249}{253}{ alongside}\myhl{255}{255}{255}{ Lock}\myhl{255}{255}{255}{,}\myhl{253}{254}{254}{ Stock} \dots\myhl{255}{255}{255}{.}\myhl{231}{248}{253}{ Last}\myhl{255}{255}{255}{ week}\myhl{255}{255}{255}{,}\myhl{251}{254}{254}{ the}\myhl{51}{198}{238}{ UK}\myhl{168}{230}{248}{ gambling}\myhl{221}{245}{252}{ regulator}\myhl{229}{247}{252}{ The}\myhl{130}{220}{245}{ Gambling}\myhl{142}{223}{246}{ Commission}\myhl{27}{191}{236}{ suspended}\myhl{255}{255}{255}{ the}\myhl{222}{245}{252}{ firm}\myhl{255}{255}{255}{'s}\myhl{47}{196}{238}{ licence}\myhl{216}{244}{251}{.}\myhl{235}{249}{253}{ The}\myhl{186}{235}{249}{ suspension}\myhl{232}{248}{253}{ led}\myhl{252}{254}{254}{ to}\myhl{255}{255}{255}{ Football}\myhl{253}{254}{254}{ League}\myhl{247}{252}{254}{ One} \dots\myhl{244}{252}{254}{ firm}\myhl{255}{255}{255}{ }\myhl{255}{255}{255}{666}\myhl{255}{255}{255}{Bet}\myhl{218}{244}{252}{ announced}\myhl{194}{237}{250}{ deals}\myhl{251}{254}{254}{ with}\myhl{183}{235}{249}{ West}\myhl{221}{245}{252}{ Brom}\myhl{224}{246}{252}{ and}\myhl{171}{231}{248}{ Leicester}\myhl{228}{247}{252}{ City}\myhl{247}{252}{254}{ last}\myhl{250}{253}{254}{ summer}\myhl{250}{253}{254}{.}\myhl{255}{255}{255}{ It} \dots\myhl{255}{255}{255}{ be}\myhl{252}{254}{254}{.'}\myhl{250}{253}{254}{ Although}\myhl{255}{255}{255}{ the}\myhl{222}{245}{252}{ Gambling}\myhl{203}{240}{250}{ Commission}\myhl{252}{254}{254}{ refused}\myhl{253}{254}{254}{ to}\myhl{253}{254}{254}{ comment}\myhl{250}{253}{254}{,}\myhl{253}{254}{254}{ it}\myhl{249}{253}{254}{ previously}\myhl{250}{253}{254}{ said}\myhl{249}{253}{254}{ the}\myhl{192}{237}{250}{ suspension}\myhl{251}{254}{254}{ of}\myhl{255}{255}{255}{ the}\myhl{238}{250}{253}{ Metro}\myhl{251}{254}{254}{ Play}\myhl{255}{255}{255}{ Limited} \dots
\end{tcolorbox}
\begin{tcolorbox}[colback=white, , title=Average attention]
\textbf{Context}:
 \dots\myhl{218}{244}{252}{ firm}\myhl{224}{246}{252}{ linked}\myhl{216}{244}{251}{ to}\myhl{206}{241}{251}{ Harry}\myhl{211}{242}{251}{ Red}\myhl{202}{240}{250}{kn}\myhl{199}{239}{250}{app}\myhl{188}{236}{249}{ and}\myhl{233}{248}{253}{ two}\myhl{196}{238}{250}{ Premier}\myhl{209}{242}{251}{ League}\myhl{225}{246}{252}{ football}\myhl{131}{220}{245}{ clubs}\myhl{215}{244}{251}{ has}\myhl{240}{251}{253}{ been}\myhl{218}{244}{252}{ arrested}\myhl{236}{249}{253}{ over}\myhl{243}{251}{254}{ a} \dots\myhl{215}{243}{251}{ online}\myhl{224}{246}{252}{ gambling}\myhl{229}{247}{252}{ firm}\myhl{238}{250}{253}{,}\myhl{227}{247}{252}{ which}\myhl{172}{231}{248}{ signed}\myhl{180}{234}{249}{ Harry}\myhl{222}{245}{252}{ Red}\myhl{236}{249}{253}{kn}\myhl{221}{245}{252}{app}\myhl{169}{231}{248}{ to}\myhl{163}{229}{247}{ star}\myhl{146}{224}{246}{ in}\myhl{189}{236}{249}{ its}\myhl{0}{183}{234}{ adverts}\myhl{134}{221}{245}{ (}\myhl{221}{245}{252}{pictured}\myhl{239}{250}{253}{ above}\myhl{207}{241}{251}{)}\myhl{219}{244}{252}{ has}\myhl{245}{252}{254}{ been} \dots\myhl{242}{251}{254}{ fraud}\myhl{227}{247}{252}{ .}\myhl{234}{249}{253}{ The}\myhl{244}{252}{254}{ online}\myhl{233}{248}{253}{ firm}\myhl{199}{239}{250}{ announced}\myhl{151}{225}{246}{ deals}\myhl{169}{230}{248}{ with}\myhl{188}{236}{249}{ West}\myhl{159}{228}{247}{ Brom}\myhl{188}{236}{249}{wich}\myhl{185}{235}{249}{ Alb}\myhl{172}{231}{248}{ion}\myhl{131}{220}{245}{ and}\myhl{105}{213}{243}{ Leicester}\myhl{160}{228}{247}{ City}\myhl{184}{235}{249}{ last}\myhl{166}{230}{247}{ summer}\myhl{102}{212}{242}{.}\myhl{229}{247}{252}{ It}\myhl{233}{249}{253}{ also}\myhl{210}{242}{251}{ signed}\myhl{227}{247}{252}{ former}\myhl{235}{249}{253}{ Q} \dots\myhl{238}{250}{253}{app}\myhl{211}{242}{251}{ to}\myhl{215}{243}{251}{ appear}\myhl{230}{248}{253}{ in}\myhl{231}{248}{253}{ its}\myhl{176}{233}{248}{ adverts}\myhl{191}{237}{249}{,}\myhl{217}{244}{252}{ alongside}\myhl{241}{251}{253}{ Lock}\myhl{244}{252}{254}{,}\myhl{246}{252}{254}{ Stock}\myhl{245}{252}{254}{ and} \dots\myhl{210}{242}{251}{.}\myhl{225}{246}{252}{ Last}\myhl{240}{251}{253}{ week}\myhl{243}{251}{254}{,}\myhl{245}{252}{254}{ the}\myhl{198}{239}{250}{ UK}\myhl{170}{231}{248}{ gambling}\myhl{115}{215}{243}{ regulator}\myhl{191}{237}{249}{ The}\myhl{192}{237}{250}{ Gambling}\myhl{131}{220}{245}{ Commission}\myhl{97}{210}{242}{ suspended}\myhl{225}{246}{252}{ the}\myhl{214}{243}{251}{ firm}\myhl{217}{244}{251}{'s}\myhl{86}{207}{241}{ licence}\myhl{64}{201}{239}{.}\myhl{227}{247}{252}{ The}\myhl{193}{237}{250}{ suspension}\myhl{227}{247}{252}{ led}\myhl{240}{250}{253}{ to}\myhl{219}{245}{252}{ Football}\myhl{232}{248}{253}{ League}\myhl{231}{248}{253}{ One} \dots\myhl{219}{245}{252}{makers}\myhl{235}{249}{253}{ in}\myhl{227}{247}{252}{ August}\myhl{228}{247}{252}{ last}\myhl{229}{247}{252}{ year}\myhl{200}{239}{250}{,}\myhl{219}{245}{252}{ terminating}\myhl{242}{251}{253}{ its}\myhl{230}{248}{253}{ deal}\myhl{246}{252}{254}{ for}\myhl{231}{248}{253}{ shirt}\myhl{209}{242}{251}{ sponsorship}\myhl{190}{236}{249}{.}\myhl{236}{249}{253}{ It}\myhl{244}{252}{254}{ is}\myhl{250}{253}{254}{ not}\myhl{249}{253}{254}{ yet}\myhl{246}{252}{254}{ known}\myhl{248}{253}{254}{ whether}\myhl{234}{249}{253}{ the}\myhl{239}{250}{253}{ two}\myhl{208}{241}{251}{ Premier}\myhl{201}{240}{250}{ League}\myhl{168}{230}{248}{ clubs}\myhl{224}{246}{252}{ will}\myhl{236}{249}{253}{ follow}\myhl{228}{247}{252}{ suit}\myhl{222}{245}{252}{.}\myhl{242}{251}{253}{ Speaking} \dots\myhl{239}{250}{253}{ firm}\myhl{230}{248}{253}{ }\myhl{233}{248}{253}{666}\myhl{230}{248}{253}{Bet}\myhl{227}{247}{252}{ announced}\myhl{203}{240}{250}{ deals}\myhl{221}{245}{252}{ with}\myhl{208}{242}{251}{ West}\myhl{210}{242}{251}{ Brom}\myhl{187}{236}{249}{ and}\myhl{170}{231}{248}{ Leicester}\myhl{169}{231}{248}{ City}\myhl{235}{249}{253}{ last}\myhl{220}{245}{252}{ summer}\myhl{208}{242}{251}{.}\myhl{239}{250}{253}{ It}\myhl{240}{250}{253}{ also} \dots\myhl{253}{254}{254}{ will}\myhl{251}{253}{254}{ be}\myhl{245}{252}{254}{.'}\myhl{247}{252}{254}{ Although}\myhl{239}{250}{253}{ the}\myhl{199}{239}{250}{ Gambling}\myhl{187}{236}{249}{ Commission}\myhl{242}{251}{254}{ refused}\myhl{248}{253}{254}{ to}\myhl{245}{252}{254}{ comment}\myhl{228}{247}{252}{,}\myhl{244}{252}{254}{ it}\myhl{246}{252}{254}{ previously}\myhl{248}{253}{254}{ said}\myhl{239}{250}{253}{ the}\myhl{184}{235}{249}{ suspension}\myhl{240}{250}{253}{ of}\myhl{228}{247}{252}{ the}\myhl{232}{248}{253}{ Metro}\myhl{240}{250}{253}{ Play}\myhl{231}{248}{253}{ Limited} \dots\myhl{247}{253}{254}{ accepting}\myhl{241}{251}{253}{ comments}\myhl{244}{252}{254}{ on}\myhl{243}{251}{254}{ this}\myhl{233}{248}{253}{ article}\myhl{197}{239}{250}{.}
\end{tcolorbox}
\begin{tcolorbox}[colback=white, , title=ESM (256 samples)]
\textbf{Context}:
 \dots\myhl{255}{255}{255}{ online}\myhl{255}{255}{255}{ gambling}\myhl{255}{255}{255}{ firm}\myhl{255}{255}{255}{ linked}\myhl{255}{255}{255}{ to}\myhl{189}{236}{249}{ Harry}\myhl{203}{240}{250}{ Red}\myhl{255}{255}{255}{kn}\myhl{226}{247}{252}{app}\myhl{255}{255}{255}{ and}\myhl{255}{255}{255}{ two}\myhl{255}{255}{255}{ Premier} \dots\myhl{255}{255}{255}{.}\myhl{255}{255}{255}{ Six}\myhl{252}{254}{254}{ other}\myhl{255}{255}{255}{ people}\myhl{255}{255}{255}{ were}\myhl{194}{238}{250}{ arrested}\myhl{255}{255}{255}{,}\myhl{255}{255}{255}{ }\myhl{255}{255}{255}{13}\myhl{255}{255}{255}{ properties}\myhl{255}{255}{255}{ were} \dots\myhl{255}{255}{255}{ an}\myhl{255}{255}{255}{ online}\myhl{255}{255}{255}{ gambling}\myhl{255}{255}{255}{ firm}\myhl{255}{255}{255}{,}\myhl{198}{239}{250}{ which}\myhl{0}{183}{234}{ signed}\myhl{212}{243}{251}{ Harry}\myhl{205}{241}{251}{ Red}\myhl{252}{254}{254}{kn}\myhl{248}{253}{254}{app}\myhl{156}{227}{247}{ to}\myhl{80}{206}{241}{ star}\myhl{255}{255}{255}{ in}\myhl{255}{255}{255}{ its}\myhl{146}{224}{246}{ adverts}\myhl{255}{255}{255}{ (}\myhl{255}{255}{255}{pictured}\myhl{255}{255}{255}{ above}\myhl{255}{255}{255}{)}\myhl{255}{255}{255}{ has} \dots\myhl{255}{255}{255}{ arrested}\myhl{218}{244}{252}{ over}\myhl{255}{255}{255}{ a}\myhl{255}{255}{255}{ PS}\myhl{213}{243}{251}{21}\myhl{202}{240}{250}{m}\myhl{255}{255}{255}{ fraud}\myhl{255}{255}{255}{ .}\myhl{255}{255}{255}{ The}\myhl{255}{255}{255}{ online}\myhl{255}{255}{255}{ firm}\myhl{255}{255}{255}{ announced}\myhl{227}{247}{252}{ deals}\myhl{255}{255}{255}{ with}\myhl{148}{225}{246}{ West}\myhl{255}{255}{255}{ Brom}\myhl{255}{255}{255}{wich}\myhl{255}{255}{255}{ Alb}\myhl{255}{255}{255}{ion}\myhl{255}{255}{255}{ and}\myhl{161}{228}{247}{ Leicester}\myhl{250}{253}{254}{ City}\myhl{255}{255}{255}{ last}\myhl{218}{244}{252}{ summer}\myhl{255}{255}{255}{.}\myhl{250}{253}{254}{ It} \dots\myhl{255}{255}{255}{ regulator}\myhl{255}{255}{255}{ The}\myhl{255}{255}{255}{ Gambling}\myhl{255}{255}{255}{ Commission}\myhl{255}{255}{255}{ suspended}\myhl{191}{237}{249}{ the}\myhl{255}{255}{255}{ firm}\myhl{255}{255}{255}{'s}\myhl{255}{255}{255}{ licence}\myhl{255}{255}{255}{.}\myhl{255}{255}{255}{ The} \dots\myhl{255}{255}{255}{ firm}\myhl{255}{255}{255}{ }\myhl{255}{255}{255}{666}\myhl{255}{255}{255}{Bet}\myhl{255}{255}{255}{ announced}\myhl{183}{235}{249}{ deals}\myhl{255}{255}{255}{ with}\myhl{176}{233}{248}{ West}\myhl{255}{255}{255}{ Brom}\myhl{255}{255}{255}{ and}\myhl{236}{249}{253}{ Leicester}\myhl{255}{255}{255}{ City}\myhl{255}{255}{255}{ last} \dots\myhl{255}{255}{255}{ just}\myhl{238}{250}{253}{ did}\myhl{255}{255}{255}{ an}\myhl{255}{255}{255}{ advert}\myhl{231}{248}{253}{ for}\myhl{155}{227}{247}{ them}\myhl{255}{255}{255}{.}\myhl{255}{255}{255}{ I}\myhl{255}{255}{255}{ don}\myhl{255}{255}{255}{'t}\myhl{255}{255}{255}{ know} \dots\myhl{255}{255}{255}{ official}\myhl{255}{255}{255}{ Twitter}\myhl{255}{255}{255}{ account}\myhl{255}{255}{255}{ said}\myhl{243}{251}{254}{ the}\myhl{202}{240}{250}{ site}\myhl{255}{255}{255}{ was}\myhl{255}{255}{255}{ under}\myhl{255}{255}{255}{ maintenance}\myhl{255}{255}{255}{ due}\myhl{255}{255}{255}{ to} \dots\myhl{255}{255}{255}{seen}\myhl{255}{255}{255}{ circumstances}\myhl{255}{255}{255}{'}\myhl{212}{243}{251}{ In}\myhl{255}{255}{255}{ another}\myhl{197}{238}{250}{ tweet}\myhl{255}{255}{255}{,}\myhl{255}{255}{255}{ it}\myhl{255}{255}{255}{ says}\myhl{255}{255}{255}{:}\myhl{255}{255}{255}{ '} \dots\myhl{255}{255}{255}{ have}\myhl{255}{255}{255}{ the}\myhl{255}{255}{255}{ site}\myhl{255}{255}{255}{ and}\myhl{255}{255}{255}{ our}\myhl{184}{235}{249}{ licence}\myhl{255}{255}{255}{ back}\myhl{255}{255}{255}{ ASAP}\myhl{255}{255}{255}{ but}\myhl{255}{255}{255}{ cannot}\myhl{255}{255}{255}{ comment} \dots\myhl{255}{255}{255}{ of}\myhl{255}{255}{255}{ the}\myhl{255}{255}{255}{ Metro}\myhl{255}{255}{255}{ Play}\myhl{255}{255}{255}{ Limited}\myhl{195}{238}{250}{ licence}\myhl{209}{242}{251}{ -}\myhl{255}{255}{255}{ which}\myhl{255}{255}{255}{ covers}\myhl{255}{255}{255}{ }\myhl{255}{255}{255}{666} \dots
\end{tcolorbox}
\textbf{Query}:
Summarize the article in up to three sentences.
\newline\newline\textbf{Response} (from \texttt{Llama-3.1-8B}):
\newline
A director of online gambling firm 666Bet, Paul Bell, has been arrested as part of a PS21m fraud probe by HM Revenue \& Customs and the National Crime Agency. \myhl{249}{237}{91}{The firm, which signed Harry Redknapp to star in its adverts and had deals with West Bromwich Albion and Leicester City, had its licence suspended by the UK Gambling Commission.} The suspension has led to one of the firm's shirt sponsorship deals being terminated, and it is unclear whether the two Premier League clubs will follow suit.
\end{tcolorbox}
\relsize{1}
\caption{Visualized attributions for a random example from CNN DailyMail for \texttt{Llama-3.1-8B} with tokens as sources.}
\label{fig:cnn_dailymail_llama_3_1_8b_token}
\end{figure*}

\begin{figure*}[t]
\centering\definecolor{forestgreen}{RGB}{34, 139, 34}
\relsize{-1}
\begin{tcolorbox}[colback=white]
\begin{tcolorbox}[colback=white, colframe=forestgreen, title=\ours (general)]
\textbf{Context}:
 \dots\myhl{186}{235}{249}{ This}\myhl{186}{235}{249}{ time}\myhl{186}{235}{249}{ she}\myhl{186}{235}{249}{ seems}\myhl{186}{235}{249}{ to}\myhl{186}{235}{249}{ have}\myhl{186}{235}{249}{ decided}\myhl{186}{235}{249}{ to}\myhl{186}{235}{249}{ fully}\myhl{186}{235}{249}{ em}\myhl{186}{235}{249}{brace}\myhl{186}{235}{249}{ her}\myhl{186}{235}{249}{ woman}\myhl{186}{235}{249}{hood}\myhl{186}{235}{249}{ as}\myhl{186}{235}{249}{ an}\myhl{186}{235}{249}{ asset}\myhl{186}{235}{249}{ in}\myhl{186}{235}{249}{ her}\myhl{186}{235}{249}{ quest}\myhl{186}{235}{249}{ for}\myhl{186}{235}{249}{ the}\myhl{186}{235}{249}{ White}\myhl{186}{235}{249}{ House}\myhl{186}{235}{249}{ and}\myhl{186}{235}{249}{ to}\myhl{186}{235}{249}{ trust}\myhl{186}{235}{249}{ that}\myhl{186}{235}{249}{ the}\myhl{186}{235}{249}{ vot}\myhl{186}{235}{249}{ers}\myhl{186}{235}{249}{ will}\myhl{186}{235}{249}{ do}\myhl{186}{235}{249}{ the}\myhl{186}{235}{249}{ same}\myhl{186}{235}{249}{.} \dots\myhl{201}{239}{250}{ This}\myhl{201}{239}{250}{ time}\myhl{201}{239}{250}{ around}\myhl{201}{239}{250}{,}\myhl{201}{239}{250}{ there}\myhl{201}{239}{250}{ seems}\myhl{201}{239}{250}{ to}\myhl{201}{239}{250}{ be}\myhl{201}{239}{250}{ a}\myhl{201}{239}{250}{ shared}\myhl{201}{239}{250}{ assumption}\myhl{201}{239}{250}{ that}\myhl{201}{239}{250}{,}\myhl{201}{239}{250}{ of}\myhl{201}{239}{250}{ course}\myhl{201}{239}{250}{,}\myhl{201}{239}{250}{ a}\myhl{201}{239}{250}{ woman}\myhl{201}{239}{250}{ would}\myhl{201}{239}{250}{ make}\myhl{201}{239}{250}{ an}\myhl{201}{239}{250}{ excellent}\myhl{201}{239}{250}{ president}\myhl{201}{239}{250}{,}\myhl{201}{239}{250}{ in}\myhl{201}{239}{250}{ part}\myhl{201}{239}{250}{ simply}\myhl{201}{239}{250}{ by}\myhl{201}{239}{250}{ virtue}\myhl{201}{239}{250}{ of}\myhl{201}{239}{250}{ being}\myhl{201}{239}{250}{ a}\myhl{201}{239}{250}{ woman}\myhl{201}{239}{250}{.} \dots\myhl{0}{183}{234}{ If}\myhl{0}{183}{234}{ Hill}\myhl{0}{183}{234}{ary}\myhl{0}{183}{234}{ seems}\myhl{0}{183}{234}{ more}\myhl{0}{183}{234}{ comfortable}\myhl{0}{183}{234}{ running}\myhl{0}{183}{234}{ as}\myhl{0}{183}{234}{ a}\myhl{0}{183}{234}{ woman}\myhl{0}{183}{234}{,}\myhl{0}{183}{234}{ it}\myhl{0}{183}{234}{'}\myhl{0}{183}{234}{s}\myhl{0}{183}{234}{ partly}\myhl{0}{183}{234}{ because}\myhl{0}{183}{234}{ society}\myhl{0}{183}{234}{ at}\myhl{0}{183}{234}{ large}\myhl{0}{183}{234}{ seems}\myhl{0}{183}{234}{ more}\myhl{0}{183}{234}{ comfortable}\myhl{0}{183}{234}{ with}\myhl{0}{183}{234}{ a}\myhl{0}{183}{234}{ popul}\myhl{0}{183}{234}{ist}\myhl{0}{183}{234}{-}\myhl{0}{183}{234}{ting}\myhl{0}{183}{234}{ed}\myhl{0}{183}{234}{ femin}\myhl{0}{183}{234}{ism}\myhl{0}{183}{234}{ than}\myhl{0}{183}{234}{ it}\myhl{0}{183}{234}{ was}\myhl{0}{183}{234}{ in}\myhl{0}{183}{234}{ }\myhl{0}{183}{234}{2}\myhl{0}{183}{234}{0}\myhl{0}{183}{234}{0}\myhl{0}{183}{234}{8}\myhl{0}{183}{234}{.} \dots\myhl{181}{234}{249}{ Women}\myhl{181}{234}{249}{ have}\myhl{181}{234}{249}{ been}\myhl{181}{234}{249}{ stead}\myhl{181}{234}{249}{ily}\myhl{181}{234}{249}{ making}\myhl{181}{234}{249}{ str}\myhl{181}{234}{249}{ides}\myhl{181}{234}{249}{ in}\myhl{181}{234}{249}{ the}\myhl{181}{234}{249}{ years}\myhl{181}{234}{249}{ since}\myhl{181}{234}{249}{ Hill}\myhl{181}{234}{249}{ary}\myhl{181}{234}{249}{'}\myhl{181}{234}{249}{s}\myhl{181}{234}{249}{ }\myhl{181}{234}{249}{2}\myhl{181}{234}{249}{0}\myhl{181}{234}{249}{0}\myhl{181}{234}{249}{8}\myhl{181}{234}{249}{ campaign}\myhl{181}{234}{249}{,}\myhl{181}{234}{249}{ and}\myhl{181}{234}{249}{ as}\myhl{181}{234}{249}{ they}\myhl{181}{234}{249}{ did}\myhl{181}{234}{249}{,}\myhl{181}{234}{249}{ they}\myhl{181}{234}{249}{ smooth}\myhl{181}{234}{249}{ed}\myhl{181}{234}{249}{ the}\myhl{181}{234}{249}{ way}\myhl{181}{234}{249}{ for}\myhl{181}{234}{249}{ one}\myhl{181}{234}{249}{ of}\myhl{181}{234}{249}{ their}\myhl{181}{234}{249}{ own}\myhl{181}{234}{249}{ to}\myhl{181}{234}{249}{ run}\myhl{181}{234}{249}{ for}\myhl{181}{234}{249}{ the}\myhl{181}{234}{249}{ highest}\myhl{181}{234}{249}{ office}\myhl{181}{234}{249}{ in}\myhl{181}{234}{249}{ the}\myhl{181}{234}{249}{ land}\myhl{181}{234}{249}{ without}\myhl{181}{234}{249}{ having}\myhl{181}{234}{249}{ to}\myhl{181}{234}{249}{ play}\myhl{181}{234}{249}{ down}\myhl{181}{234}{249}{ her}\myhl{181}{234}{249}{ gender}\myhl{181}{234}{249}{.}\myhl{172}{232}{248}{ Hill}\myhl{172}{232}{248}{ary}\myhl{172}{232}{248}{ is}\myhl{172}{232}{248}{ now}\myhl{172}{232}{248}{ wis}\myhl{172}{232}{248}{ely}\myhl{172}{232}{248}{ emb}\myhl{172}{232}{248}{rac}\myhl{172}{232}{248}{ing}\myhl{172}{232}{248}{ her}\myhl{172}{232}{248}{ gender}\myhl{172}{232}{248}{ as}\myhl{172}{232}{248}{ a}\myhl{172}{232}{248}{ way}\myhl{172}{232}{248}{ of}\myhl{172}{232}{248}{ capt}\myhl{172}{232}{248}{uring}\myhl{172}{232}{248}{ the}\myhl{172}{232}{248}{ same}\myhl{172}{232}{248}{ "}\myhl{172}{232}{248}{ho}\myhl{172}{232}{248}{pe}\myhl{172}{232}{248}{ and}\myhl{172}{232}{248}{ change}\myhl{172}{232}{248}{"}\myhl{172}{232}{248}{ historical}\myhl{172}{232}{248}{ quality}\myhl{172}{232}{248}{ of}\myhl{172}{232}{248}{ Ob}\myhl{172}{232}{248}{ama}\myhl{172}{232}{248}{'}\myhl{172}{232}{248}{s}\myhl{172}{232}{248}{ presiden}\myhl{172}{232}{248}{cy}\myhl{172}{232}{248}{.} \dots
\end{tcolorbox}
\begin{tcolorbox}[colback=white, , title=Average attention]
\textbf{Context}:
 \dots\myhl{202}{240}{250}{ Of}\myhl{202}{240}{250}{ course}\myhl{202}{240}{250}{,}\myhl{202}{240}{250}{ Hill}\myhl{202}{240}{250}{ary}\myhl{202}{240}{250}{ hasn}\myhl{202}{240}{250}{'}\myhl{202}{240}{250}{t}\myhl{202}{240}{250}{ officially}\myhl{202}{240}{250}{ announced}\myhl{202}{240}{250}{ that}\myhl{202}{240}{250}{ she}\myhl{202}{240}{250}{ will}\myhl{202}{240}{250}{ be}\myhl{202}{240}{250}{ running}\myhl{202}{240}{250}{ for}\myhl{202}{240}{250}{ president}\myhl{202}{240}{250}{ --}\myhl{202}{240}{250}{ and}\myhl{202}{240}{250}{ Universal}\myhl{202}{240}{250}{ Studios}\myhl{202}{240}{250}{ has}\myhl{202}{240}{250}{ not}\myhl{202}{240}{250}{ officially}\myhl{202}{240}{250}{ announced}\myhl{202}{240}{250}{ that}\myhl{202}{240}{250}{ there}\myhl{202}{240}{250}{ will}\myhl{202}{240}{250}{ be}\myhl{202}{240}{250}{ a}\myhl{202}{240}{250}{ sequ}\myhl{202}{240}{250}{el}\myhl{202}{240}{250}{ to}\myhl{202}{240}{250}{ the}\myhl{202}{240}{250}{ block}\myhl{202}{240}{250}{b}\myhl{202}{240}{250}{uster}\myhl{202}{240}{250}{ "}\myhl{202}{240}{250}{5}\myhl{202}{240}{250}{0}\myhl{202}{240}{250}{ Sh}\myhl{202}{240}{250}{ades}\myhl{202}{240}{250}{ of}\myhl{202}{240}{250}{ Grey}\myhl{202}{240}{250}{."} \dots\myhl{0}{183}{234}{ Hill}\myhl{0}{183}{234}{ary}\myhl{0}{183}{234}{'}\myhl{0}{183}{234}{s}\myhl{0}{183}{234}{ stress}\myhl{0}{183}{234}{ing}\myhl{0}{183}{234}{ of}\myhl{0}{183}{234}{ the}\myhl{0}{183}{234}{ importance}\myhl{0}{183}{234}{ of}\myhl{0}{183}{234}{ women}\myhl{0}{183}{234}{'}\myhl{0}{183}{234}{s}\myhl{0}{183}{234}{ work}\myhl{0}{183}{234}{place}\myhl{0}{183}{234}{ issues}\myhl{0}{183}{234}{ both}\myhl{0}{183}{234}{ to}\myhl{0}{183}{234}{ the}\myhl{0}{183}{234}{ family}\myhl{0}{183}{234}{ and}\myhl{0}{183}{234}{ to}\myhl{0}{183}{234}{ the}\myhl{0}{183}{234}{ strugg}\myhl{0}{183}{234}{les}\myhl{0}{183}{234}{ of}\myhl{0}{183}{234}{ the}\myhl{0}{183}{234}{ middle}\myhl{0}{183}{234}{-}\myhl{0}{183}{234}{class}\myhl{0}{183}{234}{ puts}\myhl{0}{183}{234}{ the}\myhl{0}{183}{234}{ Republic}\myhl{0}{183}{234}{ans}\myhl{0}{183}{234}{ on}\myhl{0}{183}{234}{ the}\myhl{0}{183}{234}{ def}\myhl{0}{183}{234}{ensive}\myhl{0}{183}{234}{ as}\myhl{0}{183}{234}{ they}\myhl{0}{183}{234}{ try}\myhl{0}{183}{234}{ to}\myhl{0}{183}{234}{ co}\myhl{0}{183}{234}{-}\myhl{0}{183}{234}{opt}\myhl{0}{183}{234}{ middle}\myhl{0}{183}{234}{-}\myhl{0}{183}{234}{class}\myhl{0}{183}{234}{ economic}\myhl{0}{183}{234}{ issues}\myhl{0}{183}{234}{.} \dots\myhl{186}{235}{249}{ Women}\myhl{186}{235}{249}{ have}\myhl{186}{235}{249}{ been}\myhl{186}{235}{249}{ stead}\myhl{186}{235}{249}{ily}\myhl{186}{235}{249}{ making}\myhl{186}{235}{249}{ str}\myhl{186}{235}{249}{ides}\myhl{186}{235}{249}{ in}\myhl{186}{235}{249}{ the}\myhl{186}{235}{249}{ years}\myhl{186}{235}{249}{ since}\myhl{186}{235}{249}{ Hill}\myhl{186}{235}{249}{ary}\myhl{186}{235}{249}{'}\myhl{186}{235}{249}{s}\myhl{186}{235}{249}{ }\myhl{186}{235}{249}{2}\myhl{186}{235}{249}{0}\myhl{186}{235}{249}{0}\myhl{186}{235}{249}{8}\myhl{186}{235}{249}{ campaign}\myhl{186}{235}{249}{,}\myhl{186}{235}{249}{ and}\myhl{186}{235}{249}{ as}\myhl{186}{235}{249}{ they}\myhl{186}{235}{249}{ did}\myhl{186}{235}{249}{,}\myhl{186}{235}{249}{ they}\myhl{186}{235}{249}{ smooth}\myhl{186}{235}{249}{ed}\myhl{186}{235}{249}{ the}\myhl{186}{235}{249}{ way}\myhl{186}{235}{249}{ for}\myhl{186}{235}{249}{ one}\myhl{186}{235}{249}{ of}\myhl{186}{235}{249}{ their}\myhl{186}{235}{249}{ own}\myhl{186}{235}{249}{ to}\myhl{186}{235}{249}{ run}\myhl{186}{235}{249}{ for}\myhl{186}{235}{249}{ the}\myhl{186}{235}{249}{ highest}\myhl{186}{235}{249}{ office}\myhl{186}{235}{249}{ in}\myhl{186}{235}{249}{ the}\myhl{186}{235}{249}{ land}\myhl{186}{235}{249}{ without}\myhl{186}{235}{249}{ having}\myhl{186}{235}{249}{ to}\myhl{186}{235}{249}{ play}\myhl{186}{235}{249}{ down}\myhl{186}{235}{249}{ her}\myhl{186}{235}{249}{ gender}\myhl{186}{235}{249}{.} \dots
\end{tcolorbox}
\begin{tcolorbox}[colback=white, , title=ESM (256 samples)]
\textbf{Context}:
 \dots\myhl{186}{235}{249}{ In}\myhl{186}{235}{249}{ }\myhl{186}{235}{249}{2}\myhl{186}{235}{249}{0}\myhl{186}{235}{249}{0}\myhl{186}{235}{249}{8}\myhl{186}{235}{249}{,}\myhl{186}{235}{249}{ she}\myhl{186}{235}{249}{ seemed}\myhl{186}{235}{249}{ to}\myhl{186}{235}{249}{ think}\myhl{186}{235}{249}{ that}\myhl{186}{235}{249}{ she}\myhl{186}{235}{249}{ had}\myhl{186}{235}{249}{ to}\myhl{186}{235}{249}{ re}\myhl{186}{235}{249}{ass}\myhl{186}{235}{249}{ure}\myhl{186}{235}{249}{ vot}\myhl{186}{235}{249}{ers}\myhl{186}{235}{249}{ that}\myhl{186}{235}{249}{ a}\myhl{186}{235}{249}{ woman}\myhl{186}{235}{249}{ could}\myhl{186}{235}{249}{ be}\myhl{186}{235}{249}{ president}\myhl{186}{235}{249}{,}\myhl{186}{235}{249}{ primarily}\myhl{186}{235}{249}{ by}\myhl{186}{235}{249}{ not}\myhl{186}{235}{249}{ acc}\myhl{186}{235}{249}{ent}\myhl{186}{235}{249}{u}\myhl{186}{235}{249}{ating}\myhl{186}{235}{249}{ the}\myhl{186}{235}{249}{ fact}\myhl{186}{235}{249}{ that}\myhl{186}{235}{249}{ she}\myhl{186}{235}{249}{ was}\myhl{186}{235}{249}{ a}\myhl{186}{235}{249}{ woman}\myhl{186}{235}{249}{.}\myhl{201}{240}{250}{ This}\myhl{201}{240}{250}{ time}\myhl{201}{240}{250}{ around}\myhl{201}{240}{250}{,}\myhl{201}{240}{250}{ there}\myhl{201}{240}{250}{ seems}\myhl{201}{240}{250}{ to}\myhl{201}{240}{250}{ be}\myhl{201}{240}{250}{ a}\myhl{201}{240}{250}{ shared}\myhl{201}{240}{250}{ assumption}\myhl{201}{240}{250}{ that}\myhl{201}{240}{250}{,}\myhl{201}{240}{250}{ of}\myhl{201}{240}{250}{ course}\myhl{201}{240}{250}{,}\myhl{201}{240}{250}{ a}\myhl{201}{240}{250}{ woman}\myhl{201}{240}{250}{ would}\myhl{201}{240}{250}{ make}\myhl{201}{240}{250}{ an}\myhl{201}{240}{250}{ excellent}\myhl{201}{240}{250}{ president}\myhl{201}{240}{250}{,}\myhl{201}{240}{250}{ in}\myhl{201}{240}{250}{ part}\myhl{201}{240}{250}{ simply}\myhl{201}{240}{250}{ by}\myhl{201}{240}{250}{ virtue}\myhl{201}{240}{250}{ of}\myhl{201}{240}{250}{ being}\myhl{201}{240}{250}{ a}\myhl{201}{240}{250}{ woman}\myhl{201}{240}{250}{.} \dots\myhl{182}{234}{249}{ Fort}\myhl{182}{234}{249}{un}\myhl{182}{234}{249}{ately}\myhl{182}{234}{249}{,}\myhl{182}{234}{249}{ the}\myhl{182}{234}{249}{ dynamics}\myhl{182}{234}{249}{ of}\myhl{182}{234}{249}{ the}\myhl{182}{234}{249}{ }\myhl{182}{234}{249}{2}\myhl{182}{234}{249}{1}\myhl{182}{234}{249}{st}\myhl{182}{234}{249}{ century}\myhl{182}{234}{249}{ economy}\myhl{182}{234}{249}{ and}\myhl{182}{234}{249}{ the}\myhl{182}{234}{249}{ place}\myhl{182}{234}{249}{ of}\myhl{182}{234}{249}{ the}\myhl{182}{234}{249}{ family}\myhl{182}{234}{249}{ within}\myhl{182}{234}{249}{ it}\myhl{182}{234}{249}{ l}\myhl{182}{234}{249}{end}\myhl{182}{234}{249}{ themselves}\myhl{182}{234}{249}{ to}\myhl{182}{234}{249}{ a}\myhl{182}{234}{249}{ family}\myhl{182}{234}{249}{-}\myhl{182}{234}{249}{friend}\myhl{182}{234}{249}{ly}\myhl{182}{234}{249}{ femin}\myhl{182}{234}{249}{ism}\myhl{182}{234}{249}{.} \dots\myhl{0}{183}{234}{ If}\myhl{0}{183}{234}{ Hill}\myhl{0}{183}{234}{ary}\myhl{0}{183}{234}{ seems}\myhl{0}{183}{234}{ more}\myhl{0}{183}{234}{ comfortable}\myhl{0}{183}{234}{ running}\myhl{0}{183}{234}{ as}\myhl{0}{183}{234}{ a}\myhl{0}{183}{234}{ woman}\myhl{0}{183}{234}{,}\myhl{0}{183}{234}{ it}\myhl{0}{183}{234}{'}\myhl{0}{183}{234}{s}\myhl{0}{183}{234}{ partly}\myhl{0}{183}{234}{ because}\myhl{0}{183}{234}{ society}\myhl{0}{183}{234}{ at}\myhl{0}{183}{234}{ large}\myhl{0}{183}{234}{ seems}\myhl{0}{183}{234}{ more}\myhl{0}{183}{234}{ comfortable}\myhl{0}{183}{234}{ with}\myhl{0}{183}{234}{ a}\myhl{0}{183}{234}{ popul}\myhl{0}{183}{234}{ist}\myhl{0}{183}{234}{-}\myhl{0}{183}{234}{ting}\myhl{0}{183}{234}{ed}\myhl{0}{183}{234}{ femin}\myhl{0}{183}{234}{ism}\myhl{0}{183}{234}{ than}\myhl{0}{183}{234}{ it}\myhl{0}{183}{234}{ was}\myhl{0}{183}{234}{ in}\myhl{0}{183}{234}{ }\myhl{0}{183}{234}{2}\myhl{0}{183}{234}{0}\myhl{0}{183}{234}{0}\myhl{0}{183}{234}{8}\myhl{0}{183}{234}{.} \dots\myhl{196}{238}{250}{ Women}\myhl{196}{238}{250}{ have}\myhl{196}{238}{250}{ been}\myhl{196}{238}{250}{ stead}\myhl{196}{238}{250}{ily}\myhl{196}{238}{250}{ making}\myhl{196}{238}{250}{ str}\myhl{196}{238}{250}{ides}\myhl{196}{238}{250}{ in}\myhl{196}{238}{250}{ the}\myhl{196}{238}{250}{ years}\myhl{196}{238}{250}{ since}\myhl{196}{238}{250}{ Hill}\myhl{196}{238}{250}{ary}\myhl{196}{238}{250}{'}\myhl{196}{238}{250}{s}\myhl{196}{238}{250}{ }\myhl{196}{238}{250}{2}\myhl{196}{238}{250}{0}\myhl{196}{238}{250}{0}\myhl{196}{238}{250}{8}\myhl{196}{238}{250}{ campaign}\myhl{196}{238}{250}{,}\myhl{196}{238}{250}{ and}\myhl{196}{238}{250}{ as}\myhl{196}{238}{250}{ they}\myhl{196}{238}{250}{ did}\myhl{196}{238}{250}{,}\myhl{196}{238}{250}{ they}\myhl{196}{238}{250}{ smooth}\myhl{196}{238}{250}{ed}\myhl{196}{238}{250}{ the}\myhl{196}{238}{250}{ way}\myhl{196}{238}{250}{ for}\myhl{196}{238}{250}{ one}\myhl{196}{238}{250}{ of}\myhl{196}{238}{250}{ their}\myhl{196}{238}{250}{ own}\myhl{196}{238}{250}{ to}\myhl{196}{238}{250}{ run}\myhl{196}{238}{250}{ for}\myhl{196}{238}{250}{ the}\myhl{196}{238}{250}{ highest}\myhl{196}{238}{250}{ office}\myhl{196}{238}{250}{ in}\myhl{196}{238}{250}{ the}\myhl{196}{238}{250}{ land}\myhl{196}{238}{250}{ without}\myhl{196}{238}{250}{ having}\myhl{196}{238}{250}{ to}\myhl{196}{238}{250}{ play}\myhl{196}{238}{250}{ down}\myhl{196}{238}{250}{ her}\myhl{196}{238}{250}{ gender}\myhl{196}{238}{250}{.} \dots
\end{tcolorbox}
\textbf{Query}:
Summarize the article in up to three sentences.
\newline\newline\textbf{Response} (from \texttt{Phi-3.5-Mini}):
\newline
Hillary Clinton is embracing her womanhood as a key aspect of her 2016 presidential campaign, highlighting women's economic issues and the importance of family-friendly policies. She is set to speak at numerous events focused on women's issues during Women's History Month, marking a stark contrast to her 2008 campaign. \myhl{249}{237}{91}{Clinton's approach reflects a societal shift towards a more inclusive feminism, with her gender now seen as an asset rather than a point of contention.}
\end{tcolorbox}
\relsize{1}
\caption{Visualized attributions for a random example from CNN DailyMail for \texttt{Phi-3.5-Mini} with sentences as sources.}
\label{fig:cnn_dailymail_phi_3_5_mini_sentence}
\end{figure*}

\begin{figure*}[t]
\centering\definecolor{forestgreen}{RGB}{34, 139, 34}
\relsize{-1}
\begin{tcolorbox}[colback=white]
\begin{tcolorbox}[colback=white, colframe=forestgreen, title=\ours (general)]
\textbf{Context}:
\myhl{120}{217}{244}{ A}\myhl{120}{217}{244}{ have}\myhl{120}{217}{244}{-a}\myhl{120}{217}{244}{-go}\myhl{120}{217}{244}{-hero}\myhl{120}{217}{244}{ was}\myhl{120}{217}{244}{ so}\myhl{120}{217}{244}{ badly}\myhl{120}{217}{244}{ beaten}\myhl{120}{217}{244}{ and}\myhl{120}{217}{244}{ covered}\myhl{120}{217}{244}{ in}\myhl{120}{217}{244}{ blood}\myhl{120}{217}{244}{ after}\myhl{120}{217}{244}{ he}\myhl{120}{217}{244}{ challenged}\myhl{120}{217}{244}{ a}\myhl{120}{217}{244}{ car}\myhl{120}{217}{244}{ thief}\myhl{120}{217}{244}{ that}\myhl{120}{217}{244}{ his}\myhl{120}{217}{244}{ family}\myhl{120}{217}{244}{ thought}\myhl{120}{217}{244}{ he}\myhl{120}{217}{244}{ had}\myhl{120}{217}{244}{ been}\myhl{120}{217}{244}{ shot}\myhl{120}{217}{244}{.}\myhl{0}{183}{234}{ Danny}\myhl{0}{183}{234}{ Mason}\myhl{0}{183}{234}{,}\myhl{0}{183}{234}{ }\myhl{0}{183}{234}{37}\myhl{0}{183}{234}{,}\myhl{0}{183}{234}{ was}\myhl{0}{183}{234}{ repeatedly}\myhl{0}{183}{234}{ punched}\myhl{0}{183}{234}{ in}\myhl{0}{183}{234}{ the}\myhl{0}{183}{234}{ face}\myhl{0}{183}{234}{ and}\myhl{0}{183}{234}{ left}\myhl{0}{183}{234}{ with}\myhl{0}{183}{234}{ a}\myhl{0}{183}{234}{ fractured}\myhl{0}{183}{234}{ cheek}\myhl{0}{183}{234}{,}\myhl{0}{183}{234}{ broken}\myhl{0}{183}{234}{ nose}\myhl{0}{183}{234}{ and}\myhl{0}{183}{234}{ cut}\myhl{0}{183}{234}{ lip}\myhl{0}{183}{234}{ after}\myhl{0}{183}{234}{ he}\myhl{0}{183}{234}{ approached}\myhl{0}{183}{234}{ the}\myhl{0}{183}{234}{ would}\myhl{0}{183}{234}{-be}\myhl{0}{183}{234}{-th}\myhl{0}{183}{234}{ief}\myhl{0}{183}{234}{ in}\myhl{0}{183}{234}{ Blackburn}\myhl{0}{183}{234}{,}\myhl{0}{183}{234}{ Lanc}\myhl{0}{183}{234}{ashire}\myhl{0}{183}{234}{.}\myhl{45}{196}{238}{ He}\myhl{45}{196}{238}{ had}\myhl{45}{196}{238}{ just}\myhl{45}{196}{238}{ left}\myhl{45}{196}{238}{ a}\myhl{45}{196}{238}{ family}\myhl{45}{196}{238}{ party}\myhl{45}{196}{238}{ at}\myhl{45}{196}{238}{ around}\myhl{45}{196}{238}{ }\myhl{45}{196}{238}{11}\myhl{45}{196}{238}{.}\myhl{45}{196}{238}{45}\myhl{45}{196}{238}{pm}\myhl{45}{196}{238}{ on}\myhl{45}{196}{238}{ Saturday}\myhl{45}{196}{238}{ when}\myhl{45}{196}{238}{ he}\myhl{45}{196}{238}{ tried}\myhl{45}{196}{238}{ to}\myhl{45}{196}{238}{ stop}\myhl{45}{196}{238}{ the}\myhl{45}{196}{238}{ would}\myhl{45}{196}{238}{-be}\myhl{45}{196}{238}{-th}\myhl{45}{196}{238}{ief}\myhl{45}{196}{238}{ from}\myhl{45}{196}{238}{ stealing}\myhl{45}{196}{238}{ a}\myhl{45}{196}{238}{ number}\myhl{45}{196}{238}{ of}\myhl{45}{196}{238}{ cars}\myhl{45}{196}{238}{.}\myhl{150}{225}{246}{ The}\myhl{150}{225}{246}{ }\myhl{150}{225}{246}{37}\myhl{150}{225}{246}{-year}\myhl{150}{225}{246}{-old}\myhl{150}{225}{246}{,}\myhl{150}{225}{246}{ who}\myhl{150}{225}{246}{ now}\myhl{150}{225}{246}{ lives}\myhl{150}{225}{246}{ in}\myhl{150}{225}{246}{ Kid}\myhl{150}{225}{246}{der}\myhl{150}{225}{246}{minster}\myhl{150}{225}{246}{,}\myhl{150}{225}{246}{ Worce}\myhl{150}{225}{246}{sters}\myhl{150}{225}{246}{hire}\myhl{150}{225}{246}{,}\myhl{150}{225}{246}{ was}\myhl{150}{225}{246}{ found}\myhl{150}{225}{246}{ sl}\myhl{150}{225}{246}{umped}\myhl{150}{225}{246}{ unconscious}\myhl{150}{225}{246}{ on}\myhl{150}{225}{246}{ the}\myhl{150}{225}{246}{ ground}\myhl{150}{225}{246}{ by}\myhl{150}{225}{246}{ family}\myhl{150}{225}{246}{ members}\myhl{150}{225}{246}{ and}\myhl{150}{225}{246}{ is}\myhl{150}{225}{246}{ now}\myhl{150}{225}{246}{ in}\myhl{150}{225}{246}{ so}\myhl{150}{225}{246}{ much}\myhl{150}{225}{246}{ pain}\myhl{150}{225}{246}{ that}\myhl{150}{225}{246}{ he}\myhl{150}{225}{246}{ cannot}\myhl{150}{225}{246}{ sleep}\myhl{150}{225}{246}{ or}\myhl{150}{225}{246}{ work}\myhl{150}{225}{246}{.}\myhl{171}{231}{248}{ Danny}\myhl{171}{231}{248}{ Mason}\myhl{171}{231}{248}{,}\myhl{171}{231}{248}{ }\myhl{171}{231}{248}{37}\myhl{171}{231}{248}{,}\myhl{171}{231}{248}{ was}\myhl{171}{231}{248}{ repeatedly}\myhl{171}{231}{248}{ punched}\myhl{171}{231}{248}{ in}\myhl{171}{231}{248}{ the}\myhl{171}{231}{248}{ face}\myhl{171}{231}{248}{ and}\myhl{171}{231}{248}{ left}\myhl{171}{231}{248}{ with}\myhl{171}{231}{248}{ a}\myhl{171}{231}{248}{ fractured}\myhl{171}{231}{248}{ cheek}\myhl{171}{231}{248}{,}\myhl{171}{231}{248}{ broken}\myhl{171}{231}{248}{ nose}\myhl{171}{231}{248}{ and}\myhl{171}{231}{248}{ cut}\myhl{171}{231}{248}{ lip}\myhl{171}{231}{248}{ after}\myhl{171}{231}{248}{ challenging}\myhl{171}{231}{248}{ a}\myhl{171}{231}{248}{ car}\myhl{171}{231}{248}{ thief}\myhl{171}{231}{248}{ .} \dots\myhl{188}{236}{249}{ He}\myhl{188}{236}{249}{ had}\myhl{188}{236}{249}{ just}\myhl{188}{236}{249}{ left}\myhl{188}{236}{249}{ a}\myhl{188}{236}{249}{ family}\myhl{188}{236}{249}{ party}\myhl{188}{236}{249}{ at}\myhl{188}{236}{249}{ around}\myhl{188}{236}{249}{ }\myhl{188}{236}{249}{11}\myhl{188}{236}{249}{.}\myhl{188}{236}{249}{45}\myhl{188}{236}{249}{pm}\myhl{188}{236}{249}{ on}\myhl{188}{236}{249}{ Saturday}\myhl{188}{236}{249}{ when}\myhl{188}{236}{249}{ he}\myhl{188}{236}{249}{ approached}\myhl{188}{236}{249}{ the}\myhl{188}{236}{249}{ would}\myhl{188}{236}{249}{-be}\myhl{188}{236}{249}{ thief}\myhl{188}{236}{249}{ and}\myhl{188}{236}{249}{ was}\myhl{188}{236}{249}{ vicious}\myhl{188}{236}{249}{ly}\myhl{188}{236}{249}{ attacked}\myhl{188}{236}{249}{ in}\myhl{188}{236}{249}{ Park}\myhl{188}{236}{249}{es}\myhl{188}{236}{249}{ Way}\myhl{188}{236}{249}{,}\myhl{188}{236}{249}{ Blackburn}\myhl{188}{236}{249}{ .} \dots
\end{tcolorbox}
\begin{tcolorbox}[colback=white, , title=Average attention]
\textbf{Context}:
\myhl{45}{196}{238}{ A}\myhl{45}{196}{238}{ have}\myhl{45}{196}{238}{-a}\myhl{45}{196}{238}{-go}\myhl{45}{196}{238}{-hero}\myhl{45}{196}{238}{ was}\myhl{45}{196}{238}{ so}\myhl{45}{196}{238}{ badly}\myhl{45}{196}{238}{ beaten}\myhl{45}{196}{238}{ and}\myhl{45}{196}{238}{ covered}\myhl{45}{196}{238}{ in}\myhl{45}{196}{238}{ blood}\myhl{45}{196}{238}{ after}\myhl{45}{196}{238}{ he}\myhl{45}{196}{238}{ challenged}\myhl{45}{196}{238}{ a}\myhl{45}{196}{238}{ car}\myhl{45}{196}{238}{ thief}\myhl{45}{196}{238}{ that}\myhl{45}{196}{238}{ his}\myhl{45}{196}{238}{ family}\myhl{45}{196}{238}{ thought}\myhl{45}{196}{238}{ he}\myhl{45}{196}{238}{ had}\myhl{45}{196}{238}{ been}\myhl{45}{196}{238}{ shot}\myhl{45}{196}{238}{.}\myhl{0}{183}{234}{ Danny}\myhl{0}{183}{234}{ Mason}\myhl{0}{183}{234}{,}\myhl{0}{183}{234}{ }\myhl{0}{183}{234}{37}\myhl{0}{183}{234}{,}\myhl{0}{183}{234}{ was}\myhl{0}{183}{234}{ repeatedly}\myhl{0}{183}{234}{ punched}\myhl{0}{183}{234}{ in}\myhl{0}{183}{234}{ the}\myhl{0}{183}{234}{ face}\myhl{0}{183}{234}{ and}\myhl{0}{183}{234}{ left}\myhl{0}{183}{234}{ with}\myhl{0}{183}{234}{ a}\myhl{0}{183}{234}{ fractured}\myhl{0}{183}{234}{ cheek}\myhl{0}{183}{234}{,}\myhl{0}{183}{234}{ broken}\myhl{0}{183}{234}{ nose}\myhl{0}{183}{234}{ and}\myhl{0}{183}{234}{ cut}\myhl{0}{183}{234}{ lip}\myhl{0}{183}{234}{ after}\myhl{0}{183}{234}{ he}\myhl{0}{183}{234}{ approached}\myhl{0}{183}{234}{ the}\myhl{0}{183}{234}{ would}\myhl{0}{183}{234}{-be}\myhl{0}{183}{234}{-th}\myhl{0}{183}{234}{ief}\myhl{0}{183}{234}{ in}\myhl{0}{183}{234}{ Blackburn}\myhl{0}{183}{234}{,}\myhl{0}{183}{234}{ Lanc}\myhl{0}{183}{234}{ashire}\myhl{0}{183}{234}{.}\myhl{32}{192}{237}{ He}\myhl{32}{192}{237}{ had}\myhl{32}{192}{237}{ just}\myhl{32}{192}{237}{ left}\myhl{32}{192}{237}{ a}\myhl{32}{192}{237}{ family}\myhl{32}{192}{237}{ party}\myhl{32}{192}{237}{ at}\myhl{32}{192}{237}{ around}\myhl{32}{192}{237}{ }\myhl{32}{192}{237}{11}\myhl{32}{192}{237}{.}\myhl{32}{192}{237}{45}\myhl{32}{192}{237}{pm}\myhl{32}{192}{237}{ on}\myhl{32}{192}{237}{ Saturday}\myhl{32}{192}{237}{ when}\myhl{32}{192}{237}{ he}\myhl{32}{192}{237}{ tried}\myhl{32}{192}{237}{ to}\myhl{32}{192}{237}{ stop}\myhl{32}{192}{237}{ the}\myhl{32}{192}{237}{ would}\myhl{32}{192}{237}{-be}\myhl{32}{192}{237}{-th}\myhl{32}{192}{237}{ief}\myhl{32}{192}{237}{ from}\myhl{32}{192}{237}{ stealing}\myhl{32}{192}{237}{ a}\myhl{32}{192}{237}{ number}\myhl{32}{192}{237}{ of}\myhl{32}{192}{237}{ cars}\myhl{32}{192}{237}{.}\myhl{100}{211}{242}{ The}\myhl{100}{211}{242}{ }\myhl{100}{211}{242}{37}\myhl{100}{211}{242}{-year}\myhl{100}{211}{242}{-old}\myhl{100}{211}{242}{,}\myhl{100}{211}{242}{ who}\myhl{100}{211}{242}{ now}\myhl{100}{211}{242}{ lives}\myhl{100}{211}{242}{ in}\myhl{100}{211}{242}{ Kid}\myhl{100}{211}{242}{der}\myhl{100}{211}{242}{minster}\myhl{100}{211}{242}{,}\myhl{100}{211}{242}{ Worce}\myhl{100}{211}{242}{sters}\myhl{100}{211}{242}{hire}\myhl{100}{211}{242}{,}\myhl{100}{211}{242}{ was}\myhl{100}{211}{242}{ found}\myhl{100}{211}{242}{ sl}\myhl{100}{211}{242}{umped}\myhl{100}{211}{242}{ unconscious}\myhl{100}{211}{242}{ on}\myhl{100}{211}{242}{ the}\myhl{100}{211}{242}{ ground}\myhl{100}{211}{242}{ by}\myhl{100}{211}{242}{ family}\myhl{100}{211}{242}{ members}\myhl{100}{211}{242}{ and}\myhl{100}{211}{242}{ is}\myhl{100}{211}{242}{ now}\myhl{100}{211}{242}{ in}\myhl{100}{211}{242}{ so}\myhl{100}{211}{242}{ much}\myhl{100}{211}{242}{ pain}\myhl{100}{211}{242}{ that}\myhl{100}{211}{242}{ he}\myhl{100}{211}{242}{ cannot}\myhl{100}{211}{242}{ sleep}\myhl{100}{211}{242}{ or}\myhl{100}{211}{242}{ work}\myhl{100}{211}{242}{.}\myhl{134}{221}{245}{ Danny}\myhl{134}{221}{245}{ Mason}\myhl{134}{221}{245}{,}\myhl{134}{221}{245}{ }\myhl{134}{221}{245}{37}\myhl{134}{221}{245}{,}\myhl{134}{221}{245}{ was}\myhl{134}{221}{245}{ repeatedly}\myhl{134}{221}{245}{ punched}\myhl{134}{221}{245}{ in}\myhl{134}{221}{245}{ the}\myhl{134}{221}{245}{ face}\myhl{134}{221}{245}{ and}\myhl{134}{221}{245}{ left}\myhl{134}{221}{245}{ with}\myhl{134}{221}{245}{ a}\myhl{134}{221}{245}{ fractured}\myhl{134}{221}{245}{ cheek}\myhl{134}{221}{245}{,}\myhl{134}{221}{245}{ broken}\myhl{134}{221}{245}{ nose}\myhl{134}{221}{245}{ and}\myhl{134}{221}{245}{ cut}\myhl{134}{221}{245}{ lip}\myhl{134}{221}{245}{ after}\myhl{134}{221}{245}{ challenging}\myhl{134}{221}{245}{ a}\myhl{134}{221}{245}{ car}\myhl{134}{221}{245}{ thief}\myhl{134}{221}{245}{ .} \dots\myhl{200}{239}{250}{ '}\myhl{200}{239}{250}{Looking}\myhl{200}{239}{250}{ back}\myhl{200}{239}{250}{,}\myhl{200}{239}{250}{ I}\myhl{200}{239}{250}{ thought}\myhl{200}{239}{250}{ I}\myhl{200}{239}{250}{ was}\myhl{200}{239}{250}{ doing}\myhl{200}{239}{250}{ the}\myhl{200}{239}{250}{ right}\myhl{200}{239}{250}{ thing}\myhl{200}{239}{250}{ by}\myhl{200}{239}{250}{ speaking}\myhl{200}{239}{250}{ to}\myhl{200}{239}{250}{ this}\myhl{200}{239}{250}{ man}\myhl{200}{239}{250}{,}\myhl{200}{239}{250}{ but}\myhl{200}{239}{250}{ now}\myhl{200}{239}{250}{ I}\myhl{200}{239}{250}{ am}\myhl{200}{239}{250}{ in}\myhl{200}{239}{250}{ so}\myhl{200}{239}{250}{ much}\myhl{200}{239}{250}{ pain}\myhl{200}{239}{250}{ that}\myhl{200}{239}{250}{ I}\myhl{200}{239}{250}{ cannot}\myhl{200}{239}{250}{ sleep}\myhl{200}{239}{250}{ or}\myhl{200}{239}{250}{ work}\myhl{200}{239}{250}{.'} \dots\myhl{193}{237}{250}{ Police}\myhl{193}{237}{250}{ have}\myhl{193}{237}{250}{ now}\myhl{193}{237}{250}{ released}\myhl{193}{237}{250}{ CCTV}\myhl{193}{237}{250}{ images}\myhl{193}{237}{250}{ of}\myhl{193}{237}{250}{ a}\myhl{193}{237}{250}{ white}\myhl{193}{237}{250}{ man}\myhl{193}{237}{250}{ around}\myhl{193}{237}{250}{ }\myhl{193}{237}{250}{6}\myhl{193}{237}{250}{ft}\myhl{193}{237}{250}{ tall}\myhl{193}{237}{250}{ who}\myhl{193}{237}{250}{ they}\myhl{193}{237}{250}{ would}\myhl{193}{237}{250}{ like}\myhl{193}{237}{250}{ to}\myhl{193}{237}{250}{ speak}\myhl{193}{237}{250}{ to}\myhl{193}{237}{250}{ in}\myhl{193}{237}{250}{ connection}\myhl{193}{237}{250}{ with}\myhl{193}{237}{250}{ the}\myhl{193}{237}{250}{ incident}\myhl{193}{237}{250}{.}\myhl{199}{239}{250}{ Mr}\myhl{199}{239}{250}{ Mason}\myhl{199}{239}{250}{'s}\myhl{199}{239}{250}{ sister}\myhl{199}{239}{250}{ Sue}\myhl{199}{239}{250}{ B}\myhl{199}{239}{250}{ann}\myhl{199}{239}{250}{ister}\myhl{199}{239}{250}{ said}\myhl{199}{239}{250}{ that}\myhl{199}{239}{250}{ the}\myhl{199}{239}{250}{ family}\myhl{199}{239}{250}{ have}\myhl{199}{239}{250}{ been}\myhl{199}{239}{250}{ '}\myhl{199}{239}{250}{deep}\myhl{199}{239}{250}{ly}\myhl{199}{239}{250}{ shocked}\myhl{199}{239}{250}{ and}\myhl{199}{239}{250}{ upset}\myhl{199}{239}{250}{'}\myhl{199}{239}{250}{ by}\myhl{199}{239}{250}{ the}\myhl{199}{239}{250}{ attack}\myhl{199}{239}{250}{ and}\myhl{199}{239}{250}{ asked}\myhl{199}{239}{250}{ anyone}\myhl{199}{239}{250}{ with}\myhl{199}{239}{250}{ information}\myhl{199}{239}{250}{ to}\myhl{199}{239}{250}{ come}\myhl{199}{239}{250}{ forward}\myhl{199}{239}{250}{.} \dots\myhl{176}{233}{248}{ '}\myhl{176}{233}{248}{Danny}\myhl{176}{233}{248}{ had}\myhl{176}{233}{248}{ gone}\myhl{176}{233}{248}{ outside}\myhl{176}{233}{248}{ for}\myhl{176}{233}{248}{ a}\myhl{176}{233}{248}{ cigarette}\myhl{176}{233}{248}{ when}\myhl{176}{233}{248}{ he}\myhl{176}{233}{248}{ spotted}\myhl{176}{233}{248}{ a}\myhl{176}{233}{248}{ man}\myhl{176}{233}{248}{ who}\myhl{176}{233}{248}{ looked}\myhl{176}{233}{248}{ like}\myhl{176}{233}{248}{ he}\myhl{176}{233}{248}{ was}\myhl{176}{233}{248}{ trying}\myhl{176}{233}{248}{ to}\myhl{176}{233}{248}{ break}\myhl{176}{233}{248}{ into}\myhl{176}{233}{248}{ vehicles}\myhl{176}{233}{248}{.} \dots\myhl{143}{223}{246}{ The}\myhl{143}{223}{246}{ }\myhl{143}{223}{246}{37}\myhl{143}{223}{246}{-year}\myhl{143}{223}{246}{-old}\myhl{143}{223}{246}{ (}\myhl{143}{223}{246}{before}\myhl{143}{223}{246}{ the}\myhl{143}{223}{246}{ attack}\myhl{143}{223}{246}{,}\myhl{143}{223}{246}{ left}\myhl{143}{223}{246}{)}\myhl{143}{223}{246}{ was}\myhl{143}{223}{246}{ found}\myhl{143}{223}{246}{ sl}\myhl{143}{223}{246}{umped}\myhl{143}{223}{246}{ unconscious}\myhl{143}{223}{246}{ on}\myhl{143}{223}{246}{ the}\myhl{143}{223}{246}{ ground}\myhl{143}{223}{246}{ by}\myhl{143}{223}{246}{ family}\myhl{143}{223}{246}{ members}\myhl{143}{223}{246}{ and}\myhl{143}{223}{246}{ is}\myhl{143}{223}{246}{ now}\myhl{143}{223}{246}{ in}\myhl{143}{223}{246}{ so}\myhl{143}{223}{246}{ much}\myhl{143}{223}{246}{ pain}\myhl{143}{223}{246}{ that}\myhl{143}{223}{246}{ he}\myhl{143}{223}{246}{ cannot}\myhl{143}{223}{246}{ sleep}\myhl{143}{223}{246}{ or}\myhl{143}{223}{246}{ work}\myhl{143}{223}{246}{ (}\myhl{143}{223}{246}{after}\myhl{143}{223}{246}{ the}\myhl{143}{223}{246}{ attack}\myhl{143}{223}{246}{,}\myhl{143}{223}{246}{ right}\myhl{143}{223}{246}{)}\myhl{143}{223}{246}{ Police}\myhl{143}{223}{246}{ described}\myhl{143}{223}{246}{ the}\myhl{143}{223}{246}{ attacker}\myhl{143}{223}{246}{ as}\myhl{143}{223}{246}{ white}\myhl{143}{223}{246}{,}\myhl{143}{223}{246}{ between}\myhl{143}{223}{246}{ }\myhl{143}{223}{246}{6}\myhl{143}{223}{246}{ft}\myhl{143}{223}{246}{ and}\myhl{143}{223}{246}{ }\myhl{143}{223}{246}{6}\myhl{143}{223}{246}{ft}\myhl{143}{223}{246}{ }\myhl{143}{223}{246}{2}\myhl{143}{223}{246}{in}\myhl{143}{223}{246}{ tall}\myhl{143}{223}{246}{,}\myhl{143}{223}{246}{ of}\myhl{143}{223}{246}{ an}\myhl{143}{223}{246}{ average}\myhl{143}{223}{246}{ to}\myhl{143}{223}{246}{ medium}\myhl{143}{223}{246}{ build}\myhl{143}{223}{246}{ with}\myhl{143}{223}{246}{ short}\myhl{143}{223}{246}{ light}\myhl{143}{223}{246}{ brown}\myhl{143}{223}{246}{ or}\myhl{143}{223}{246}{ dark}\myhl{143}{223}{246}{ blonde}\myhl{143}{223}{246}{ hair}\myhl{143}{223}{246}{ .} \dots\myhl{108}{213}{243}{ He}\myhl{108}{213}{243}{ had}\myhl{108}{213}{243}{ just}\myhl{108}{213}{243}{ left}\myhl{108}{213}{243}{ a}\myhl{108}{213}{243}{ family}\myhl{108}{213}{243}{ party}\myhl{108}{213}{243}{ at}\myhl{108}{213}{243}{ around}\myhl{108}{213}{243}{ }\myhl{108}{213}{243}{11}\myhl{108}{213}{243}{.}\myhl{108}{213}{243}{45}\myhl{108}{213}{243}{pm}\myhl{108}{213}{243}{ on}\myhl{108}{213}{243}{ Saturday}\myhl{108}{213}{243}{ when}\myhl{108}{213}{243}{ he}\myhl{108}{213}{243}{ approached}\myhl{108}{213}{243}{ the}\myhl{108}{213}{243}{ would}\myhl{108}{213}{243}{-be}\myhl{108}{213}{243}{ thief}\myhl{108}{213}{243}{ and}\myhl{108}{213}{243}{ was}\myhl{108}{213}{243}{ vicious}\myhl{108}{213}{243}{ly}\myhl{108}{213}{243}{ attacked}\myhl{108}{213}{243}{ in}\myhl{108}{213}{243}{ Park}\myhl{108}{213}{243}{es}\myhl{108}{213}{243}{ Way}\myhl{108}{213}{243}{,}\myhl{108}{213}{243}{ Blackburn}\myhl{108}{213}{243}{ .} \dots\myhl{164}{229}{247}{ Anyone}\myhl{164}{229}{247}{ with}\myhl{164}{229}{247}{ information}\myhl{164}{229}{247}{ about}\myhl{164}{229}{247}{ the}\myhl{164}{229}{247}{ incident}\myhl{164}{229}{247}{ which}\myhl{164}{229}{247}{ happened}\myhl{164}{229}{247}{ at}\myhl{164}{229}{247}{ }\myhl{164}{229}{247}{11}\myhl{164}{229}{247}{.}\myhl{164}{229}{247}{45}\myhl{164}{229}{247}{pm}\myhl{164}{229}{247}{ on}\myhl{164}{229}{247}{ Saturday}\myhl{164}{229}{247}{ should}\myhl{164}{229}{247}{ contact}\myhl{164}{229}{247}{ police}\myhl{164}{229}{247}{ on}\myhl{164}{229}{247}{ }\myhl{164}{229}{247}{101}\myhl{164}{229}{247}{ quoting}\myhl{164}{229}{247}{ log}\myhl{164}{229}{247}{ number}\myhl{164}{229}{247}{ }\myhl{164}{229}{247}{156}\myhl{164}{229}{247}{7}\myhl{164}{229}{247}{ of}\myhl{164}{229}{247}{ February}\myhl{164}{229}{247}{ }\myhl{164}{229}{247}{28}\myhl{164}{229}{247}{ or}\myhl{164}{229}{247}{ Cr}\myhl{164}{229}{247}{imest}\myhl{164}{229}{247}{oppers}\myhl{164}{229}{247}{ on}\myhl{164}{229}{247}{ }\myhl{164}{229}{247}{080}\myhl{164}{229}{247}{0}\myhl{164}{229}{247}{ }\myhl{164}{229}{247}{555}\myhl{164}{229}{247}{ }\myhl{164}{229}{247}{111}\myhl{164}{229}{247}{.}
\end{tcolorbox}
\begin{tcolorbox}[colback=white, , title=ESM (256 samples)]
\textbf{Context}:
\myhl{107}{213}{243}{ A}\myhl{107}{213}{243}{ have}\myhl{107}{213}{243}{-a}\myhl{107}{213}{243}{-go}\myhl{107}{213}{243}{-hero}\myhl{107}{213}{243}{ was}\myhl{107}{213}{243}{ so}\myhl{107}{213}{243}{ badly}\myhl{107}{213}{243}{ beaten}\myhl{107}{213}{243}{ and}\myhl{107}{213}{243}{ covered}\myhl{107}{213}{243}{ in}\myhl{107}{213}{243}{ blood}\myhl{107}{213}{243}{ after}\myhl{107}{213}{243}{ he}\myhl{107}{213}{243}{ challenged}\myhl{107}{213}{243}{ a}\myhl{107}{213}{243}{ car}\myhl{107}{213}{243}{ thief}\myhl{107}{213}{243}{ that}\myhl{107}{213}{243}{ his}\myhl{107}{213}{243}{ family}\myhl{107}{213}{243}{ thought}\myhl{107}{213}{243}{ he}\myhl{107}{213}{243}{ had}\myhl{107}{213}{243}{ been}\myhl{107}{213}{243}{ shot}\myhl{107}{213}{243}{.}\myhl{4}{184}{234}{ Danny}\myhl{4}{184}{234}{ Mason}\myhl{4}{184}{234}{,}\myhl{4}{184}{234}{ }\myhl{4}{184}{234}{37}\myhl{4}{184}{234}{,}\myhl{4}{184}{234}{ was}\myhl{4}{184}{234}{ repeatedly}\myhl{4}{184}{234}{ punched}\myhl{4}{184}{234}{ in}\myhl{4}{184}{234}{ the}\myhl{4}{184}{234}{ face}\myhl{4}{184}{234}{ and}\myhl{4}{184}{234}{ left}\myhl{4}{184}{234}{ with}\myhl{4}{184}{234}{ a}\myhl{4}{184}{234}{ fractured}\myhl{4}{184}{234}{ cheek}\myhl{4}{184}{234}{,}\myhl{4}{184}{234}{ broken}\myhl{4}{184}{234}{ nose}\myhl{4}{184}{234}{ and}\myhl{4}{184}{234}{ cut}\myhl{4}{184}{234}{ lip}\myhl{4}{184}{234}{ after}\myhl{4}{184}{234}{ he}\myhl{4}{184}{234}{ approached}\myhl{4}{184}{234}{ the}\myhl{4}{184}{234}{ would}\myhl{4}{184}{234}{-be}\myhl{4}{184}{234}{-th}\myhl{4}{184}{234}{ief}\myhl{4}{184}{234}{ in}\myhl{4}{184}{234}{ Blackburn}\myhl{4}{184}{234}{,}\myhl{4}{184}{234}{ Lanc}\myhl{4}{184}{234}{ashire}\myhl{4}{184}{234}{.}\myhl{0}{183}{234}{ He}\myhl{0}{183}{234}{ had}\myhl{0}{183}{234}{ just}\myhl{0}{183}{234}{ left}\myhl{0}{183}{234}{ a}\myhl{0}{183}{234}{ family}\myhl{0}{183}{234}{ party}\myhl{0}{183}{234}{ at}\myhl{0}{183}{234}{ around}\myhl{0}{183}{234}{ }\myhl{0}{183}{234}{11}\myhl{0}{183}{234}{.}\myhl{0}{183}{234}{45}\myhl{0}{183}{234}{pm}\myhl{0}{183}{234}{ on}\myhl{0}{183}{234}{ Saturday}\myhl{0}{183}{234}{ when}\myhl{0}{183}{234}{ he}\myhl{0}{183}{234}{ tried}\myhl{0}{183}{234}{ to}\myhl{0}{183}{234}{ stop}\myhl{0}{183}{234}{ the}\myhl{0}{183}{234}{ would}\myhl{0}{183}{234}{-be}\myhl{0}{183}{234}{-th}\myhl{0}{183}{234}{ief}\myhl{0}{183}{234}{ from}\myhl{0}{183}{234}{ stealing}\myhl{0}{183}{234}{ a}\myhl{0}{183}{234}{ number}\myhl{0}{183}{234}{ of}\myhl{0}{183}{234}{ cars}\myhl{0}{183}{234}{.} \dots\myhl{61}{200}{239}{ Danny}\myhl{61}{200}{239}{ Mason}\myhl{61}{200}{239}{,}\myhl{61}{200}{239}{ }\myhl{61}{200}{239}{37}\myhl{61}{200}{239}{,}\myhl{61}{200}{239}{ was}\myhl{61}{200}{239}{ repeatedly}\myhl{61}{200}{239}{ punched}\myhl{61}{200}{239}{ in}\myhl{61}{200}{239}{ the}\myhl{61}{200}{239}{ face}\myhl{61}{200}{239}{ and}\myhl{61}{200}{239}{ left}\myhl{61}{200}{239}{ with}\myhl{61}{200}{239}{ a}\myhl{61}{200}{239}{ fractured}\myhl{61}{200}{239}{ cheek}\myhl{61}{200}{239}{,}\myhl{61}{200}{239}{ broken}\myhl{61}{200}{239}{ nose}\myhl{61}{200}{239}{ and}\myhl{61}{200}{239}{ cut}\myhl{61}{200}{239}{ lip}\myhl{61}{200}{239}{ after}\myhl{61}{200}{239}{ challenging}\myhl{61}{200}{239}{ a}\myhl{61}{200}{239}{ car}\myhl{61}{200}{239}{ thief}\myhl{61}{200}{239}{ .} \dots\myhl{166}{230}{247}{ '}\myhl{166}{230}{247}{Danny}\myhl{166}{230}{247}{ had}\myhl{166}{230}{247}{ gone}\myhl{166}{230}{247}{ outside}\myhl{166}{230}{247}{ for}\myhl{166}{230}{247}{ a}\myhl{166}{230}{247}{ cigarette}\myhl{166}{230}{247}{ when}\myhl{166}{230}{247}{ he}\myhl{166}{230}{247}{ spotted}\myhl{166}{230}{247}{ a}\myhl{166}{230}{247}{ man}\myhl{166}{230}{247}{ who}\myhl{166}{230}{247}{ looked}\myhl{166}{230}{247}{ like}\myhl{166}{230}{247}{ he}\myhl{166}{230}{247}{ was}\myhl{166}{230}{247}{ trying}\myhl{166}{230}{247}{ to}\myhl{166}{230}{247}{ break}\myhl{166}{230}{247}{ into}\myhl{166}{230}{247}{ vehicles}\myhl{166}{230}{247}{.} \dots
\end{tcolorbox}
\textbf{Query}:
Summarize the article in up to three sentences.
\newline\newline\textbf{Response} (from \texttt{Llama-3.1-8B}):
\newline
\myhl{249}{237}{91}{A 37-year-old man, Danny Mason, was brutally attacked by a car thief in Blackburn, Lancashire, after he tried to stop the individual from stealing vehicles.} Mason suffered a fractured cheek, broken nose, and cut lip in the assault, which left him in so much pain that he cannot sleep or work. Police have released CCTV images of a suspect and are appealing for anyone with information to come forward to help identify the attacker.
\end{tcolorbox}
\relsize{1}
\caption{Visualized attributions for a random example from CNN DailyMail for \texttt{Llama-3.1-8B} with sentences as sources.}
\label{fig:cnn_dailymail_llama_3_1_8b_sentence}
\end{figure*}

\begin{figure*}[t]
\centering\definecolor{forestgreen}{RGB}{34, 139, 34}
\relsize{-1}
\begin{tcolorbox}[colback=white]
\begin{tcolorbox}[colback=white, colframe=forestgreen, title=\ours (general)]
\textbf{Context}:
 \dots\myhl{253}{254}{254}{1}\myhl{246}{252}{254}{6}\myhl{247}{252}{254}{.}\myhl{255}{255}{255}{}\myhl{255}{255}{255}{}\myhl{0}{183}{234}{Right}\myhl{0}{183}{234}{ On}\myhl{150}{225}{246}{!}\myhl{219}{245}{252}{ }\myhl{230}{248}{253}{ was}\myhl{255}{255}{255}{ an}\myhl{255}{255}{255}{ American}\myhl{255}{255}{255}{ te} \dots
\end{tcolorbox}
\begin{tcolorbox}[colback=white, , title=Average attention]
\textbf{Context}:
 \dots\myhl{254}{254}{254}{yle}\myhl{254}{254}{254}{ and}\myhl{254}{254}{254}{ culture}\myhl{254}{254}{254}{ published}\myhl{254}{254}{254}{ in}\myhl{126}{219}{244}{ the}\myhl{254}{254}{254}{ United}\myhl{254}{254}{254}{ States}\myhl{251}{253}{254}{.}\myhl{254}{254}{254}{ }\myhl{254}{254}{254}{ It} \dots\myhl{254}{254}{254}{1}\myhl{254}{254}{254}{9}\myhl{254}{254}{254}{2}\myhl{254}{254}{254}{6}\myhl{254}{254}{254}{ during}\myhl{177}{233}{248}{ the}\myhl{254}{254}{254}{ Har}\myhl{254}{254}{254}{lem}\myhl{254}{254}{254}{ Renaissance}\myhl{252}{254}{254}{.}\myhl{254}{254}{254}{ } \dots\myhl{254}{254}{254}{anz}\myhl{254}{254}{254}{ine}\myhl{254}{254}{254}{ three}\myhl{254}{254}{254}{ times}\myhl{254}{254}{254}{ (}\myhl{192}{237}{250}{2}\myhl{254}{254}{254}{0}\myhl{254}{254}{254}{0}\myhl{254}{254}{254}{8}\myhl{254}{254}{254}{-}\myhl{206}{241}{251}{2} \dots\myhl{254}{254}{254}{1}\myhl{254}{254}{254}{9}\myhl{254}{254}{254}{6}\myhl{254}{254}{254}{3}\myhl{254}{254}{254}{ into}\myhl{167}{230}{248}{ the}\myhl{254}{254}{254}{ early}\myhl{254}{254}{254}{ }\myhl{254}{254}{254}{1}\myhl{254}{254}{254}{9}\myhl{254}{254}{254}{7} \dots\myhl{254}{254}{254}{ the}\myhl{254}{254}{254}{ National}\myhl{254}{254}{254}{ Geographic}\myhl{254}{254}{254}{ Society}\myhl{254}{254}{254}{ in}\myhl{177}{233}{248}{ the}\myhl{254}{254}{254}{ United}\myhl{254}{254}{254}{ States}\myhl{251}{253}{254}{.}\myhl{254}{254}{254}{ }\myhl{254}{254}{254}{ The} \dots\myhl{254}{254}{254}{0}\myhl{254}{254}{254}{0}\myhl{254}{254}{254}{9}\myhl{254}{254}{254}{/}\myhl{254}{254}{254}{J}\myhl{0}{183}{234}{anu}\myhl{254}{254}{254}{ary}\myhl{254}{254}{254}{ }\myhl{254}{254}{254}{2}\myhl{254}{254}{254}{0}\myhl{254}{254}{254}{1} \dots
\end{tcolorbox}
\begin{tcolorbox}[colback=white, , title=ESM (256 samples)]
\textbf{Context}:
 \dots\myhl{255}{255}{255}{ was}\myhl{255}{255}{255}{ a}\myhl{255}{255}{255}{ month}\myhl{255}{255}{255}{ly}\myhl{255}{255}{255}{ print}\myhl{78}{205}{240}{ magazine}\myhl{255}{255}{255}{ dedicated}\myhl{147}{224}{246}{ to}\myhl{255}{255}{255}{ news}\myhl{255}{255}{255}{,}\myhl{255}{255}{255}{ comment}\myhl{246}{252}{254}{ary}\myhl{255}{255}{255}{,} \dots\myhl{255}{255}{255}{ circul}\myhl{255}{255}{255}{ated}\myhl{255}{255}{255}{ English}\myhl{255}{255}{255}{-}\myhl{255}{255}{255}{language}\myhl{188}{236}{249}{ month}\myhl{255}{255}{255}{ly}\myhl{255}{255}{255}{ magazine}\myhl{255}{255}{255}{ for}\myhl{255}{255}{255}{ the}\myhl{255}{255}{255}{ Asian} \dots\myhl{255}{255}{255}{ants}\myhl{255}{255}{255}{ and}\myhl{255}{255}{255}{ communities}\myhl{255}{255}{255}{ of}\myhl{255}{255}{255}{ color}\myhl{198}{239}{250}{.}\myhl{255}{255}{255}{ }\myhl{167}{230}{248}{ The}\myhl{255}{255}{255}{ magazine}\myhl{255}{255}{255}{ ended}\myhl{255}{255}{255}{ print}\myhl{255}{255}{255}{ issue}\myhl{255}{255}{255}{ in} \dots\myhl{255}{255}{255}{!!}\myhl{255}{255}{255}{ }\myhl{255}{255}{255}{ was}\myhl{255}{255}{255}{ an}\myhl{255}{255}{255}{ African}\myhl{144}{223}{246}{-}\myhl{255}{255}{255}{American}\myhl{255}{255}{255}{ literary}\myhl{255}{255}{255}{ magazine}\myhl{255}{255}{255}{ published}\myhl{255}{255}{255}{ in} \dots\myhl{255}{255}{255}{9}\myhl{255}{255}{255}{2}\myhl{255}{255}{255}{6}\myhl{255}{255}{255}{ during}\myhl{255}{255}{255}{ the}\myhl{146}{224}{246}{ Har}\myhl{255}{255}{255}{lem}\myhl{255}{255}{255}{ Renaissance}\myhl{255}{255}{255}{.}\myhl{255}{255}{255}{ }\myhl{255}{255}{255}{ The} \dots\myhl{255}{255}{255}{ Ne}\myhl{255}{255}{255}{ale}\myhl{255}{255}{255}{ Hur}\myhl{255}{255}{255}{ston}\myhl{255}{255}{255}{,}\myhl{192}{237}{250}{ A}\myhl{255}{255}{255}{aron}\myhl{255}{255}{255}{ Douglas}\myhl{255}{255}{255}{,}\myhl{255}{255}{255}{ John}\myhl{255}{255}{255}{ P} \dots\myhl{255}{255}{255}{ Ben}\myhl{255}{255}{255}{nett}\myhl{255}{255}{255}{,}\myhl{255}{255}{255}{ Lewis}\myhl{255}{255}{255}{ Grand}\myhl{190}{237}{249}{ison}\myhl{255}{255}{255}{ Alexander}\myhl{255}{255}{255}{,}\myhl{255}{255}{255}{ Coun}\myhl{255}{255}{255}{te}\myhl{255}{255}{255}{e} \dots\myhl{255}{255}{255}{ qu}\myhl{255}{255}{255}{arters}\myhl{255}{255}{255}{ burn}\myhl{255}{255}{255}{ed}\myhl{255}{255}{255}{ down}\myhl{146}{224}{246}{,}\myhl{255}{255}{255}{ and}\myhl{255}{255}{255}{ the}\myhl{255}{255}{255}{ magazine}\myhl{255}{255}{255}{ ended}\myhl{255}{255}{255}{.} \dots\myhl{255}{255}{255}{ F}\myhl{255}{255}{255}{anz}\myhl{255}{255}{255}{ine}\myhl{255}{255}{255}{ three}\myhl{255}{255}{255}{ times}\myhl{122}{217}{244}{ (}\myhl{255}{255}{255}{2}\myhl{255}{255}{255}{0}\myhl{255}{255}{255}{0}\myhl{255}{255}{255}{8}\myhl{255}{255}{255}{-}\myhl{255}{255}{255}{2}\myhl{255}{255}{255}{0}\myhl{255}{255}{255}{1}\myhl{178}{233}{248}{0}\myhl{255}{255}{255}{).}\myhl{255}{255}{255}{ }\myhl{255}{255}{255}{ The}\myhl{255}{255}{255}{ magazine}\myhl{255}{255}{255}{ ended} \dots\myhl{255}{255}{255}{ }\myhl{255}{255}{255}{ The}\myhl{255}{255}{255}{ magazine}\myhl{255}{255}{255}{ was}\myhl{255}{255}{255}{ published}\myhl{126}{218}{244}{ on}\myhl{255}{255}{255}{ a}\myhl{255}{255}{255}{ bi}\myhl{255}{255}{255}{week}\myhl{255}{255}{255}{ly}\myhl{255}{255}{255}{ basis} \dots\myhl{255}{255}{255}{1}\myhl{255}{255}{255}{6}\myhl{255}{255}{255}{.}\myhl{255}{255}{255}{ }\myhl{226}{246}{252}{ Fred}\myhl{193}{237}{250}{rik}\myhl{205}{241}{251}{ke}\myhl{255}{255}{255}{ M}\myhl{255}{255}{255}{or}\myhl{255}{255}{255}{ck}\myhl{255}{255}{255}{ took} \dots\myhl{254}{254}{254}{9}\myhl{255}{255}{255}{1}\myhl{255}{255}{255}{6}\myhl{255}{255}{255}{.}\myhl{255}{255}{255}{ }\myhl{76}{205}{240}{ The}\myhl{255}{255}{255}{ magazine}\myhl{255}{255}{255}{ ended}\myhl{255}{255}{255}{ its}\myhl{255}{255}{255}{ publication}\myhl{158}{227}{247}{ in}\myhl{255}{255}{255}{ }\myhl{255}{255}{255}{1}\myhl{255}{255}{255}{9}\myhl{255}{255}{255}{2}\myhl{255}{255}{255}{7} \dots\myhl{255}{255}{255}{2}\myhl{255}{255}{255}{ and}\myhl{255}{255}{255}{ }\myhl{255}{255}{255}{1}\myhl{255}{255}{255}{9}\myhl{192}{237}{249}{7}\myhl{255}{255}{255}{5}\myhl{255}{255}{255}{ by}\myhl{255}{255}{255}{ Cal}\myhl{255}{255}{255}{vin}\myhl{255}{255}{255}{ Thomas} \dots\myhl{255}{255}{255}{ Goth}\myhl{255}{255}{255}{ic}\myhl{255}{255}{255}{ Castle}\myhl{255}{255}{255}{ Publishing}\myhl{255}{255}{255}{ Company}\myhl{40}{194}{237}{,}\myhl{255}{255}{255}{ distributed}\myhl{255}{255}{255}{ by}\myhl{255}{255}{255}{ K}\myhl{255}{255}{255}{able}\myhl{255}{255}{255}{ News} \dots\myhl{255}{255}{255}{ Ken}\myhl{255}{255}{255}{ Be}\myhl{255}{255}{255}{ale}\myhl{255}{255}{255}{ edited}\myhl{255}{255}{255}{ the}\myhl{199}{239}{250}{ first}\myhl{255}{255}{255}{ three}\myhl{255}{255}{255}{ issues}\myhl{255}{255}{255}{.}\myhl{255}{255}{255}{ }\myhl{255}{255}{255}{ W} \dots\myhl{255}{255}{255}{ }\myhl{255}{255}{255}{1}\myhl{255}{255}{255}{9}\myhl{255}{255}{255}{6}\myhl{255}{255}{255}{3}\myhl{128}{219}{244}{ into}\myhl{255}{255}{255}{ the}\myhl{255}{255}{255}{ early}\myhl{255}{255}{255}{ }\myhl{255}{255}{255}{1}\myhl{255}{255}{255}{9} \dots\myhl{255}{255}{255}{ Although}\myhl{255}{255}{255}{ promoted}\myhl{255}{255}{255}{ and}\myhl{255}{255}{255}{ sold}\myhl{255}{255}{255}{ as}\myhl{158}{228}{247}{ a}\myhl{248}{253}{254}{ "}\myhl{255}{255}{255}{mon}\myhl{255}{255}{255}{ster}\myhl{255}{255}{255}{ magazine}\myhl{255}{255}{255}{,"} \dots\myhl{255}{255}{255}{ and}\myhl{255}{255}{255}{ f}\myhl{255}{255}{255}{anz}\myhl{255}{255}{255}{ines}\myhl{255}{255}{255}{.}\myhl{174}{232}{248}{ }\myhl{255}{255}{255}{ Its}\myhl{255}{255}{255}{ advert}\myhl{255}{255}{255}{ising}\myhl{255}{255}{255}{ pages}\myhl{255}{255}{255}{ sold} \dots\myhl{255}{255}{255}{ issue}\myhl{255}{255}{255}{ was}\myhl{255}{255}{255}{ published}\myhl{255}{255}{255}{ in}\myhl{248}{253}{254}{ Spring}\myhl{202}{240}{250}{ }\myhl{255}{255}{255}{1}\myhl{255}{255}{255}{9}\myhl{197}{238}{250}{9}\myhl{255}{255}{255}{9}\myhl{255}{255}{255}{.}\myhl{255}{255}{255}{ }\myhl{255}{255}{255}{ Reg}\myhl{255}{255}{255}{ular}\myhl{255}{255}{255}{ publication}\myhl{255}{255}{255}{ of}\myhl{255}{255}{255}{ the}\myhl{255}{255}{255}{ magazine}\myhl{106}{213}{243}{ ended}\myhl{255}{255}{255}{ in}\myhl{255}{255}{255}{ December}\myhl{255}{255}{255}{ }\myhl{255}{255}{255}{2}\myhl{255}{255}{255}{0}\myhl{255}{255}{255}{0}\myhl{170}{231}{248}{9}\myhl{255}{255}{255}{,}\myhl{255}{255}{255}{ and}\myhl{255}{255}{255}{ the}\myhl{255}{255}{255}{ name}\myhl{255}{255}{255}{ was}\myhl{255}{255}{255}{ re}\myhl{255}{255}{255}{used}\myhl{193}{237}{250}{ for}\myhl{255}{255}{255}{ a}\myhl{255}{255}{255}{ b}\myhl{255}{255}{255}{ian}\myhl{154}{226}{246}{n}\myhl{255}{255}{255}{ual}\myhl{255}{255}{255}{ news}\myhl{255}{255}{255}{stand}\myhl{255}{255}{255}{ publication}\myhl{255}{255}{255}{.} \dots\myhl{255}{255}{255}{.}\myhl{255}{255}{255}{S}\myhl{255}{255}{255}{.}\myhl{255}{255}{255}{ "}\myhl{255}{255}{255}{Time}\myhl{185}{235}{249}{"}\myhl{255}{255}{255}{ magazine}\myhl{255}{255}{255}{.}\myhl{255}{255}{255}{ }\myhl{255}{255}{255}{ It}\myhl{255}{255}{255}{ was} \dots\myhl{245}{252}{254}{ London}\myhl{255}{255}{255}{.}\myhl{255}{255}{255}{ }\myhl{255}{255}{255}{ The}\myhl{255}{255}{255}{ magazine}\myhl{188}{236}{249}{ ended}\myhl{139}{222}{245}{ its}\myhl{255}{255}{255}{ run}\myhl{255}{255}{255}{ by}\myhl{255}{255}{255}{ eventually}\myhl{255}{255}{255}{ being}\myhl{255}{255}{255}{ am} \dots\myhl{255}{255}{255}{-}\myhl{255}{255}{255}{form}\myhl{255}{255}{255}{ feature}\myhl{255}{255}{255}{ stories}\myhl{255}{255}{255}{,}\myhl{90}{208}{241}{ phot}\myhl{255}{255}{255}{ography}\myhl{255}{255}{255}{,}\myhl{255}{255}{255}{ and}\myhl{255}{255}{255}{ illustr}\myhl{255}{255}{255}{ations} \dots\myhl{255}{255}{255}{itch}\myhl{255}{255}{255}{f}\myhl{255}{255}{255}{ork}\myhl{255}{255}{255}{'}\myhl{255}{255}{255}{s}\myhl{99}{211}{242}{ online}\myhl{255}{255}{255}{ content}\myhl{255}{255}{255}{.}\myhl{255}{255}{255}{ }\myhl{255}{255}{255}{ The}\myhl{255}{255}{255}{ magazine} \dots\myhl{255}{255}{255}{1}\myhl{255}{255}{255}{ issues}\myhl{255}{255}{255}{ }\myhl{255}{255}{255}{I}\myhl{255}{255}{255}{n}\myhl{130}{220}{245}{ November}\myhl{255}{255}{255}{ }\myhl{255}{255}{255}{2}\myhl{255}{255}{255}{0}\myhl{255}{255}{255}{1}\myhl{255}{255}{255}{6} \dots\myhl{255}{255}{255}{ an}\myhl{255}{255}{255}{ American}\myhl{255}{255}{255}{ te}\myhl{255}{255}{255}{en}\myhl{255}{255}{255}{ magazine}\myhl{182}{234}{249}{ first}\myhl{255}{255}{255}{ published}\myhl{255}{255}{255}{ by}\myhl{255}{255}{255}{ the}\myhl{255}{255}{255}{ Lau}\myhl{255}{255}{255}{fer}\myhl{255}{255}{255}{ Company}\myhl{255}{255}{255}{ in}\myhl{25}{190}{236}{ }\myhl{255}{255}{255}{1}\myhl{255}{255}{255}{9}\myhl{0}{183}{234}{7}\myhl{255}{255}{255}{2}\myhl{200}{239}{250}{ with}\myhl{255}{255}{255}{ editor}\myhl{255}{255}{255}{/}\myhl{255}{255}{255}{cre}\myhl{255}{255}{255}{ator}\myhl{255}{255}{255}{ Jud} \dots\myhl{255}{255}{255}{.}\myhl{255}{255}{255}{ }\myhl{255}{255}{255}{2}\myhl{255}{255}{255}{0}\myhl{255}{255}{255}{1}\myhl{199}{239}{250}{1}\myhl{255}{255}{255}{ and}\myhl{255}{255}{255}{ focused}\myhl{255}{255}{255}{ on}\myhl{255}{255}{255}{ African}\myhl{255}{255}{255}{-} \dots\myhl{255}{255}{255}{ rail}\myhl{224}{246}{252}{ro}\myhl{255}{255}{255}{ads}\myhl{255}{255}{255}{ and}\myhl{255}{255}{255}{ short}\myhl{188}{236}{249}{ line}\myhl{255}{255}{255}{ rail}\myhl{255}{255}{255}{ro}\myhl{255}{255}{255}{ads}\myhl{255}{255}{255}{,}\myhl{255}{255}{255}{ and} \dots\myhl{255}{255}{255}{ and}\myhl{255}{255}{255}{ model}\myhl{255}{255}{255}{ing}\myhl{255}{255}{255}{.}\myhl{255}{255}{255}{ }\myhl{186}{235}{249}{ The}\myhl{255}{255}{255}{ magazine}\myhl{106}{213}{243}{ ended}\myhl{255}{255}{255}{ publication}\myhl{255}{255}{255}{ in}\myhl{255}{255}{255}{ August}\myhl{255}{255}{255}{ }\myhl{255}{255}{255}{2} \dots
\end{tcolorbox}
\textbf{Query}:
The publication of which magazine ended first, Right On! or Castle of Frankenstein?
\newline\newline\textbf{Response} (from \texttt{Phi-3.5-Mini}):
\newline
The publication of Right On! ended first. \myhl{249}{237}{91}{Right On!} was a teen magazine that continued publishing until around 2011. On the other hand, Castle of Frankenstein, an American horror, science fiction, and fantasy film magazine, ended its publication in 1975. Therefore, Right On! ended its publication before Castle of Frankenstein.
\end{tcolorbox}
\relsize{1}
\caption{Visualized attributions for a random example from Hotpot QA for \texttt{Phi-3.5-Mini} with tokens as sources.}
\label{fig:hotpot_qa_passages_phi_3_5_mini_token}
\end{figure*}

\begin{figure*}[t]
\centering\definecolor{forestgreen}{RGB}{34, 139, 34}
\relsize{-1}
\begin{tcolorbox}[colback=white]
\begin{tcolorbox}[colback=white, colframe=forestgreen, title=\ours (general)]
\textbf{Context}:
 \dots\myhl{253}{254}{254}{ LG}\myhl{253}{254}{254}{V}\myhl{254}{254}{254}{ Nord}\myhl{251}{254}{254}{.}\myhl{204}{240}{250}{Paris}\myhl{189}{236}{249}{ Charles}\myhl{246}{252}{254}{ de}\myhl{245}{252}{254}{ Gaul}\myhl{247}{253}{254}{le}\myhl{223}{246}{252}{ Airport}\myhl{252}{254}{254}{ (} \dots\myhl{255}{255}{255}{ to}\myhl{254}{254}{254}{ various}\myhl{244}{251}{254}{ others}\myhl{230}{248}{253}{ cities}\myhl{254}{254}{254}{ in}\myhl{202}{240}{250}{ France}\myhl{249}{253}{254}{,}\myhl{205}{241}{251}{ as}\myhl{214}{243}{251}{ well}\myhl{212}{242}{251}{ as}\myhl{255}{255}{255}{ to}\myhl{0}{183}{234}{ Belgium}\myhl{255}{255}{255}{.}
\end{tcolorbox}
\begin{tcolorbox}[colback=white, , title=Average attention]
\textbf{Context}:
 \dots\myhl{251}{253}{254}{ LG}\myhl{247}{252}{254}{V}\myhl{249}{253}{254}{ Nord}\myhl{229}{247}{252}{.}\myhl{208}{242}{251}{Paris}\myhl{181}{234}{249}{ Charles}\myhl{214}{243}{251}{ de}\myhl{226}{247}{252}{ Gaul}\myhl{203}{240}{250}{le}\myhl{196}{238}{250}{ Airport}\myhl{222}{245}{252}{ (}\myhl{244}{251}{254}{French}\myhl{247}{253}{254}{:}\myhl{248}{253}{254}{ "}\myhl{251}{254}{254}{A} \dots\myhl{249}{253}{254}{PG}\myhl{231}{248}{253}{ ),}\myhl{239}{250}{253}{ also}\myhl{233}{248}{253}{ known}\myhl{229}{247}{252}{ as}\myhl{194}{238}{250}{ Ro}\myhl{213}{243}{251}{issy}\myhl{180}{234}{249}{ Airport}\myhl{226}{246}{252}{ (}\myhl{245}{252}{254}{name}\myhl{248}{253}{254}{ of}\myhl{249}{253}{254}{ the}\myhl{249}{253}{254}{ local} \dots\myhl{249}{253}{254}{ueling}\myhl{248}{253}{254}{ and}\myhl{245}{252}{254}{ Norwegian}\myhl{252}{254}{254}{ Air}\myhl{249}{253}{254}{ Shuttle}\myhl{180}{234}{249}{.}\myhl{216}{244}{251}{Ro}\myhl{213}{243}{251}{issy}\myhl{231}{248}{253}{-en}\myhl{241}{251}{253}{-F}\myhl{243}{251}{254}{rance} \dots\myhl{241}{251}{253}{199}\myhl{241}{251}{253}{4}\myhl{237}{250}{253}{.}\myhl{240}{251}{253}{ }\myhl{240}{250}{253}{ It}\myhl{203}{240}{250}{ connects}\myhl{237}{250}{253}{ the}\myhl{222}{246}{252}{ airport}\myhl{213}{243}{251}{ to}\myhl{213}{243}{251}{ Paris}\myhl{205}{241}{251}{ and}\myhl{208}{242}{251}{ to}\myhl{198}{239}{250}{ various}\myhl{210}{242}{251}{ others}\myhl{162}{229}{247}{ cities}\myhl{213}{243}{251}{ in}\myhl{166}{230}{247}{ France}\myhl{204}{240}{250}{,}\myhl{226}{247}{252}{ as}\myhl{186}{235}{249}{ well}\myhl{177}{233}{248}{ as}\myhl{199}{239}{250}{ to}\myhl{0}{183}{234}{ Belgium}\myhl{133}{220}{245}{.}
\end{tcolorbox}
\begin{tcolorbox}[colback=white, , title=ESM (256 samples)]
\textbf{Context}:
 \dots\myhl{255}{255}{255}{,}\myhl{255}{255}{255}{ as}\myhl{246}{252}{254}{ well}\myhl{255}{255}{255}{ as}\myhl{231}{248}{253}{ to}\myhl{0}{183}{234}{ Belgium}\myhl{255}{255}{255}{.}
\end{tcolorbox}
\textbf{Query}:
The Roissy Airport connects to Paris and cities in what countries?
\newline\newline\textbf{Response} (from \texttt{Llama-3.1-8B}):
\newline
\myhl{249}{237}{91}{The Roissy Airport, also known as Paris Charles de Gaulle Airport, connects to Paris and cities in France, as well as to Belgium.}
\end{tcolorbox}
\relsize{1}
\caption{Visualized attributions for a random example from Hotpot QA for \texttt{Llama-3.1-8B} with tokens as sources.}
\label{fig:hotpot_qa_passages_llama_3_1_8b_token}
\end{figure*}

\begin{figure*}[t]
\centering\definecolor{forestgreen}{RGB}{34, 139, 34}
\relsize{-2}
\begin{tcolorbox}[colback=white]
\begin{tcolorbox}[colback=white, colframe=forestgreen, title=\ours (general)]
\textbf{Context}:
\myhl{71}{203}{240}{M}\myhl{71}{203}{240}{anga}\myhl{71}{203}{240}{ Life}\myhl{71}{203}{240}{ (}\myhl{71}{203}{240}{ma}\myhl{71}{203}{240}{n}\myhl{71}{203}{240}{ga}\myhl{71}{203}{240}{ra}\myhl{71}{203}{240}{i}\myhl{71}{203}{240}{hu}\myhl{71}{203}{240}{ ,}\myhl{71}{203}{240}{ M}\myhl{71}{203}{240}{anga}\myhl{71}{203}{240}{ Ra}\myhl{71}{203}{240}{if}\myhl{71}{203}{240}{u}\myhl{71}{203}{240}{ )}\myhl{71}{203}{240}{ is}\myhl{71}{203}{240}{ a}\myhl{71}{203}{240}{ m}\myhl{71}{203}{240}{anga}\myhl{71}{203}{240}{ magazine}\myhl{71}{203}{240}{ published}\myhl{71}{203}{240}{ month}\myhl{71}{203}{240}{ly}\myhl{71}{203}{240}{ by}\myhl{71}{203}{240}{ T}\myhl{71}{203}{240}{akes}\myhl{71}{203}{240}{h}\myhl{71}{203}{240}{ob}\myhl{71}{203}{240}{o}\myhl{71}{203}{240}{ in}\myhl{71}{203}{240}{ Japan}\myhl{71}{203}{240}{ since}\myhl{71}{203}{240}{ the}\myhl{71}{203}{240}{ November}\myhl{71}{203}{240}{ }\myhl{71}{203}{240}{1}\myhl{71}{203}{240}{9}\myhl{71}{203}{240}{8}\myhl{71}{203}{240}{4}\myhl{71}{203}{240}{ issue}\myhl{71}{203}{240}{ (}\myhl{71}{203}{240}{pub}\myhl{71}{203}{240}{lished}\myhl{71}{203}{240}{ in}\myhl{71}{203}{240}{ October}\myhl{71}{203}{240}{ }\myhl{71}{203}{240}{1}\myhl{71}{203}{240}{9}\myhl{71}{203}{240}{8}\myhl{71}{203}{240}{4}\myhl{71}{203}{240}{).} \dots\myhl{183}{234}{249}{ Its}\myhl{183}{234}{249}{ original}\myhl{183}{234}{249}{ title}\myhl{183}{234}{249}{ was}\myhl{183}{234}{249}{ G}\myhl{183}{234}{249}{ag}\myhl{183}{234}{249}{ da}\myhl{183}{234}{249}{ (}\myhl{183}{234}{249}{gi}\myhl{183}{234}{249}{ya}\myhl{183}{234}{249}{gu}\myhl{183}{234}{249}{da}\myhl{183}{234}{249}{ ,}\myhl{183}{234}{249}{ Gy}\myhl{183}{234}{249}{agu}\myhl{183}{234}{249}{ da}\myhl{183}{234}{249}{ )}\myhl{183}{234}{249}{ ,}\myhl{183}{234}{249}{ and}\myhl{183}{234}{249}{ the}\myhl{183}{234}{249}{ change}\myhl{183}{234}{249}{ to}\myhl{183}{234}{249}{ "}\myhl{183}{234}{249}{M}\myhl{183}{234}{249}{anga}\myhl{183}{234}{249}{ Life}\myhl{183}{234}{249}{"}\myhl{183}{234}{249}{ was}\myhl{183}{234}{249}{ made}\myhl{183}{234}{249}{ to}\myhl{183}{234}{249}{ better}\myhl{183}{234}{249}{ comp}\myhl{183}{234}{249}{ete}\myhl{183}{234}{249}{ with}\myhl{183}{234}{249}{ "}\myhl{183}{234}{249}{M}\myhl{183}{234}{249}{anga}\myhl{183}{234}{249}{ Time}\myhl{183}{234}{249}{",}\myhl{183}{234}{249}{ a}\myhl{183}{234}{249}{ rival}\myhl{183}{234}{249}{ magazine}\myhl{183}{234}{249}{ published}\myhl{183}{234}{249}{ by}\myhl{183}{234}{249}{ Hou}\myhl{183}{234}{249}{b}\myhl{183}{234}{249}{un}\myhl{183}{234}{249}{sha}\myhl{183}{234}{249}{.} \dots\myhl{150}{225}{246}{ Most}\myhl{150}{225}{246}{ of}\myhl{150}{225}{246}{ the}\myhl{150}{225}{246}{ series}\myhl{150}{225}{246}{ appearing}\myhl{150}{225}{246}{ in}\myhl{150}{225}{246}{ the}\myhl{150}{225}{246}{ magazine}\myhl{150}{225}{246}{ use}\myhl{150}{225}{246}{ the}\myhl{150}{225}{246}{ y}\myhl{150}{225}{246}{onk}\myhl{150}{225}{246}{oma}\myhl{150}{225}{246}{ format}\myhl{150}{225}{246}{.} \dots\myhl{198}{239}{250}{ The}\myhl{198}{239}{250}{ magazine}\myhl{198}{239}{250}{ is}\myhl{198}{239}{250}{ released}\myhl{198}{239}{250}{ month}\myhl{198}{239}{250}{ly}\myhl{198}{239}{250}{ on}\myhl{198}{239}{250}{ the}\myhl{198}{239}{250}{ }\myhl{198}{239}{250}{1}\myhl{198}{239}{250}{7}\myhl{198}{239}{250}{th}\myhl{198}{239}{250}{,}\myhl{198}{239}{250}{ though}\myhl{198}{239}{250}{ it}\myhl{198}{239}{250}{ sometimes}\myhl{198}{239}{250}{ appears}\myhl{198}{239}{250}{ on}\myhl{198}{239}{250}{ sh}\myhl{198}{239}{250}{el}\myhl{198}{239}{250}{ves}\myhl{198}{239}{250}{ slightly}\myhl{198}{239}{250}{ before}\myhl{198}{239}{250}{ or}\myhl{198}{239}{250}{ after}\myhl{198}{239}{250}{ that}\myhl{198}{239}{250}{,}\myhl{198}{239}{250}{ depending}\myhl{198}{239}{250}{ on}\myhl{198}{239}{250}{ speed}\myhl{198}{239}{250}{ of}\myhl{198}{239}{250}{ actual}\myhl{198}{239}{250}{ distribution}\myhl{198}{239}{250}{.} \dots\myhl{191}{237}{249}{Den}\myhl{191}{237}{249}{ge}\myhl{191}{237}{249}{ki}\myhl{191}{237}{249}{ G}\myhl{191}{237}{249}{'}\myhl{191}{237}{249}{s}\myhl{191}{237}{249}{ Com}\myhl{191}{237}{249}{ic}\myhl{191}{237}{249}{ (}\myhl{191}{237}{249}{Dian }\myhl{191}{237}{249}{Ji }\myhl{191}{237}{249}{Ji }\myhl{191}{237}{249}{Ji }\myhl{191}{237}{249}{G}\myhl{191}{237}{249}{'}\myhl{191}{237}{249}{s}\myhl{191}{237}{249}{ko}\myhl{191}{237}{249}{mi}\myhl{191}{237}{249}{tsu}\myhl{191}{237}{249}{ku}\myhl{191}{237}{249}{ ,}\myhl{191}{237}{249}{ Den}\myhl{191}{237}{249}{ge}\myhl{191}{237}{249}{ki}\myhl{191}{237}{249}{ J}\myhl{191}{237}{249}{i}\myhl{191}{237}{249}{zu}\myhl{191}{237}{249}{ Kom}\myhl{191}{237}{249}{ik}\myhl{191}{237}{249}{ku}\myhl{191}{237}{249}{ )}\myhl{191}{237}{249}{ is}\myhl{191}{237}{249}{ a}\myhl{191}{237}{249}{ Japanese}\myhl{191}{237}{249}{ seinen}\myhl{191}{237}{249}{ m}\myhl{191}{237}{249}{anga}\myhl{191}{237}{249}{ magazine}\myhl{191}{237}{249}{ published}\myhl{191}{237}{249}{ by}\myhl{191}{237}{249}{ ASCII}\myhl{191}{237}{249}{ Media}\myhl{191}{237}{249}{ Works}\myhl{191}{237}{249}{.} \dots\myhl{0}{183}{234}{ The}\myhl{0}{183}{234}{ magazine}\myhl{0}{183}{234}{ was}\myhl{0}{183}{234}{ first}\myhl{0}{183}{234}{ published}\myhl{0}{183}{234}{ digit}\myhl{0}{183}{234}{ally}\myhl{0}{183}{234}{ on}\myhl{0}{183}{234}{ August}\myhl{0}{183}{234}{ }\myhl{0}{183}{234}{9}\myhl{0}{183}{234}{,}\myhl{0}{183}{234}{ }\myhl{0}{183}{234}{2}\myhl{0}{183}{234}{0}\myhl{0}{183}{234}{1}\myhl{0}{183}{234}{2}\myhl{0}{183}{234}{ with}\myhl{0}{183}{234}{ volume}\myhl{0}{183}{234}{ }\myhl{0}{183}{234}{0}\myhl{0}{183}{234}{,}\myhl{0}{183}{234}{ and}\myhl{0}{183}{234}{ started}\myhl{0}{183}{234}{ month}\myhl{0}{183}{234}{ly}\myhl{0}{183}{234}{ publication}\myhl{0}{183}{234}{ with}\myhl{0}{183}{234}{ the}\myhl{0}{183}{234}{ following}\myhl{0}{183}{234}{ issue}\myhl{0}{183}{234}{ released}\myhl{0}{183}{234}{ on}\myhl{0}{183}{234}{ October}\myhl{0}{183}{234}{ }\myhl{0}{183}{234}{1}\myhl{0}{183}{234}{5}\myhl{0}{183}{234}{,}\myhl{0}{183}{234}{ }\myhl{0}{183}{234}{2}\myhl{0}{183}{234}{0}\myhl{0}{183}{234}{1}\myhl{0}{183}{234}{2}\myhl{0}{183}{234}{.} \dots\myhl{192}{237}{249}{Life}\myhl{192}{237}{249}{ was}\myhl{192}{237}{249}{ an}\myhl{192}{237}{249}{ American}\myhl{192}{237}{249}{ magazine}\myhl{192}{237}{249}{ that}\myhl{192}{237}{249}{ ran}\myhl{192}{237}{249}{ week}\myhl{192}{237}{249}{ly}\myhl{192}{237}{249}{ from}\myhl{192}{237}{249}{ }\myhl{192}{237}{249}{1}\myhl{192}{237}{249}{8}\myhl{192}{237}{249}{8}\myhl{192}{237}{249}{3}\myhl{192}{237}{249}{ to}\myhl{192}{237}{249}{ }\myhl{192}{237}{249}{1}\myhl{192}{237}{249}{9}\myhl{192}{237}{249}{3}\myhl{192}{237}{249}{6}\myhl{192}{237}{249}{ as}\myhl{192}{237}{249}{ a}\myhl{192}{237}{249}{ humor}\myhl{192}{237}{249}{ magazine}\myhl{192}{237}{249}{ with}\myhl{192}{237}{249}{ limited}\myhl{192}{237}{249}{ circul}\myhl{192}{237}{249}{ation}\myhl{192}{237}{249}{.}\myhl{192}{237}{249}{ "}\myhl{180}{234}{249}{ Time}\myhl{180}{234}{249}{"}\myhl{180}{234}{249}{ owner}\myhl{180}{234}{249}{ Henry}\myhl{180}{234}{249}{ Lu}\myhl{180}{234}{249}{ce}\myhl{180}{234}{249}{ bought}\myhl{180}{234}{249}{ the}\myhl{180}{234}{249}{ magazine}\myhl{180}{234}{249}{ in}\myhl{180}{234}{249}{ }\myhl{180}{234}{249}{1}\myhl{180}{234}{249}{9}\myhl{180}{234}{249}{3}\myhl{180}{234}{249}{6}\myhl{180}{234}{249}{,}\myhl{180}{234}{249}{ sole}\myhl{180}{234}{249}{ly}\myhl{180}{234}{249}{ so}\myhl{180}{234}{249}{ that}\myhl{180}{234}{249}{ he}\myhl{180}{234}{249}{ could}\myhl{180}{234}{249}{ ac}\myhl{180}{234}{249}{quire}\myhl{180}{234}{249}{ the}\myhl{180}{234}{249}{ rights}\myhl{180}{234}{249}{ to}\myhl{180}{234}{249}{ its}\myhl{180}{234}{249}{ name}\myhl{180}{234}{249}{,}\myhl{180}{234}{249}{ and}\myhl{180}{234}{249}{ launched}\myhl{180}{234}{249}{ a}\myhl{180}{234}{249}{ major}\myhl{180}{234}{249}{ week}\myhl{180}{234}{249}{ly}\myhl{180}{234}{249}{ news}\myhl{180}{234}{249}{ magazine}\myhl{180}{234}{249}{ with}\myhl{180}{234}{249}{ a}\myhl{180}{234}{249}{ strong}\myhl{180}{234}{249}{ emphas}\myhl{180}{234}{249}{is}\myhl{180}{234}{249}{ on}\myhl{180}{234}{249}{ photo}\myhl{180}{234}{249}{j}\myhl{180}{234}{249}{ournal}\myhl{180}{234}{249}{ism}\myhl{180}{234}{249}{.} \dots\myhl{187}{235}{249}{ "}\myhl{187}{235}{249}{Life}\myhl{187}{235}{249}{"}\myhl{187}{235}{249}{ was}\myhl{187}{235}{249}{ published}\myhl{187}{235}{249}{ week}\myhl{187}{235}{249}{ly}\myhl{187}{235}{249}{ until}\myhl{187}{235}{249}{ }\myhl{187}{235}{249}{1}\myhl{187}{235}{249}{9}\myhl{187}{235}{249}{7}\myhl{187}{235}{249}{2}\myhl{187}{235}{249}{,}\myhl{187}{235}{249}{ as}\myhl{187}{235}{249}{ an}\myhl{187}{235}{249}{ inter}\myhl{187}{235}{249}{mitt}\myhl{187}{235}{249}{ent}\myhl{187}{235}{249}{ "}\myhl{187}{235}{249}{special}\myhl{187}{235}{249}{"}\myhl{187}{235}{249}{ until}\myhl{187}{235}{249}{ }\myhl{187}{235}{249}{1}\myhl{187}{235}{249}{9}\myhl{187}{235}{249}{7}\myhl{187}{235}{249}{8}\myhl{187}{235}{249}{,}\myhl{187}{235}{249}{ and}\myhl{187}{235}{249}{ as}\myhl{187}{235}{249}{ a}\myhl{187}{235}{249}{ month}\myhl{187}{235}{249}{ly}\myhl{187}{235}{249}{ from}\myhl{187}{235}{249}{ }\myhl{187}{235}{249}{1}\myhl{187}{235}{249}{9}\myhl{187}{235}{249}{7}\myhl{187}{235}{249}{8}\myhl{187}{235}{249}{ to}\myhl{187}{235}{249}{ }\myhl{187}{235}{249}{2}\myhl{187}{235}{249}{0}\myhl{187}{235}{249}{0}\myhl{187}{235}{249}{0}\myhl{187}{235}{249}{.} \dots\myhl{202}{240}{250}{ F}\myhl{202}{240}{250}{ounded}\myhl{202}{240}{250}{ in}\myhl{202}{240}{250}{ }\myhl{202}{240}{250}{1}\myhl{202}{240}{250}{9}\myhl{202}{240}{250}{9}\myhl{202}{240}{250}{3}\myhl{202}{240}{250}{,}\myhl{202}{240}{250}{ the}\myhl{202}{240}{250}{ magazine}\myhl{202}{240}{250}{ is}\myhl{202}{240}{250}{ published}\myhl{202}{240}{250}{ eight}\myhl{202}{240}{250}{ times}\myhl{202}{240}{250}{ ann}\myhl{202}{240}{250}{ually}\myhl{202}{240}{250}{ and}\myhl{202}{240}{250}{ covers}\myhl{202}{240}{250}{ the}\myhl{202}{240}{250}{ events}\myhl{202}{240}{250}{,}\myhl{202}{240}{250}{ people}\myhl{202}{240}{250}{,}\myhl{202}{240}{250}{ history}\myhl{202}{240}{250}{ and}\myhl{202}{240}{250}{ places}\myhl{202}{240}{250}{ of}\myhl{202}{240}{250}{ New}\myhl{202}{240}{250}{port}\myhl{202}{240}{250}{ County}\myhl{202}{240}{250}{.} \dots\myhl{140}{222}{245}{ New}\myhl{140}{222}{245}{port}\myhl{140}{222}{245}{ Life}\myhl{140}{222}{245}{ Magazine}\myhl{140}{222}{245}{ is}\myhl{140}{222}{245}{ located}\myhl{140}{222}{245}{ at}\myhl{140}{222}{245}{ }\myhl{140}{222}{245}{1}\myhl{140}{222}{245}{0}\myhl{140}{222}{245}{1}\myhl{140}{222}{245}{ Mal}\myhl{140}{222}{245}{bone}\myhl{140}{222}{245}{ Road}\myhl{140}{222}{245}{ in}\myhl{140}{222}{245}{ the}\myhl{140}{222}{245}{ New}\myhl{140}{222}{245}{port}\myhl{140}{222}{245}{ Daily}\myhl{140}{222}{245}{ News}\myhl{140}{222}{245}{ building}\myhl{140}{222}{245}{.} \dots\myhl{181}{234}{249}{S}\myhl{181}{234}{249}{yl}\myhl{181}{234}{249}{ph}\myhl{181}{234}{249}{ (}\myhl{181}{234}{249}{shi}\myhl{181}{234}{249}{ru}\myhl{181}{234}{249}{hu}\myhl{181}{234}{249}{ ,}\myhl{181}{234}{249}{ Sh}\myhl{181}{234}{249}{ir}\myhl{181}{234}{249}{uf}\myhl{181}{234}{249}{u}\myhl{181}{234}{249}{ )}\myhl{181}{234}{249}{ is}\myhl{181}{234}{249}{ a}\myhl{181}{234}{249}{ Japanese}\myhl{181}{234}{249}{ sh}\myhl{181}{234}{249}{o}\myhl{181}{234}{249}{jo}\myhl{181}{234}{249}{ m}\myhl{181}{234}{249}{anga}\myhl{181}{234}{249}{ magazine}\myhl{181}{234}{249}{ published}\myhl{181}{234}{249}{ by}\myhl{181}{234}{249}{ ASCII}\myhl{181}{234}{249}{ Media}\myhl{181}{234}{249}{ Works}\myhl{181}{234}{249}{ (}\myhl{181}{234}{249}{former}\myhl{181}{234}{249}{ly}\myhl{181}{234}{249}{ Media}\myhl{181}{234}{249}{Work}\myhl{181}{234}{249}{s}\myhl{181}{234}{249}{)}\myhl{181}{234}{249}{ and}\myhl{181}{234}{249}{ is}\myhl{181}{234}{249}{ sold}\myhl{181}{234}{249}{ month}\myhl{181}{234}{249}{ly}\myhl{181}{234}{249}{.} \dots\myhl{183}{235}{249}{ The}\myhl{183}{235}{249}{ magazine}\myhl{183}{235}{249}{ was}\myhl{183}{235}{249}{ originally}\myhl{183}{235}{249}{ published}\myhl{183}{235}{249}{ on}\myhl{183}{235}{249}{ December}\myhl{183}{235}{249}{ }\myhl{183}{235}{249}{9}\myhl{183}{235}{249}{,}\myhl{183}{235}{249}{ }\myhl{183}{235}{249}{2}\myhl{183}{235}{249}{0}\myhl{183}{235}{249}{0}\myhl{183}{235}{249}{6}\myhl{183}{235}{249}{ as}\myhl{183}{235}{249}{ a}\myhl{183}{235}{249}{ special}\myhl{183}{235}{249}{ edition}\myhl{183}{235}{249}{ version}\myhl{183}{235}{249}{ of}\myhl{183}{235}{249}{ Media}\myhl{183}{235}{249}{Work}\myhl{183}{235}{249}{s}\myhl{183}{235}{249}{'}\myhl{183}{235}{249}{ now}\myhl{183}{235}{249}{-}\myhl{183}{235}{249}{def}\myhl{183}{235}{249}{unct}\myhl{183}{235}{249}{ "}\myhl{183}{235}{249}{Den}\myhl{183}{235}{249}{ge}\myhl{183}{235}{249}{ki}\myhl{183}{235}{249}{ Com}\myhl{183}{235}{249}{ic}\myhl{183}{235}{249}{ G}\myhl{183}{235}{249}{ao}\myhl{183}{235}{249}{!}\myhl{183}{235}{249}{ "}\myhl{192}{237}{250}{ under}\myhl{192}{237}{250}{ the}\myhl{192}{237}{250}{ title}\myhl{192}{237}{250}{ "}\myhl{192}{237}{250}{Com}\myhl{192}{237}{250}{ic}\myhl{192}{237}{250}{ Syl}\myhl{192}{237}{250}{ph}\myhl{192}{237}{250}{"}\myhl{192}{237}{250}{ (}\myhl{192}{237}{250}{ko}\myhl{192}{237}{250}{mi}\myhl{192}{237}{250}{tsu}\myhl{192}{237}{250}{ku}\myhl{192}{237}{250}{shi}\myhl{192}{237}{250}{ru}\myhl{192}{237}{250}{hu}\myhl{192}{237}{250}{ ,}\myhl{192}{237}{250}{ Kom}\myhl{192}{237}{250}{ik}\myhl{192}{237}{250}{ku}\myhl{192}{237}{250}{ Sh}\myhl{192}{237}{250}{ir}\myhl{192}{237}{250}{uf}\myhl{192}{237}{250}{u}\myhl{192}{237}{250}{ ,}\myhl{192}{237}{250}{ normally}\myhl{192}{237}{250}{ written}\myhl{192}{237}{250}{ as}\myhl{192}{237}{250}{ "}\myhl{192}{237}{250}{com}\myhl{192}{237}{250}{ic}\myhl{192}{237}{250}{ SY}\myhl{192}{237}{250}{LP}\myhl{192}{237}{250}{H}\myhl{192}{237}{250}{")}\myhl{192}{237}{250}{ as}\myhl{192}{237}{250}{ a}\myhl{192}{237}{250}{ quarter}\myhl{192}{237}{250}{ly}\myhl{192}{237}{250}{ publication}\myhl{192}{237}{250}{.} \dots\myhl{169}{231}{248}{ On}\myhl{169}{231}{248}{ March}\myhl{169}{231}{248}{ }\myhl{169}{231}{248}{2}\myhl{169}{231}{248}{1}\myhl{169}{231}{248}{,}\myhl{169}{231}{248}{ }\myhl{169}{231}{248}{2}\myhl{169}{231}{248}{0}\myhl{169}{231}{248}{0}\myhl{169}{231}{248}{8}\myhl{169}{231}{248}{,}\myhl{169}{231}{248}{ with}\myhl{169}{231}{248}{ the}\myhl{169}{231}{248}{ release}\myhl{169}{231}{248}{ of}\myhl{169}{231}{248}{ the}\myhl{169}{231}{248}{ sixth}\myhl{169}{231}{248}{ volume}\myhl{169}{231}{248}{,}\myhl{169}{231}{248}{ the}\myhl{169}{231}{248}{ magazine}\myhl{169}{231}{248}{ was}\myhl{169}{231}{248}{ transferred}\myhl{169}{231}{248}{ over}\myhl{169}{231}{248}{ as}\myhl{169}{231}{248}{ a}\myhl{169}{231}{248}{ special}\myhl{169}{231}{248}{ edition}\myhl{169}{231}{248}{ version}\myhl{169}{231}{248}{ of}\myhl{169}{231}{248}{ ASCII}\myhl{169}{231}{248}{ Media}\myhl{169}{231}{248}{ Works}\myhl{169}{231}{248}{'}\myhl{169}{231}{248}{ sh}\myhl{169}{231}{248}{o}\myhl{169}{231}{248}{nen}\myhl{169}{231}{248}{ m}\myhl{169}{231}{248}{anga}\myhl{169}{231}{248}{ magazine}\myhl{169}{231}{248}{ "}\myhl{169}{231}{248}{Den}\myhl{169}{231}{248}{ge}\myhl{169}{231}{248}{ki}\myhl{169}{231}{248}{ Da}\myhl{169}{231}{248}{io}\myhl{169}{231}{248}{h}\myhl{169}{231}{248}{".} \dots\myhl{109}{214}{243}{ On}\myhl{109}{214}{243}{ May}\myhl{109}{214}{243}{ }\myhl{109}{214}{243}{2}\myhl{109}{214}{243}{2}\myhl{109}{214}{243}{,}\myhl{109}{214}{243}{ }\myhl{109}{214}{243}{2}\myhl{109}{214}{243}{0}\myhl{109}{214}{243}{0}\myhl{109}{214}{243}{8}\myhl{109}{214}{243}{,}\myhl{109}{214}{243}{ the}\myhl{109}{214}{243}{ magazine}\myhl{109}{214}{243}{ became}\myhl{109}{214}{243}{ independent}\myhl{109}{214}{243}{ of}\myhl{109}{214}{243}{ "}\myhl{109}{214}{243}{Den}\myhl{109}{214}{243}{ge}\myhl{109}{214}{243}{ki}\myhl{109}{214}{243}{ Da}\myhl{109}{214}{243}{io}\myhl{109}{214}{243}{h}\myhl{109}{214}{243}{"}\myhl{109}{214}{243}{ and}\myhl{109}{214}{243}{ was}\myhl{109}{214}{243}{ published}\myhl{109}{214}{243}{ as}\myhl{109}{214}{243}{ volume}\myhl{109}{214}{243}{ one}\myhl{109}{214}{243}{ of}\myhl{109}{214}{243}{ "}\myhl{109}{214}{243}{S}\myhl{109}{214}{243}{yl}\myhl{109}{214}{243}{ph}\myhl{109}{214}{243}{"}\myhl{109}{214}{243}{ as}\myhl{109}{214}{243}{ the}\myhl{109}{214}{243}{ July}\myhl{109}{214}{243}{ }\myhl{109}{214}{243}{2}\myhl{109}{214}{243}{0}\myhl{109}{214}{243}{0}\myhl{109}{214}{243}{8}\myhl{109}{214}{243}{ issue}\myhl{109}{214}{243}{ as}\myhl{109}{214}{243}{ a}\myhl{109}{214}{243}{ b}\myhl{109}{214}{243}{imon}\myhl{109}{214}{243}{th}\myhl{109}{214}{243}{ly}\myhl{109}{214}{243}{ publication}\myhl{109}{214}{243}{.} \dots\myhl{36}{193}{237}{ "}\myhl{36}{193}{237}{S}\myhl{36}{193}{237}{yl}\myhl{36}{193}{237}{ph}\myhl{36}{193}{237}{"}\myhl{36}{193}{237}{ is}\myhl{36}{193}{237}{ one}\myhl{36}{193}{237}{ of}\myhl{36}{193}{237}{ the}\myhl{36}{193}{237}{ few}\myhl{36}{193}{237}{ mag}\myhl{36}{193}{237}{az}\myhl{36}{193}{237}{ines}\myhl{36}{193}{237}{ originally}\myhl{36}{193}{237}{ published}\myhl{36}{193}{237}{ by}\myhl{36}{193}{237}{ Media}\myhl{36}{193}{237}{Work}\myhl{36}{193}{237}{s}\myhl{36}{193}{237}{ not}\myhl{36}{193}{237}{ under}\myhl{36}{193}{237}{ the}\myhl{36}{193}{237}{ "}\myhl{36}{193}{237}{Den}\myhl{36}{193}{237}{ge}\myhl{36}{193}{237}{ki}\myhl{36}{193}{237}{"}\myhl{36}{193}{237}{ naming}\myhl{36}{193}{237}{ line}\myhl{36}{193}{237}{,}\myhl{36}{193}{237}{ such}\myhl{36}{193}{237}{ as}\myhl{36}{193}{237}{ with}\myhl{36}{193}{237}{ "}\myhl{36}{193}{237}{Den}\myhl{36}{193}{237}{ge}\myhl{36}{193}{237}{ki}\myhl{36}{193}{237}{ Da}\myhl{36}{193}{237}{io}\myhl{36}{193}{237}{h}\myhl{36}{193}{237}{",}\myhl{36}{193}{237}{ and}\myhl{36}{193}{237}{ "}\myhl{36}{193}{237}{Den}\myhl{36}{193}{237}{ge}\myhl{36}{193}{237}{ki}\myhl{36}{193}{237}{ G}\myhl{36}{193}{237}{'}\myhl{36}{193}{237}{s}\myhl{36}{193}{237}{ Magazine}\myhl{36}{193}{237}{",}\myhl{36}{193}{237}{ the}\myhl{36}{193}{237}{ first}\myhl{36}{193}{237}{ of}\myhl{36}{193}{237}{ which}\myhl{36}{193}{237}{ being}\myhl{36}{193}{237}{ "}\myhl{36}{193}{237}{Active}\myhl{36}{193}{237}{ Japan}\myhl{36}{193}{237}{"}\myhl{36}{193}{237}{ in}\myhl{36}{193}{237}{ }\myhl{36}{193}{237}{1}\myhl{36}{193}{237}{9}\myhl{36}{193}{237}{9}\myhl{36}{193}{237}{5}\myhl{36}{193}{237}{ which}\myhl{36}{193}{237}{ has}\myhl{36}{193}{237}{ been}\myhl{36}{193}{237}{ dis}\myhl{36}{193}{237}{cont}\myhl{36}{193}{237}{in}\myhl{36}{193}{237}{ued}\myhl{36}{193}{237}{ since}\myhl{36}{193}{237}{ }\myhl{36}{193}{237}{1}\myhl{36}{193}{237}{9}\myhl{36}{193}{237}{9}\myhl{36}{193}{237}{8}\myhl{36}{193}{237}{.} \dots\myhl{202}{240}{250}{ The}\myhl{202}{240}{250}{ magazine}\myhl{202}{240}{250}{ include}\myhl{202}{240}{250}{ articles}\myhl{202}{240}{250}{ about}\myhl{202}{240}{250}{ the}\myhl{202}{240}{250}{ following}\myhl{202}{240}{250}{:}\myhl{202}{240}{250}{ history}\myhl{202}{240}{250}{ of}\myhl{202}{240}{250}{ slot}\myhl{202}{240}{250}{ machine}\myhl{202}{240}{250}{,}\myhl{202}{240}{250}{ the}\myhl{202}{240}{250}{ principle}\myhl{202}{240}{250}{ of}\myhl{202}{240}{250}{ slot}\myhl{202}{240}{250}{ machine}\myhl{202}{240}{250}{ work}\myhl{202}{240}{250}{,}\myhl{202}{240}{250}{ new}\myhl{202}{240}{250}{ strateg}\myhl{202}{240}{250}{ies}\myhl{202}{240}{250}{ for}\myhl{202}{240}{250}{ playing}\myhl{202}{240}{250}{ video}\myhl{202}{240}{250}{ pok}\myhl{202}{240}{250}{er}\myhl{202}{240}{250}{,}\myhl{202}{240}{250}{ cas}\myhl{202}{240}{250}{ino}\myhl{202}{240}{250}{ reviews}\myhl{202}{240}{250}{ and}\myhl{202}{240}{250}{ latest}\myhl{202}{240}{250}{ cas}\myhl{202}{240}{250}{ino}\myhl{202}{240}{250}{ news}\myhl{202}{240}{250}{ etc}\myhl{202}{240}{250}{.} \dots\myhl{151}{225}{246}{The}\myhl{151}{225}{246}{ Le}\myhl{151}{225}{246}{eds}\myhl{151}{225}{246}{ Guide}\myhl{151}{225}{246}{ was}\myhl{151}{225}{246}{ a}\myhl{151}{225}{246}{ month}\myhl{151}{225}{246}{ly}\myhl{151}{225}{246}{ "}\myhl{151}{225}{246}{What}\myhl{151}{225}{246}{'}\myhl{151}{225}{246}{s}\myhl{151}{225}{246}{ on}\myhl{151}{225}{246}{"}\myhl{151}{225}{246}{ magazine}\myhl{151}{225}{246}{ published}\myhl{151}{225}{246}{ in}\myhl{151}{225}{246}{ Le}\myhl{151}{225}{246}{eds}\myhl{151}{225}{246}{,}\myhl{151}{225}{246}{ West}\myhl{151}{225}{246}{ York}\myhl{151}{225}{246}{shire}\myhl{151}{225}{246}{,}\myhl{151}{225}{246}{ England}\myhl{151}{225}{246}{ from}\myhl{151}{225}{246}{ }\myhl{151}{225}{246}{1}\myhl{151}{225}{246}{9}\myhl{151}{225}{246}{9}\myhl{151}{225}{246}{7}\myhl{151}{225}{246}{ until}\myhl{151}{225}{246}{ }\myhl{151}{225}{246}{2}\myhl{151}{225}{246}{0}\myhl{151}{225}{246}{1}\myhl{151}{225}{246}{2}\myhl{151}{225}{246}{.} \dots\myhl{163}{229}{247}{ Origin}\myhl{163}{229}{247}{ally}\myhl{163}{229}{247}{ a}\myhl{163}{229}{247}{ month}\myhl{163}{229}{247}{ly}\myhl{163}{229}{247}{ A}\myhl{163}{229}{247}{5}\myhl{163}{229}{247}{ magazine}\myhl{163}{229}{247}{ printed}\myhl{163}{229}{247}{ in}\myhl{163}{229}{247}{ black}\myhl{163}{229}{247}{ and}\myhl{163}{229}{247}{ white}\myhl{163}{229}{247}{,}\myhl{163}{229}{247}{ "}\myhl{163}{229}{247}{The}\myhl{163}{229}{247}{ Le}\myhl{163}{229}{247}{eds}\myhl{163}{229}{247}{ Guide}\myhl{163}{229}{247}{"}\myhl{163}{229}{247}{ changed}\myhl{163}{229}{247}{ to}\myhl{163}{229}{247}{ an}\myhl{163}{229}{247}{ A}\myhl{163}{229}{247}{4}\myhl{163}{229}{247}{ fort}\myhl{163}{229}{247}{night}\myhl{163}{229}{247}{ly}\myhl{163}{229}{247}{ in}\myhl{163}{229}{247}{ }\myhl{163}{229}{247}{2}\myhl{163}{229}{247}{0}\myhl{163}{229}{247}{0}\myhl{163}{229}{247}{3}\myhl{163}{229}{247}{ and}\myhl{163}{229}{247}{ subsequently}\myhl{163}{229}{247}{ went}\myhl{163}{229}{247}{ back}\myhl{163}{229}{247}{ to}\myhl{163}{229}{247}{ being}\myhl{163}{229}{247}{ published}\myhl{163}{229}{247}{ month}\myhl{163}{229}{247}{ly}\myhl{163}{229}{247}{.} \dots\myhl{156}{227}{247}{Se}\myhl{156}{227}{247}{ka}\myhl{156}{227}{247}{i}\myhl{156}{227}{247}{ (}\myhl{156}{227}{247}{J}\myhl{156}{227}{247}{apan}\myhl{156}{227}{247}{ese}\myhl{156}{227}{247}{:}\myhl{156}{227}{247}{ }\myhl{156}{227}{247}{Shi }\myhl{156}{227}{247}{Jie }\myhl{156}{227}{247}{ "}\myhl{156}{227}{247}{World}\myhl{156}{227}{247}{")}\myhl{156}{227}{247}{ is}\myhl{156}{227}{247}{ a}\myhl{156}{227}{247}{ Japanese}\myhl{156}{227}{247}{ month}\myhl{156}{227}{247}{ly}\myhl{156}{227}{247}{ political}\myhl{156}{227}{247}{ magazine}\myhl{156}{227}{247}{ published}\myhl{156}{227}{247}{ by}\myhl{156}{227}{247}{ I}\myhl{156}{227}{247}{wan}\myhl{156}{227}{247}{ami}\myhl{156}{227}{247}{ Sh}\myhl{156}{227}{247}{oten}\myhl{156}{227}{247}{,}\myhl{156}{227}{247}{ which}\myhl{156}{227}{247}{ was}\myhl{156}{227}{247}{ founded}\myhl{156}{227}{247}{ in}\myhl{156}{227}{247}{ December}\myhl{156}{227}{247}{ }\myhl{156}{227}{247}{1}\myhl{156}{227}{247}{9}\myhl{156}{227}{247}{4}\myhl{156}{227}{247}{5}\myhl{156}{227}{247}{.} \dots\myhl{200}{239}{250}{ The}\myhl{200}{239}{250}{ magazine}\myhl{200}{239}{250}{'}\myhl{200}{239}{250}{s}\myhl{200}{239}{250}{ found}\myhl{200}{239}{250}{ing}\myhl{200}{239}{250}{ principles}\myhl{200}{239}{250}{ were}\myhl{200}{239}{250}{ "}\myhl{200}{239}{250}{pe}\myhl{200}{239}{250}{ace}\myhl{200}{239}{250}{ and}\myhl{200}{239}{250}{ social}\myhl{200}{239}{250}{ justice}\myhl{200}{239}{250}{,}\myhl{200}{239}{250}{ freedom}\myhl{200}{239}{250}{ and}\myhl{200}{239}{250}{ equality}\myhl{200}{239}{250}{,}\myhl{200}{239}{250}{ and}\myhl{200}{239}{250}{ harm}\myhl{200}{239}{250}{ony}\myhl{200}{239}{250}{ and}\myhl{200}{239}{250}{ solid}\myhl{200}{239}{250}{ar}\myhl{200}{239}{250}{ity}\myhl{200}{239}{250}{ with}\myhl{200}{239}{250}{ the}\myhl{200}{239}{250}{ pe}\myhl{200}{239}{250}{op}\myhl{200}{239}{250}{les}\myhl{200}{239}{250}{ of}\myhl{200}{239}{250}{ East}\myhl{200}{239}{250}{ Asia}\myhl{200}{239}{250}{."} \dots\myhl{152}{226}{246}{Gre}\myhl{152}{226}{246}{at}\myhl{152}{226}{246}{ L}\myhl{152}{226}{246}{akes}\myhl{152}{226}{246}{ Life}\myhl{152}{226}{246}{ Magazine}\myhl{152}{226}{246}{ was}\myhl{152}{226}{246}{ a}\myhl{152}{226}{246}{ regional}\myhl{152}{226}{246}{ magazine}\myhl{152}{226}{246}{ that}\myhl{152}{226}{246}{ was}\myhl{152}{226}{246}{ published}\myhl{152}{226}{246}{ in}\myhl{152}{226}{246}{ West}\myhl{152}{226}{246}{field}\myhl{152}{226}{246}{,}\myhl{152}{226}{246}{ New}\myhl{152}{226}{246}{ York}\myhl{152}{226}{246}{.} \dots\myhl{155}{227}{247}{ It}\myhl{155}{227}{247}{ was}\myhl{155}{227}{247}{ founded}\myhl{155}{227}{247}{ by}\myhl{155}{227}{247}{ editor}\myhl{155}{227}{247}{-}\myhl{155}{227}{247}{in}\myhl{155}{227}{247}{-}\myhl{155}{227}{247}{ch}\myhl{155}{227}{247}{ief}\myhl{155}{227}{247}{ and}\myhl{155}{227}{247}{ publish}\myhl{155}{227}{247}{er}\myhl{155}{227}{247}{ Rena}\myhl{155}{227}{247}{ T}\myhl{155}{227}{247}{ran}\myhl{155}{227}{247}{ and}\myhl{155}{227}{247}{ began}\myhl{155}{227}{247}{ publication}\myhl{155}{227}{247}{ in}\myhl{155}{227}{247}{ February}\myhl{155}{227}{247}{ }\myhl{155}{227}{247}{2}\myhl{155}{227}{247}{0}\myhl{155}{227}{247}{0}\myhl{155}{227}{247}{8}\myhl{155}{227}{247}{ under}\myhl{155}{227}{247}{ the}\myhl{155}{227}{247}{ title}\myhl{155}{227}{247}{ Er}\myhl{155}{227}{247}{ie}\myhl{155}{227}{247}{ Life}\myhl{155}{227}{247}{ Magazine}\myhl{155}{227}{247}{,}\myhl{155}{227}{247}{ but}\myhl{155}{227}{247}{ went}\myhl{155}{227}{247}{ out}\myhl{155}{227}{247}{ of}\myhl{155}{227}{247}{ production}\myhl{155}{227}{247}{ in}\myhl{155}{227}{247}{ }\myhl{155}{227}{247}{2}\myhl{155}{227}{247}{0}\myhl{155}{227}{247}{1}\myhl{155}{227}{247}{1}\myhl{155}{227}{247}{.} \dots
\end{tcolorbox}
\begin{tcolorbox}[colback=white, , title=Average attention]
\textbf{Context}:
\myhl{169}{231}{248}{M}\myhl{169}{231}{248}{anga}\myhl{169}{231}{248}{ Life}\myhl{169}{231}{248}{ (}\myhl{169}{231}{248}{ma}\myhl{169}{231}{248}{n}\myhl{169}{231}{248}{ga}\myhl{169}{231}{248}{ra}\myhl{169}{231}{248}{i}\myhl{169}{231}{248}{hu}\myhl{169}{231}{248}{ ,}\myhl{169}{231}{248}{ M}\myhl{169}{231}{248}{anga}\myhl{169}{231}{248}{ Ra}\myhl{169}{231}{248}{if}\myhl{169}{231}{248}{u}\myhl{169}{231}{248}{ )}\myhl{169}{231}{248}{ is}\myhl{169}{231}{248}{ a}\myhl{169}{231}{248}{ m}\myhl{169}{231}{248}{anga}\myhl{169}{231}{248}{ magazine}\myhl{169}{231}{248}{ published}\myhl{169}{231}{248}{ month}\myhl{169}{231}{248}{ly}\myhl{169}{231}{248}{ by}\myhl{169}{231}{248}{ T}\myhl{169}{231}{248}{akes}\myhl{169}{231}{248}{h}\myhl{169}{231}{248}{ob}\myhl{169}{231}{248}{o}\myhl{169}{231}{248}{ in}\myhl{169}{231}{248}{ Japan}\myhl{169}{231}{248}{ since}\myhl{169}{231}{248}{ the}\myhl{169}{231}{248}{ November}\myhl{169}{231}{248}{ }\myhl{169}{231}{248}{1}\myhl{169}{231}{248}{9}\myhl{169}{231}{248}{8}\myhl{169}{231}{248}{4}\myhl{169}{231}{248}{ issue}\myhl{169}{231}{248}{ (}\myhl{169}{231}{248}{pub}\myhl{169}{231}{248}{lished}\myhl{169}{231}{248}{ in}\myhl{169}{231}{248}{ October}\myhl{169}{231}{248}{ }\myhl{169}{231}{248}{1}\myhl{169}{231}{248}{9}\myhl{169}{231}{248}{8}\myhl{169}{231}{248}{4}\myhl{169}{231}{248}{).} \dots\myhl{179}{233}{248}{ Most}\myhl{179}{233}{248}{ of}\myhl{179}{233}{248}{ the}\myhl{179}{233}{248}{ series}\myhl{179}{233}{248}{ appearing}\myhl{179}{233}{248}{ in}\myhl{179}{233}{248}{ the}\myhl{179}{233}{248}{ magazine}\myhl{179}{233}{248}{ use}\myhl{179}{233}{248}{ the}\myhl{179}{233}{248}{ y}\myhl{179}{233}{248}{onk}\myhl{179}{233}{248}{oma}\myhl{179}{233}{248}{ format}\myhl{179}{233}{248}{.} \dots\myhl{0}{183}{234}{ The}\myhl{0}{183}{234}{ magazine}\myhl{0}{183}{234}{ was}\myhl{0}{183}{234}{ first}\myhl{0}{183}{234}{ published}\myhl{0}{183}{234}{ digit}\myhl{0}{183}{234}{ally}\myhl{0}{183}{234}{ on}\myhl{0}{183}{234}{ August}\myhl{0}{183}{234}{ }\myhl{0}{183}{234}{9}\myhl{0}{183}{234}{,}\myhl{0}{183}{234}{ }\myhl{0}{183}{234}{2}\myhl{0}{183}{234}{0}\myhl{0}{183}{234}{1}\myhl{0}{183}{234}{2}\myhl{0}{183}{234}{ with}\myhl{0}{183}{234}{ volume}\myhl{0}{183}{234}{ }\myhl{0}{183}{234}{0}\myhl{0}{183}{234}{,}\myhl{0}{183}{234}{ and}\myhl{0}{183}{234}{ started}\myhl{0}{183}{234}{ month}\myhl{0}{183}{234}{ly}\myhl{0}{183}{234}{ publication}\myhl{0}{183}{234}{ with}\myhl{0}{183}{234}{ the}\myhl{0}{183}{234}{ following}\myhl{0}{183}{234}{ issue}\myhl{0}{183}{234}{ released}\myhl{0}{183}{234}{ on}\myhl{0}{183}{234}{ October}\myhl{0}{183}{234}{ }\myhl{0}{183}{234}{1}\myhl{0}{183}{234}{5}\myhl{0}{183}{234}{,}\myhl{0}{183}{234}{ }\myhl{0}{183}{234}{2}\myhl{0}{183}{234}{0}\myhl{0}{183}{234}{1}\myhl{0}{183}{234}{2}\myhl{0}{183}{234}{.} \dots\myhl{167}{230}{248}{ New}\myhl{167}{230}{248}{port}\myhl{167}{230}{248}{ Life}\myhl{167}{230}{248}{ Magazine}\myhl{167}{230}{248}{ is}\myhl{167}{230}{248}{ located}\myhl{167}{230}{248}{ at}\myhl{167}{230}{248}{ }\myhl{167}{230}{248}{1}\myhl{167}{230}{248}{0}\myhl{167}{230}{248}{1}\myhl{167}{230}{248}{ Mal}\myhl{167}{230}{248}{bone}\myhl{167}{230}{248}{ Road}\myhl{167}{230}{248}{ in}\myhl{167}{230}{248}{ the}\myhl{167}{230}{248}{ New}\myhl{167}{230}{248}{port}\myhl{167}{230}{248}{ Daily}\myhl{167}{230}{248}{ News}\myhl{167}{230}{248}{ building}\myhl{167}{230}{248}{.} \dots\myhl{172}{231}{248}{ On}\myhl{172}{231}{248}{ May}\myhl{172}{231}{248}{ }\myhl{172}{231}{248}{2}\myhl{172}{231}{248}{2}\myhl{172}{231}{248}{,}\myhl{172}{231}{248}{ }\myhl{172}{231}{248}{2}\myhl{172}{231}{248}{0}\myhl{172}{231}{248}{0}\myhl{172}{231}{248}{8}\myhl{172}{231}{248}{,}\myhl{172}{231}{248}{ the}\myhl{172}{231}{248}{ magazine}\myhl{172}{231}{248}{ became}\myhl{172}{231}{248}{ independent}\myhl{172}{231}{248}{ of}\myhl{172}{231}{248}{ "}\myhl{172}{231}{248}{Den}\myhl{172}{231}{248}{ge}\myhl{172}{231}{248}{ki}\myhl{172}{231}{248}{ Da}\myhl{172}{231}{248}{io}\myhl{172}{231}{248}{h}\myhl{172}{231}{248}{"}\myhl{172}{231}{248}{ and}\myhl{172}{231}{248}{ was}\myhl{172}{231}{248}{ published}\myhl{172}{231}{248}{ as}\myhl{172}{231}{248}{ volume}\myhl{172}{231}{248}{ one}\myhl{172}{231}{248}{ of}\myhl{172}{231}{248}{ "}\myhl{172}{231}{248}{S}\myhl{172}{231}{248}{yl}\myhl{172}{231}{248}{ph}\myhl{172}{231}{248}{"}\myhl{172}{231}{248}{ as}\myhl{172}{231}{248}{ the}\myhl{172}{231}{248}{ July}\myhl{172}{231}{248}{ }\myhl{172}{231}{248}{2}\myhl{172}{231}{248}{0}\myhl{172}{231}{248}{0}\myhl{172}{231}{248}{8}\myhl{172}{231}{248}{ issue}\myhl{172}{231}{248}{ as}\myhl{172}{231}{248}{ a}\myhl{172}{231}{248}{ b}\myhl{172}{231}{248}{imon}\myhl{172}{231}{248}{th}\myhl{172}{231}{248}{ly}\myhl{172}{231}{248}{ publication}\myhl{172}{231}{248}{.} \dots\myhl{108}{213}{243}{ "}\myhl{108}{213}{243}{S}\myhl{108}{213}{243}{yl}\myhl{108}{213}{243}{ph}\myhl{108}{213}{243}{"}\myhl{108}{213}{243}{ is}\myhl{108}{213}{243}{ one}\myhl{108}{213}{243}{ of}\myhl{108}{213}{243}{ the}\myhl{108}{213}{243}{ few}\myhl{108}{213}{243}{ mag}\myhl{108}{213}{243}{az}\myhl{108}{213}{243}{ines}\myhl{108}{213}{243}{ originally}\myhl{108}{213}{243}{ published}\myhl{108}{213}{243}{ by}\myhl{108}{213}{243}{ Media}\myhl{108}{213}{243}{Work}\myhl{108}{213}{243}{s}\myhl{108}{213}{243}{ not}\myhl{108}{213}{243}{ under}\myhl{108}{213}{243}{ the}\myhl{108}{213}{243}{ "}\myhl{108}{213}{243}{Den}\myhl{108}{213}{243}{ge}\myhl{108}{213}{243}{ki}\myhl{108}{213}{243}{"}\myhl{108}{213}{243}{ naming}\myhl{108}{213}{243}{ line}\myhl{108}{213}{243}{,}\myhl{108}{213}{243}{ such}\myhl{108}{213}{243}{ as}\myhl{108}{213}{243}{ with}\myhl{108}{213}{243}{ "}\myhl{108}{213}{243}{Den}\myhl{108}{213}{243}{ge}\myhl{108}{213}{243}{ki}\myhl{108}{213}{243}{ Da}\myhl{108}{213}{243}{io}\myhl{108}{213}{243}{h}\myhl{108}{213}{243}{",}\myhl{108}{213}{243}{ and}\myhl{108}{213}{243}{ "}\myhl{108}{213}{243}{Den}\myhl{108}{213}{243}{ge}\myhl{108}{213}{243}{ki}\myhl{108}{213}{243}{ G}\myhl{108}{213}{243}{'}\myhl{108}{213}{243}{s}\myhl{108}{213}{243}{ Magazine}\myhl{108}{213}{243}{",}\myhl{108}{213}{243}{ the}\myhl{108}{213}{243}{ first}\myhl{108}{213}{243}{ of}\myhl{108}{213}{243}{ which}\myhl{108}{213}{243}{ being}\myhl{108}{213}{243}{ "}\myhl{108}{213}{243}{Active}\myhl{108}{213}{243}{ Japan}\myhl{108}{213}{243}{"}\myhl{108}{213}{243}{ in}\myhl{108}{213}{243}{ }\myhl{108}{213}{243}{1}\myhl{108}{213}{243}{9}\myhl{108}{213}{243}{9}\myhl{108}{213}{243}{5}\myhl{108}{213}{243}{ which}\myhl{108}{213}{243}{ has}\myhl{108}{213}{243}{ been}\myhl{108}{213}{243}{ dis}\myhl{108}{213}{243}{cont}\myhl{108}{213}{243}{in}\myhl{108}{213}{243}{ued}\myhl{108}{213}{243}{ since}\myhl{108}{213}{243}{ }\myhl{108}{213}{243}{1}\myhl{108}{213}{243}{9}\myhl{108}{213}{243}{9}\myhl{108}{213}{243}{8}\myhl{108}{213}{243}{.} \dots
\end{tcolorbox}
\begin{tcolorbox}[colback=white, , title=ESM (256 samples)]
\textbf{Context}:
 \dots\myhl{198}{239}{250}{ From}\myhl{198}{239}{250}{ April}\myhl{198}{239}{250}{ }\myhl{198}{239}{250}{2}\myhl{198}{239}{250}{0}\myhl{198}{239}{250}{1}\myhl{198}{239}{250}{3}\myhl{198}{239}{250}{ to}\myhl{198}{239}{250}{ April}\myhl{198}{239}{250}{ }\myhl{198}{239}{250}{2}\myhl{198}{239}{250}{0}\myhl{198}{239}{250}{1}\myhl{198}{239}{250}{4}\myhl{198}{239}{250}{,}\myhl{198}{239}{250}{ the}\myhl{198}{239}{250}{ magazine}\myhl{198}{239}{250}{ was}\myhl{198}{239}{250}{ released}\myhl{198}{239}{250}{ bi}\myhl{198}{239}{250}{week}\myhl{198}{239}{250}{ly}\myhl{198}{239}{250}{.} \dots\myhl{0}{183}{234}{Life}\myhl{0}{183}{234}{ was}\myhl{0}{183}{234}{ an}\myhl{0}{183}{234}{ American}\myhl{0}{183}{234}{ magazine}\myhl{0}{183}{234}{ that}\myhl{0}{183}{234}{ ran}\myhl{0}{183}{234}{ week}\myhl{0}{183}{234}{ly}\myhl{0}{183}{234}{ from}\myhl{0}{183}{234}{ }\myhl{0}{183}{234}{1}\myhl{0}{183}{234}{8}\myhl{0}{183}{234}{8}\myhl{0}{183}{234}{3}\myhl{0}{183}{234}{ to}\myhl{0}{183}{234}{ }\myhl{0}{183}{234}{1}\myhl{0}{183}{234}{9}\myhl{0}{183}{234}{3}\myhl{0}{183}{234}{6}\myhl{0}{183}{234}{ as}\myhl{0}{183}{234}{ a}\myhl{0}{183}{234}{ humor}\myhl{0}{183}{234}{ magazine}\myhl{0}{183}{234}{ with}\myhl{0}{183}{234}{ limited}\myhl{0}{183}{234}{ circul}\myhl{0}{183}{234}{ation}\myhl{0}{183}{234}{.}\myhl{0}{183}{234}{ "} \dots\myhl{197}{239}{250}{S}\myhl{197}{239}{250}{yl}\myhl{197}{239}{250}{ph}\myhl{197}{239}{250}{ (}\myhl{197}{239}{250}{shi}\myhl{197}{239}{250}{ru}\myhl{197}{239}{250}{hu}\myhl{197}{239}{250}{ ,}\myhl{197}{239}{250}{ Sh}\myhl{197}{239}{250}{ir}\myhl{197}{239}{250}{uf}\myhl{197}{239}{250}{u}\myhl{197}{239}{250}{ )}\myhl{197}{239}{250}{ is}\myhl{197}{239}{250}{ a}\myhl{197}{239}{250}{ Japanese}\myhl{197}{239}{250}{ sh}\myhl{197}{239}{250}{o}\myhl{197}{239}{250}{jo}\myhl{197}{239}{250}{ m}\myhl{197}{239}{250}{anga}\myhl{197}{239}{250}{ magazine}\myhl{197}{239}{250}{ published}\myhl{197}{239}{250}{ by}\myhl{197}{239}{250}{ ASCII}\myhl{197}{239}{250}{ Media}\myhl{197}{239}{250}{ Works}\myhl{197}{239}{250}{ (}\myhl{197}{239}{250}{former}\myhl{197}{239}{250}{ly}\myhl{197}{239}{250}{ Media}\myhl{197}{239}{250}{Work}\myhl{197}{239}{250}{s}\myhl{197}{239}{250}{)}\myhl{197}{239}{250}{ and}\myhl{197}{239}{250}{ is}\myhl{197}{239}{250}{ sold}\myhl{197}{239}{250}{ month}\myhl{197}{239}{250}{ly}\myhl{197}{239}{250}{.} \dots\myhl{178}{233}{248}{ under}\myhl{178}{233}{248}{ the}\myhl{178}{233}{248}{ title}\myhl{178}{233}{248}{ "}\myhl{178}{233}{248}{Com}\myhl{178}{233}{248}{ic}\myhl{178}{233}{248}{ Syl}\myhl{178}{233}{248}{ph}\myhl{178}{233}{248}{"}\myhl{178}{233}{248}{ (}\myhl{178}{233}{248}{ko}\myhl{178}{233}{248}{mi}\myhl{178}{233}{248}{tsu}\myhl{178}{233}{248}{ku}\myhl{178}{233}{248}{shi}\myhl{178}{233}{248}{ru}\myhl{178}{233}{248}{hu}\myhl{178}{233}{248}{ ,}\myhl{178}{233}{248}{ Kom}\myhl{178}{233}{248}{ik}\myhl{178}{233}{248}{ku}\myhl{178}{233}{248}{ Sh}\myhl{178}{233}{248}{ir}\myhl{178}{233}{248}{uf}\myhl{178}{233}{248}{u}\myhl{178}{233}{248}{ ,}\myhl{178}{233}{248}{ normally}\myhl{178}{233}{248}{ written}\myhl{178}{233}{248}{ as}\myhl{178}{233}{248}{ "}\myhl{178}{233}{248}{com}\myhl{178}{233}{248}{ic}\myhl{178}{233}{248}{ SY}\myhl{178}{233}{248}{LP}\myhl{178}{233}{248}{H}\myhl{178}{233}{248}{")}\myhl{178}{233}{248}{ as}\myhl{178}{233}{248}{ a}\myhl{178}{233}{248}{ quarter}\myhl{178}{233}{248}{ly}\myhl{178}{233}{248}{ publication}\myhl{178}{233}{248}{.} \dots\myhl{160}{228}{247}{ "}\myhl{160}{228}{247}{S}\myhl{160}{228}{247}{yl}\myhl{160}{228}{247}{ph}\myhl{160}{228}{247}{"}\myhl{160}{228}{247}{ is}\myhl{160}{228}{247}{ one}\myhl{160}{228}{247}{ of}\myhl{160}{228}{247}{ the}\myhl{160}{228}{247}{ few}\myhl{160}{228}{247}{ mag}\myhl{160}{228}{247}{az}\myhl{160}{228}{247}{ines}\myhl{160}{228}{247}{ originally}\myhl{160}{228}{247}{ published}\myhl{160}{228}{247}{ by}\myhl{160}{228}{247}{ Media}\myhl{160}{228}{247}{Work}\myhl{160}{228}{247}{s}\myhl{160}{228}{247}{ not}\myhl{160}{228}{247}{ under}\myhl{160}{228}{247}{ the}\myhl{160}{228}{247}{ "}\myhl{160}{228}{247}{Den}\myhl{160}{228}{247}{ge}\myhl{160}{228}{247}{ki}\myhl{160}{228}{247}{"}\myhl{160}{228}{247}{ naming}\myhl{160}{228}{247}{ line}\myhl{160}{228}{247}{,}\myhl{160}{228}{247}{ such}\myhl{160}{228}{247}{ as}\myhl{160}{228}{247}{ with}\myhl{160}{228}{247}{ "}\myhl{160}{228}{247}{Den}\myhl{160}{228}{247}{ge}\myhl{160}{228}{247}{ki}\myhl{160}{228}{247}{ Da}\myhl{160}{228}{247}{io}\myhl{160}{228}{247}{h}\myhl{160}{228}{247}{",}\myhl{160}{228}{247}{ and}\myhl{160}{228}{247}{ "}\myhl{160}{228}{247}{Den}\myhl{160}{228}{247}{ge}\myhl{160}{228}{247}{ki}\myhl{160}{228}{247}{ G}\myhl{160}{228}{247}{'}\myhl{160}{228}{247}{s}\myhl{160}{228}{247}{ Magazine}\myhl{160}{228}{247}{",}\myhl{160}{228}{247}{ the}\myhl{160}{228}{247}{ first}\myhl{160}{228}{247}{ of}\myhl{160}{228}{247}{ which}\myhl{160}{228}{247}{ being}\myhl{160}{228}{247}{ "}\myhl{160}{228}{247}{Active}\myhl{160}{228}{247}{ Japan}\myhl{160}{228}{247}{"}\myhl{160}{228}{247}{ in}\myhl{160}{228}{247}{ }\myhl{160}{228}{247}{1}\myhl{160}{228}{247}{9}\myhl{160}{228}{247}{9}\myhl{160}{228}{247}{5}\myhl{160}{228}{247}{ which}\myhl{160}{228}{247}{ has}\myhl{160}{228}{247}{ been}\myhl{160}{228}{247}{ dis}\myhl{160}{228}{247}{cont}\myhl{160}{228}{247}{in}\myhl{160}{228}{247}{ued}\myhl{160}{228}{247}{ since}\myhl{160}{228}{247}{ }\myhl{160}{228}{247}{1}\myhl{160}{228}{247}{9}\myhl{160}{228}{247}{9}\myhl{160}{228}{247}{8}\myhl{160}{228}{247}{.} \dots\myhl{112}{215}{243}{Str}\myhl{112}{215}{243}{ict}\myhl{112}{215}{243}{ly}\myhl{112}{215}{243}{ Sl}\myhl{112}{215}{243}{ots}\myhl{112}{215}{243}{ is}\myhl{112}{215}{243}{ a}\myhl{112}{215}{243}{ month}\myhl{112}{215}{243}{ly}\myhl{112}{215}{243}{ magazine}\myhl{112}{215}{243}{ aim}\myhl{112}{215}{243}{ed}\myhl{112}{215}{243}{ at}\myhl{112}{215}{243}{ slot}\myhl{112}{215}{243}{ machine}\myhl{112}{215}{243}{ and}\myhl{112}{215}{243}{ video}\myhl{112}{215}{243}{ pok}\myhl{112}{215}{243}{er}\myhl{112}{215}{243}{ players}\myhl{112}{215}{243}{.} \dots\myhl{51}{197}{238}{ The}\myhl{51}{197}{238}{ magazine}\myhl{51}{197}{238}{ was}\myhl{51}{197}{238}{ founded}\myhl{51}{197}{238}{ in}\myhl{51}{197}{238}{ }\myhl{51}{197}{238}{1}\myhl{51}{197}{238}{9}\myhl{51}{197}{238}{9}\myhl{51}{197}{238}{8}\myhl{51}{197}{238}{.} \dots\myhl{88}{208}{241}{ Str}\myhl{88}{208}{241}{ict}\myhl{88}{208}{241}{ly}\myhl{88}{208}{241}{ Sl}\myhl{88}{208}{241}{ots}\myhl{88}{208}{241}{ is}\myhl{88}{208}{241}{ published}\myhl{88}{208}{241}{ by}\myhl{88}{208}{241}{ the}\myhl{88}{208}{241}{ Cas}\myhl{88}{208}{241}{ino}\myhl{88}{208}{241}{ Player}\myhl{88}{208}{241}{ Publishing}\myhl{88}{208}{241}{,}\myhl{88}{208}{241}{ which}\myhl{88}{208}{241}{ also}\myhl{88}{208}{241}{ publish}\myhl{88}{208}{241}{es}\myhl{88}{208}{241}{ "}\myhl{88}{208}{241}{C}\myhl{88}{208}{241}{as}\myhl{88}{208}{241}{ino}\myhl{88}{208}{241}{ Player}\myhl{88}{208}{241}{"}\myhl{88}{208}{241}{ magazine}\myhl{88}{208}{241}{.} \dots
\end{tcolorbox}
\textbf{Query}:
Were both Life magazine and Strictly Slots magazine published monthly in 1998?
\newline\newline\textbf{Response} (from \texttt{Phi-3.5-Mini}):
\newline
\myhl{249}{237}{91}{No, they were not.} Life magazine, which ran from 1883 to 1936 and then again from 1978 to 2000, was not published monthly in 1998. It was a weekly magazine during its second run and then became an intermittent "special" until 1978, after which it was published monthly.On the other hand, Strictly Slots is a monthly magazine that was founded in 1998. It is aimed at slot machine and video poker players and includes articles about the history of slot machines, strategies for playing video poker, casino reviews, and the latest casino news.So, only Strictly Slots was published monthly in 1998.
\end{tcolorbox}
\relsize{2}
\caption{Visualized attributions for a random example from Hotpot QA for \texttt{Phi-3.5-Mini} with sentences as sources.}
\label{fig:hotpot_qa_passages_phi_3_5_mini_sentence}
\end{figure*}

\begin{figure*}[t]
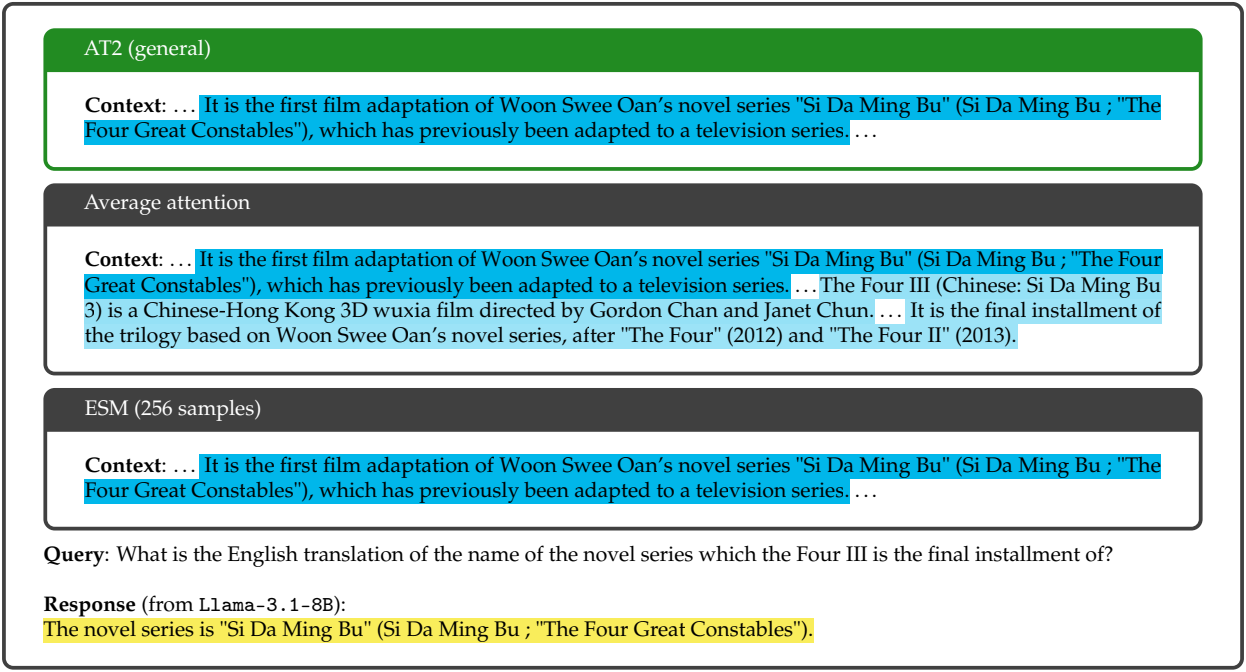

\centering\definecolor{forestgreen}{RGB}{34, 139, 34}
\relsize{-1}
\begin{tcolorbox}[colback=white]
\begin{tcolorbox}[colback=white, colframe=forestgreen, title=\ours (general)]
\textbf{Context}:
 \dots\myhl{0}{183}{234}{ It}\myhl{0}{183}{234}{ is}\myhl{0}{183}{234}{ the}\myhl{0}{183}{234}{ first}\myhl{0}{183}{234}{ film}\myhl{0}{183}{234}{ adaptation}\myhl{0}{183}{234}{ of}\myhl{0}{183}{234}{ W}\myhl{0}{183}{234}{oon}\myhl{0}{183}{234}{ Swe}\myhl{0}{183}{234}{e}\myhl{0}{183}{234}{ O}\myhl{0}{183}{234}{an}\myhl{0}{183}{234}{'s}\myhl{0}{183}{234}{ novel}\myhl{0}{183}{234}{ series}\myhl{0}{183}{234}{ "}\myhl{0}{183}{234}{Si}\myhl{0}{183}{234}{ Da}\myhl{0}{183}{234}{ Ming}\myhl{0}{183}{234}{ Bu}\myhl{0}{183}{234}{"}\myhl{0}{183}{234}{ (}\myhl{0}{183}{234}{Si }\myhl{0}{183}{234}{Da }\myhl{0}{183}{234}{Ming }\myhl{0}{183}{234}{Bu }\myhl{0}{183}{234}{;}\myhl{0}{183}{234}{ "}\myhl{0}{183}{234}{The}\myhl{0}{183}{234}{ Four}\myhl{0}{183}{234}{ Great}\myhl{0}{183}{234}{ Const}\myhl{0}{183}{234}{ables}\myhl{0}{183}{234}{"),}\myhl{0}{183}{234}{ which}\myhl{0}{183}{234}{ has}\myhl{0}{183}{234}{ previously}\myhl{0}{183}{234}{ been}\myhl{0}{183}{234}{ adapted}\myhl{0}{183}{234}{ to}\myhl{0}{183}{234}{ a}\myhl{0}{183}{234}{ television}\myhl{0}{183}{234}{ series}\myhl{0}{183}{234}{.} \dots
\end{tcolorbox}
\begin{tcolorbox}[colback=white, , title=Average attention]
\textbf{Context}:
 \dots\myhl{0}{183}{234}{ It}\myhl{0}{183}{234}{ is}\myhl{0}{183}{234}{ the}\myhl{0}{183}{234}{ first}\myhl{0}{183}{234}{ film}\myhl{0}{183}{234}{ adaptation}\myhl{0}{183}{234}{ of}\myhl{0}{183}{234}{ W}\myhl{0}{183}{234}{oon}\myhl{0}{183}{234}{ Swe}\myhl{0}{183}{234}{e}\myhl{0}{183}{234}{ O}\myhl{0}{183}{234}{an}\myhl{0}{183}{234}{'s}\myhl{0}{183}{234}{ novel}\myhl{0}{183}{234}{ series}\myhl{0}{183}{234}{ "}\myhl{0}{183}{234}{Si}\myhl{0}{183}{234}{ Da}\myhl{0}{183}{234}{ Ming}\myhl{0}{183}{234}{ Bu}\myhl{0}{183}{234}{"}\myhl{0}{183}{234}{ (}\myhl{0}{183}{234}{Si }\myhl{0}{183}{234}{Da }\myhl{0}{183}{234}{Ming }\myhl{0}{183}{234}{Bu }\myhl{0}{183}{234}{;}\myhl{0}{183}{234}{ "}\myhl{0}{183}{234}{The}\myhl{0}{183}{234}{ Four}\myhl{0}{183}{234}{ Great}\myhl{0}{183}{234}{ Const}\myhl{0}{183}{234}{ables}\myhl{0}{183}{234}{"),}\myhl{0}{183}{234}{ which}\myhl{0}{183}{234}{ has}\myhl{0}{183}{234}{ previously}\myhl{0}{183}{234}{ been}\myhl{0}{183}{234}{ adapted}\myhl{0}{183}{234}{ to}\myhl{0}{183}{234}{ a}\myhl{0}{183}{234}{ television}\myhl{0}{183}{234}{ series}\myhl{0}{183}{234}{.} \dots\myhl{149}{225}{246}{The}\myhl{149}{225}{246}{ Four}\myhl{149}{225}{246}{ III}\myhl{149}{225}{246}{ (}\myhl{149}{225}{246}{Chinese}\myhl{149}{225}{246}{:}\myhl{149}{225}{246}{ Si }\myhl{149}{225}{246}{Da }\myhl{149}{225}{246}{Ming }\myhl{149}{225}{246}{Bu }\myhl{149}{225}{246}{3}\myhl{149}{225}{246}{)}\myhl{149}{225}{246}{ is}\myhl{149}{225}{246}{ a}\myhl{149}{225}{246}{ Chinese}\myhl{149}{225}{246}{-H}\myhl{149}{225}{246}{ong}\myhl{149}{225}{246}{ Kong}\myhl{149}{225}{246}{ }\myhl{149}{225}{246}{3}\myhl{149}{225}{246}{D}\myhl{149}{225}{246}{ w}\myhl{149}{225}{246}{ux}\myhl{149}{225}{246}{ia}\myhl{149}{225}{246}{ film}\myhl{149}{225}{246}{ directed}\myhl{149}{225}{246}{ by}\myhl{149}{225}{246}{ Gordon}\myhl{149}{225}{246}{ Chan}\myhl{149}{225}{246}{ and}\myhl{149}{225}{246}{ Janet}\myhl{149}{225}{246}{ Chun}\myhl{149}{225}{246}{.} \dots\myhl{156}{227}{247}{ It}\myhl{156}{227}{247}{ is}\myhl{156}{227}{247}{ the}\myhl{156}{227}{247}{ final}\myhl{156}{227}{247}{ installment}\myhl{156}{227}{247}{ of}\myhl{156}{227}{247}{ the}\myhl{156}{227}{247}{ trilogy}\myhl{156}{227}{247}{ based}\myhl{156}{227}{247}{ on}\myhl{156}{227}{247}{ W}\myhl{156}{227}{247}{oon}\myhl{156}{227}{247}{ Swe}\myhl{156}{227}{247}{e}\myhl{156}{227}{247}{ O}\myhl{156}{227}{247}{an}\myhl{156}{227}{247}{'s}\myhl{156}{227}{247}{ novel}\myhl{156}{227}{247}{ series}\myhl{156}{227}{247}{,}\myhl{156}{227}{247}{ after}\myhl{156}{227}{247}{ "}\myhl{156}{227}{247}{The}\myhl{156}{227}{247}{ Four}\myhl{156}{227}{247}{"}\myhl{156}{227}{247}{ (}\myhl{156}{227}{247}{201}\myhl{156}{227}{247}{2}\myhl{156}{227}{247}{)}\myhl{156}{227}{247}{ and}\myhl{156}{227}{247}{ "}\myhl{156}{227}{247}{The}\myhl{156}{227}{247}{ Four}\myhl{156}{227}{247}{ II}\myhl{156}{227}{247}{"}\myhl{156}{227}{247}{ (}\myhl{156}{227}{247}{201}\myhl{156}{227}{247}{3}\myhl{156}{227}{247}{).}
\end{tcolorbox}
\begin{tcolorbox}[colback=white, , title=ESM (256 samples)]
\textbf{Context}:
 \dots\myhl{0}{183}{234}{ It}\myhl{0}{183}{234}{ is}\myhl{0}{183}{234}{ the}\myhl{0}{183}{234}{ first}\myhl{0}{183}{234}{ film}\myhl{0}{183}{234}{ adaptation}\myhl{0}{183}{234}{ of}\myhl{0}{183}{234}{ W}\myhl{0}{183}{234}{oon}\myhl{0}{183}{234}{ Swe}\myhl{0}{183}{234}{e}\myhl{0}{183}{234}{ O}\myhl{0}{183}{234}{an}\myhl{0}{183}{234}{'s}\myhl{0}{183}{234}{ novel}\myhl{0}{183}{234}{ series}\myhl{0}{183}{234}{ "}\myhl{0}{183}{234}{Si}\myhl{0}{183}{234}{ Da}\myhl{0}{183}{234}{ Ming}\myhl{0}{183}{234}{ Bu}\myhl{0}{183}{234}{"}\myhl{0}{183}{234}{ (}\myhl{0}{183}{234}{Si }\myhl{0}{183}{234}{Da }\myhl{0}{183}{234}{Ming }\myhl{0}{183}{234}{Bu }\myhl{0}{183}{234}{;}\myhl{0}{183}{234}{ "}\myhl{0}{183}{234}{The}\myhl{0}{183}{234}{ Four}\myhl{0}{183}{234}{ Great}\myhl{0}{183}{234}{ Const}\myhl{0}{183}{234}{ables}\myhl{0}{183}{234}{"),}\myhl{0}{183}{234}{ which}\myhl{0}{183}{234}{ has}\myhl{0}{183}{234}{ previously}\myhl{0}{183}{234}{ been}\myhl{0}{183}{234}{ adapted}\myhl{0}{183}{234}{ to}\myhl{0}{183}{234}{ a}\myhl{0}{183}{234}{ television}\myhl{0}{183}{234}{ series}\myhl{0}{183}{234}{.} \dots
\end{tcolorbox}
\textbf{Query}:
What is the English translation of the name of the novel series which the Four III is the final installment of?
\newline\newline\textbf{Response} (from \texttt{Llama-3.1-8B}):
\newline
\myhl{249}{237}{91}{The novel series is "Si Da Ming Bu" (Si Da Ming Bu ; "The Four Great Constables").}
\end{tcolorbox}
\relsize{1}
\caption{Visualized attributions for a random example from Hotpot QA for \texttt{Llama-3.1-8B} with sentences as sources.}
\label{fig:hotpot_qa_passages_llama_3_1_8b_sentence}
\end{figure*}

\begin{figure*}[t]
\centering\definecolor{forestgreen}{RGB}{34, 139, 34}
\relsize{-1}
\begin{tcolorbox}[colback=white]
\textbf{Query}:
Thinking about your country's future, how important is it that: e. Copts and other religious minorities can practice their religion freely? Is it very important, somewhat important, not too important or not important at all?
\begin{tcolorbox}[colback=white, colframe=forestgreen, title=\ours (general)]
\textbf{Thoughts}:
 \dots\myhl{254}{254}{254}{ the}\myhl{255}{255}{255}{ nation}\myhl{255}{255}{255}{'s}\myhl{239}{250}{253}{ unity}\myhl{253}{254}{254}{ or}\myhl{197}{238}{250}{ harmony}\myhl{255}{255}{255}{?}\myhl{255}{255}{255}{ On}\myhl{255}{255}{255}{ one}\myhl{254}{254}{254}{ hand}\myhl{252}{254}{254}{,} \dots\myhl{250}{253}{254}{istic}\myhl{238}{250}{253}{ society}\myhl{251}{253}{254}{,}\myhl{246}{252}{254}{ which}\myhl{249}{253}{254}{ might}\myhl{83}{207}{241}{ foster}\myhl{247}{252}{254}{ tolerance}\myhl{252}{254}{254}{ and}\myhl{251}{253}{254}{ reduce}\myhl{255}{255}{255}{ conflicts}\myhl{255}{255}{255}{.} \dots\myhl{244}{251}{254}{ would}\myhl{246}{252}{254}{ help}\myhl{253}{254}{254}{ in}\myhl{255}{255}{255}{ maintaining}\myhl{235}{249}{253}{ social}\myhl{186}{235}{249}{ harmony}\myhl{228}{247}{252}{,}\myhl{208}{242}{251}{ economic}\myhl{156}{227}{247}{ diversity}\myhl{209}{242}{251}{,}\myhl{236}{249}{253}{ and}\myhl{171}{231}{248}{ political}\myhl{204}{240}{250}{ stability}\myhl{236}{249}{253}{.}\myhl{252}{254}{254}{But}\myhl{254}{254}{254}{ I}\myhl{250}{253}{254}{'m} \dots\myhl{255}{255}{255}{ future}\myhl{230}{248}{253}{.}\myhl{210}{242}{251}{ It}\myhl{240}{250}{253}{ contributes}\myhl{255}{255}{255}{ to}\myhl{171}{231}{248}{ social}\myhl{160}{228}{247}{ harmony}\myhl{184}{235}{249}{,}\myhl{176}{232}{248}{ economic}\myhl{148}{225}{246}{ diversity}\myhl{185}{235}{249}{,}\myhl{186}{235}{249}{ and}\myhl{133}{220}{245}{ political}\myhl{130}{220}{245}{ stability}\myhl{189}{236}{249}{,}\myhl{174}{232}{248}{ which}\myhl{166}{230}{247}{ are}\myhl{218}{244}{252}{ all}\myhl{57}{199}{239}{ essential}\myhl{223}{246}{252}{ for}\myhl{214}{243}{251}{ a}\myhl{0}{183}{234}{ prosperous}\myhl{194}{237}{250}{ and}\myhl{112}{215}{243}{ peaceful}\myhl{140}{222}{245}{ society}\myhl{162}{229}{247}{.}
\end{tcolorbox}
\begin{tcolorbox}[colback=white, , title=Average attention]
\textbf{Thoughts}:
 \dots\myhl{238}{250}{253}{ positively}\myhl{242}{251}{254}{ to}\myhl{240}{250}{253}{ the}\myhl{235}{249}{253}{ nation}\myhl{249}{253}{254}{'s}\myhl{202}{240}{250}{ unity}\myhl{240}{250}{253}{ or}\myhl{208}{241}{251}{ harmony}\myhl{231}{248}{253}{?}\myhl{249}{253}{254}{ On}\myhl{252}{254}{254}{ one} \dots\myhl{238}{250}{253}{ would}\myhl{230}{248}{253}{ help}\myhl{236}{249}{253}{ in}\myhl{227}{247}{252}{ maintaining}\myhl{225}{246}{252}{ social}\myhl{185}{235}{249}{ harmony}\myhl{211}{242}{251}{,}\myhl{210}{242}{251}{ economic}\myhl{171}{231}{248}{ diversity}\myhl{189}{236}{249}{,}\myhl{224}{246}{252}{ and}\myhl{199}{239}{250}{ political}\myhl{176}{232}{248}{ stability}\myhl{218}{244}{252}{.}\myhl{246}{252}{254}{But}\myhl{249}{253}{254}{ I}\myhl{250}{253}{254}{'m}\myhl{250}{253}{254}{ not} \dots\myhl{238}{250}{253}{ might}\myhl{248}{253}{254}{ be}\myhl{234}{249}{253}{ crucial}\myhl{234}{249}{253}{ for}\myhl{227}{247}{252}{ a}\myhl{193}{237}{250}{ peaceful}\myhl{226}{246}{252}{ and}\myhl{224}{246}{252}{ stable}\myhl{199}{239}{250}{ society}\myhl{217}{244}{251}{.}\myhl{243}{251}{254}{I}\myhl{247}{252}{254}{ should}\myhl{246}{252}{254}{ also}\myhl{246}{252}{254}{ think} \dots\myhl{242}{251}{254}{ often}\myhl{241}{251}{253}{ had}\myhl{239}{250}{253}{ more}\myhl{228}{247}{252}{ stable}\myhl{231}{248}{253}{ and}\myhl{196}{238}{250}{ prosperous}\myhl{209}{242}{251}{ societies}\myhl{202}{240}{250}{.}\myhl{239}{250}{253}{ Conversely}\myhl{236}{249}{253}{,}\myhl{246}{252}{254}{ those}\myhl{245}{252}{254}{ that}\myhl{248}{253}{254}{ have} \dots\myhl{235}{249}{253}{ in}\myhl{232}{248}{253}{ a}\myhl{232}{248}{253}{ country}\myhl{235}{249}{253}{'s}\myhl{215}{244}{251}{ future}\myhl{193}{237}{250}{.}\myhl{238}{250}{253}{In}\myhl{222}{245}{252}{ conclusion}\myhl{215}{244}{251}{,}\myhl{224}{246}{252}{ ensuring}\myhl{237}{250}{253}{ that} \dots\myhl{213}{243}{251}{ for}\myhl{218}{244}{252}{ a}\myhl{220}{245}{252}{ country}\myhl{232}{248}{253}{'s}\myhl{215}{243}{251}{ future}\myhl{161}{228}{247}{.}\myhl{197}{238}{250}{ It}\myhl{170}{231}{248}{ contributes}\myhl{192}{237}{250}{ to}\myhl{172}{231}{248}{ social}\myhl{114}{215}{243}{ harmony}\myhl{144}{224}{246}{,}\myhl{162}{229}{247}{ economic}\myhl{130}{220}{245}{ diversity}\myhl{150}{225}{246}{,}\myhl{153}{226}{246}{ and}\myhl{132}{220}{245}{ political}\myhl{73}{204}{240}{ stability}\myhl{132}{220}{245}{,}\myhl{140}{222}{245}{ which}\myhl{156}{227}{247}{ are}\myhl{156}{227}{247}{ all}\myhl{83}{207}{241}{ essential}\myhl{113}{215}{243}{ for}\myhl{167}{230}{247}{ a}\myhl{0}{183}{234}{ prosperous}\myhl{119}{217}{244}{ and}\myhl{16}{188}{235}{ peaceful}\myhl{58}{199}{239}{ society}\myhl{67}{202}{240}{.}
\end{tcolorbox}
\begin{tcolorbox}[colback=white, , title=ESM (256 samples)]
\textbf{Thoughts}:
 \dots\myhl{255}{255}{255}{ religion}\myhl{255}{255}{255}{ might}\myhl{244}{252}{254}{ contribute}\myhl{255}{255}{255}{ to}\myhl{255}{255}{255}{ a}\myhl{159}{228}{247}{ diverse}\myhl{255}{255}{255}{ workforce}\myhl{255}{255}{255}{,}\myhl{255}{255}{255}{ which}\myhl{255}{255}{255}{ is}\myhl{255}{255}{255}{ beneficial} \dots\myhl{255}{255}{255}{ ability}\myhl{255}{255}{255}{ to}\myhl{255}{255}{255}{ practice}\myhl{255}{255}{255}{ freely}\myhl{255}{255}{255}{ can}\myhl{192}{237}{250}{ influence}\myhl{255}{255}{255}{ international}\myhl{255}{255}{255}{ relations}\myhl{255}{255}{255}{,}\myhl{247}{252}{254}{ though}\myhl{255}{255}{255}{ that} \dots\myhl{255}{255}{255}{ This}\myhl{255}{255}{255}{ would}\myhl{255}{255}{255}{ help}\myhl{255}{255}{255}{ in}\myhl{255}{255}{255}{ maintaining}\myhl{183}{234}{249}{ social}\myhl{255}{255}{255}{ harmony}\myhl{213}{243}{251}{,}\myhl{193}{237}{250}{ economic}\myhl{152}{226}{246}{ diversity}\myhl{233}{249}{253}{,}\myhl{255}{255}{255}{ and}\myhl{202}{240}{250}{ political}\myhl{255}{255}{255}{ stability}\myhl{255}{255}{255}{.}\myhl{255}{255}{255}{But}\myhl{255}{255}{255}{ I}\myhl{255}{255}{255}{'m} \dots\myhl{255}{255}{255}{ contributes}\myhl{255}{255}{255}{ to}\myhl{216}{244}{251}{ social}\myhl{255}{255}{255}{ harmony}\myhl{255}{255}{255}{,}\myhl{18}{188}{236}{ economic}\myhl{228}{247}{252}{ diversity}\myhl{255}{255}{255}{,}\myhl{241}{251}{253}{ and}\myhl{176}{232}{248}{ political}\myhl{255}{255}{255}{ stability}\myhl{255}{255}{255}{,}\myhl{255}{255}{255}{ which}\myhl{255}{255}{255}{ are}\myhl{255}{255}{255}{ all}\myhl{255}{255}{255}{ essential}\myhl{255}{255}{255}{ for}\myhl{203}{240}{250}{ a}\myhl{0}{183}{234}{ prosperous}\myhl{255}{255}{255}{ and}\myhl{255}{255}{255}{ peaceful}\myhl{255}{255}{255}{ society}\myhl{255}{255}{255}{.}
\end{tcolorbox}
\textbf{Response} (from \texttt{DeepSeek-R1-Qwen-7B}):
\newline
Ensuring that Copts and other religious minorities can practice their religion freely is very important for a country's future. \myhl{249}{237}{91}{It fosters social harmony, economic diversity, and political stability, which are essential for a prosperous and peaceful society.}
\end{tcolorbox}
\relsize{1}
\caption{Visualized attributions for a random example from Global Opinions QA for \texttt{DeepSeek-R1-Qwen-7B} with tokens as sources.}
\label{fig:global_opinions_r1_qwen_7b_dtype=bfloat16_token_0}
\end{figure*}

\begin{figure*}[t]
\centering\definecolor{forestgreen}{RGB}{34, 139, 34}
\relsize{-1}
\begin{tcolorbox}[colback=white]
\textbf{Query}:
Thinking about the United Nations, what kind of influence is the UN having on the way things are going in your country.   Is the influence of the UN very good, somewhat good, somewhat bad or very bad in your country?
\begin{tcolorbox}[colback=white, colframe=forestgreen, title=\ours (general)]
\textbf{Thoughts}:
 \dots\myhl{239}{250}{253}{ the}\myhl{255}{255}{255}{ influence}\myhl{249}{253}{254}{ of}\myhl{223}{246}{252}{ the}\myhl{241}{251}{253}{ United}\myhl{178}{233}{248}{ Nations}\myhl{255}{255}{255}{ on}\myhl{248}{253}{254}{ their}\myhl{251}{254}{254}{ country}\myhl{234}{249}{253}{.}\myhl{253}{254}{254}{ They} \dots\myhl{252}{254}{254}{'s}\myhl{237}{250}{253}{ heavily}\myhl{243}{251}{254}{ involved}\myhl{240}{250}{253}{ with}\myhl{251}{254}{254}{ the}\myhl{182}{234}{249}{ UN}\myhl{244}{251}{254}{,}\myhl{255}{255}{255}{ or}\myhl{254}{254}{254}{ maybe}\myhl{255}{255}{255}{ they}\myhl{250}{253}{254}{'re} \dots\myhl{250}{253}{254}{ general}\myhl{252}{254}{254}{ enough}\myhl{253}{254}{254}{ to}\myhl{250}{253}{254}{ apply}\myhl{254}{254}{254}{ to}\myhl{183}{235}{249}{ many}\myhl{230}{248}{253}{ countries}\myhl{255}{255}{255}{ but}\myhl{252}{254}{254}{ specific}\myhl{254}{254}{254}{ enough}\myhl{255}{255}{255}{ to} \dots\myhl{228}{247}{252}{.}\myhl{252}{254}{254}{I}\myhl{244}{252}{254}{ remember}\myhl{251}{254}{254}{ that}\myhl{248}{253}{254}{ the}\myhl{202}{240}{250}{ UN}\myhl{253}{254}{254}{ has}\myhl{237}{250}{253}{ a}\myhl{247}{253}{254}{ significant}\myhl{255}{255}{255}{ impact}\myhl{253}{254}{254}{ on}\myhl{220}{245}{252}{ many}\myhl{137}{222}{245}{ nations}\myhl{242}{251}{253}{,}\myhl{251}{253}{254}{ especially}\myhl{250}{253}{254}{ in}\myhl{252}{254}{254}{ terms}\myhl{246}{252}{254}{ of} \dots\myhl{229}{247}{252}{ often}\myhl{241}{251}{253}{ rely}\myhl{254}{254}{254}{ on}\myhl{245}{252}{254}{ the}\myhl{216}{244}{251}{ UN}\myhl{172}{231}{248}{ for}\myhl{209}{242}{251}{ peace}\myhl{251}{254}{254}{keeping}\myhl{254}{254}{254}{ missions}\myhl{224}{246}{252}{,}\myhl{239}{250}{253}{ like} \dots\myhl{247}{252}{254}{ to}\myhl{253}{254}{254}{ respond}\myhl{253}{254}{254}{,}\myhl{254}{254}{254}{ especially}\myhl{255}{255}{255}{ in}\myhl{103}{212}{242}{ crises}\myhl{244}{252}{254}{.}\myhl{248}{253}{254}{ Also}\myhl{255}{255}{255}{,}\myhl{248}{253}{254}{ member}\myhl{255}{255}{255}{ states} \dots\myhl{245}{252}{254}{ the}\myhl{240}{250}{253}{ UN}\myhl{255}{255}{255}{ promotes}\myhl{255}{255}{255}{ cooperation}\myhl{252}{254}{254}{ and}\myhl{172}{231}{248}{ addresses}\myhl{83}{207}{241}{ global}\myhl{0}{183}{234}{ issues}\myhl{226}{247}{252}{,}\myhl{254}{254}{254}{ but}\myhl{252}{254}{254}{ also}\myhl{253}{254}{254}{ note}\myhl{248}{253}{254}{ the} \dots
\end{tcolorbox}
\begin{tcolorbox}[colback=white, , title=Average attention]
\textbf{Thoughts}:
\myhl{233}{249}{253}{Okay}\myhl{203}{240}{250}{,}\myhl{235}{249}{253}{ so}\myhl{222}{245}{252}{ the}\myhl{222}{245}{252}{ user}\myhl{243}{251}{254}{ is}\myhl{239}{250}{253}{ asking}\myhl{235}{249}{253}{ about}\myhl{198}{239}{250}{ the}\myhl{191}{237}{249}{ influence}\myhl{219}{244}{252}{ of}\myhl{173}{232}{248}{ the}\myhl{198}{239}{250}{ United}\myhl{52}{198}{238}{ Nations}\myhl{171}{231}{248}{ on}\myhl{216}{244}{251}{ their}\myhl{139}{222}{245}{ country}\myhl{205}{241}{251}{.}\myhl{242}{251}{254}{ They}\myhl{244}{252}{254}{ want}\myhl{238}{250}{253}{ a}\myhl{216}{244}{251}{ paragraph} \dots\myhl{233}{248}{253}{.}\myhl{245}{252}{254}{ They}\myhl{245}{252}{254}{ might}\myhl{245}{252}{254}{ be}\myhl{235}{249}{253}{ a}\myhl{189}{236}{249}{ citizen}\myhl{223}{246}{252}{ of}\myhl{214}{243}{251}{ a}\myhl{207}{241}{251}{ country}\myhl{218}{244}{252}{ that}\myhl{234}{249}{253}{'s}\myhl{219}{245}{252}{ heavily}\myhl{210}{242}{251}{ involved}\myhl{184}{235}{249}{ with}\myhl{194}{238}{250}{ the}\myhl{80}{206}{241}{ UN}\myhl{200}{239}{250}{,}\myhl{235}{249}{253}{ or}\myhl{242}{251}{254}{ maybe}\myhl{243}{251}{254}{ they}\myhl{244}{252}{254}{'re}\myhl{230}{248}{253}{ from}\myhl{231}{248}{253}{ a}\myhl{228}{247}{252}{ less}\myhl{200}{239}{250}{ involved}\myhl{209}{242}{251}{ country}\myhl{223}{246}{252}{.}\myhl{238}{250}{253}{ The}\myhl{241}{251}{253}{ response}\myhl{247}{252}{254}{ should} \dots\myhl{244}{252}{254}{ enough}\myhl{246}{252}{254}{ to}\myhl{241}{251}{253}{ apply}\myhl{237}{250}{253}{ to}\myhl{225}{246}{252}{ many}\myhl{175}{232}{248}{ countries}\myhl{239}{250}{253}{ but}\myhl{241}{251}{253}{ specific}\myhl{247}{252}{254}{ enough}\myhl{248}{253}{254}{ to}\myhl{249}{253}{254}{ be} \dots\myhl{219}{245}{252}{.}\myhl{244}{252}{254}{I}\myhl{239}{250}{253}{ remember}\myhl{226}{247}{252}{ that}\myhl{205}{241}{251}{ the}\myhl{105}{213}{243}{ UN}\myhl{214}{243}{251}{ has}\myhl{219}{245}{252}{ a}\myhl{224}{246}{252}{ significant}\myhl{193}{237}{250}{ impact}\myhl{178}{233}{248}{ on}\myhl{220}{245}{252}{ many}\myhl{120}{217}{244}{ nations}\myhl{186}{235}{249}{,}\myhl{225}{246}{252}{ especially}\myhl{163}{229}{247}{ in}\myhl{227}{247}{252}{ terms}\myhl{148}{225}{246}{ of}\myhl{51}{197}{238}{ international}\myhl{22}{189}{236}{ law}\myhl{103}{212}{242}{ and}\myhl{42}{195}{238}{ security}\myhl{149}{225}{246}{.}\myhl{184}{235}{249}{ Countries}\myhl{207}{241}{251}{ often}\myhl{178}{233}{248}{ rely}\myhl{203}{240}{250}{ on}\myhl{179}{233}{248}{ the}\myhl{123}{218}{244}{ UN}\myhl{148}{225}{246}{ for}\myhl{75}{204}{240}{ peace}\myhl{82}{206}{241}{keeping}\myhl{108}{213}{243}{ missions}\myhl{107}{213}{243}{,}\myhl{171}{231}{248}{ like}\myhl{170}{231}{248}{ in}\myhl{124}{218}{244}{ Afghanistan}\myhl{149}{225}{246}{ or}\myhl{224}{246}{252}{ South}\myhl{178}{233}{248}{ Sudan}\myhl{143}{223}{246}{.}\myhl{232}{248}{253}{ That}\myhl{239}{250}{253}{'s}\myhl{235}{249}{253}{ a}\myhl{201}{239}{250}{ positive}\myhl{224}{246}{252}{ aspect}\myhl{188}{236}{249}{.}\myhl{238}{250}{253}{On}\myhl{222}{245}{252}{ the}\myhl{218}{244}{252}{ other}\myhl{230}{248}{253}{ hand}\myhl{197}{238}{250}{,}\myhl{233}{249}{253}{ there}\myhl{234}{249}{253}{ are}\myhl{208}{241}{251}{ challenges}\myhl{202}{240}{250}{.}\myhl{215}{243}{251}{ The}\myhl{152}{226}{246}{ UN}\myhl{217}{244}{251}{ can}\myhl{233}{249}{253}{ be}\myhl{192}{237}{250}{ slow}\myhl{212}{243}{251}{ to}\myhl{180}{234}{249}{ respond}\myhl{198}{239}{250}{,}\myhl{234}{249}{253}{ especially}\myhl{202}{240}{250}{ in}\myhl{0}{183}{234}{ crises}\myhl{193}{237}{250}{.}\myhl{231}{248}{253}{ Also}\myhl{215}{243}{251}{,}\myhl{175}{232}{248}{ member}\myhl{187}{236}{249}{ states}\myhl{232}{248}{253}{ might}\myhl{235}{249}{253}{ not}\myhl{235}{249}{253}{ always}\myhl{210}{242}{251}{ agree}\myhl{198}{239}{250}{,}\myhl{237}{250}{253}{ leading}\myhl{227}{247}{252}{ to}\myhl{224}{246}{252}{ issues}\myhl{214}{243}{251}{ like}\myhl{187}{236}{249}{ the}\myhl{182}{234}{249}{ situation}\myhl{187}{236}{249}{ in}\myhl{113}{215}{243}{ Syria}\myhl{161}{228}{247}{.}\myhl{237}{250}{253}{ That}\myhl{242}{251}{254}{'s}\myhl{239}{250}{253}{ a}\myhl{229}{247}{252}{ negative}\myhl{237}{250}{253}{ point}\myhl{196}{238}{250}{.}\myhl{240}{250}{253}{I}\myhl{244}{252}{254}{ should}\myhl{230}{248}{253}{ mention}\myhl{228}{247}{252}{ how}\myhl{205}{241}{251}{ the}\myhl{186}{235}{249}{ UN}\myhl{125}{218}{244}{ promotes}\myhl{116}{216}{243}{ cooperation}\myhl{190}{236}{249}{ and}\myhl{115}{215}{243}{ addresses}\myhl{48}{197}{238}{ global}\myhl{35}{193}{237}{ issues}\myhl{158}{227}{247}{,}\myhl{223}{246}{252}{ but}\myhl{238}{250}{253}{ also}\myhl{241}{251}{253}{ note}\myhl{220}{245}{252}{ the}\myhl{214}{243}{251}{ difficulties}\myhl{224}{246}{252}{ in}\myhl{152}{226}{246}{ implementation}\myhl{212}{243}{251}{.}\myhl{236}{249}{253}{ It}\myhl{243}{251}{254}{'s}\myhl{243}{251}{254}{ important}\myhl{243}{251}{254}{ to} \dots\myhl{241}{251}{253}{ it}\myhl{241}{251}{253}{ all}\myhl{239}{250}{253}{ together}\myhl{217}{244}{251}{:}\myhl{220}{245}{252}{ acknowledge}\myhl{177}{233}{248}{ the}\myhl{217}{244}{251}{ positive}\myhl{217}{244}{251}{ aspects}\myhl{201}{240}{250}{ like}\myhl{130}{220}{245}{ peace}\myhl{118}{216}{244}{keeping}\myhl{139}{222}{245}{ and}\myhl{140}{222}{245}{ international}\myhl{109}{214}{243}{ law}\myhl{197}{238}{250}{,}\myhl{235}{249}{253}{ then}\myhl{230}{248}{253}{ mention}\myhl{203}{240}{250}{ the}\myhl{212}{243}{251}{ challenges}\myhl{215}{243}{251}{ like}\myhl{211}{242}{251}{ slow}\myhl{193}{237}{250}{ response}\myhl{206}{241}{251}{ and}\myhl{187}{236}{249}{ political}\myhl{168}{230}{248}{ disagreements}\myhl{192}{237}{249}{.}\myhl{239}{250}{253}{ Con}\myhl{233}{249}{253}{clude}\myhl{235}{249}{253}{ by}\myhl{235}{249}{253}{ saying}\myhl{205}{241}{251}{ the} \dots\myhl{229}{247}{252}{ is}\myhl{210}{242}{251}{ mixed}\myhl{207}{241}{251}{,}\myhl{222}{246}{252}{ depending}\myhl{222}{245}{252}{ on}\myhl{188}{236}{249}{ the}\myhl{158}{228}{247}{ country}\myhl{220}{245}{252}{'s}\myhl{208}{241}{251}{ context}\myhl{187}{236}{249}{.}\myhl{230}{248}{253}{I}\myhl{235}{249}{253}{ should}\myhl{239}{250}{253}{ make}\myhl{237}{250}{253}{ sure}\myhl{213}{243}{251}{ the} \dots\myhl{229}{247}{252}{ j}\myhl{224}{246}{252}{argon}\myhl{232}{248}{253}{ that}\myhl{238}{250}{253}{ might}\myhl{229}{247}{252}{ confuse}\myhl{185}{235}{249}{ the}\myhl{221}{245}{252}{ user}\myhl{190}{236}{249}{.}\myhl{218}{244}{252}{ The}\myhl{231}{248}{253}{ goal}\myhl{232}{248}{253}{ is}\myhl{228}{247}{252}{ to}\myhl{228}{247}{252}{ provide}\myhl{210}{242}{251}{ a}\myhl{218}{244}{252}{ balanced}\myhl{223}{246}{252}{ view}\myhl{224}{246}{252}{ that}\myhl{191}{237}{249}{ the}\myhl{226}{246}{252}{ user}\myhl{231}{248}{253}{ can}\myhl{223}{246}{252}{ understand}\myhl{220}{245}{252}{ and}\myhl{224}{246}{252}{ relate}\myhl{225}{246}{252}{ to}\myhl{144}{224}{246}{.}
\end{tcolorbox}
\begin{tcolorbox}[colback=white, , title=ESM (256 samples)]
\textbf{Thoughts}:
\myhl{255}{255}{255}{Okay}\myhl{255}{255}{255}{,}\myhl{255}{255}{255}{ so}\myhl{255}{255}{255}{ the}\myhl{0}{183}{234}{ user}\myhl{255}{255}{255}{ is}\myhl{255}{255}{255}{ asking}\myhl{255}{255}{255}{ about}\myhl{255}{255}{255}{ the}\myhl{255}{255}{255}{ influence} \dots\myhl{255}{255}{255}{ and}\myhl{255}{255}{255}{ security}\myhl{255}{255}{255}{.}\myhl{255}{255}{255}{ Countries}\myhl{255}{255}{255}{ often}\myhl{136}{221}{245}{ rely}\myhl{255}{255}{255}{ on}\myhl{255}{255}{255}{ the}\myhl{255}{255}{255}{ UN}\myhl{247}{252}{254}{ for}\myhl{227}{247}{252}{ peace}\myhl{255}{255}{255}{keeping}\myhl{199}{239}{250}{ missions}\myhl{255}{255}{255}{,}\myhl{248}{253}{254}{ like}\myhl{255}{255}{255}{ in}\myhl{255}{255}{255}{ Afghanistan}\myhl{255}{255}{255}{ or} \dots
\end{tcolorbox}
\textbf{Response} (from \texttt{DeepSeek-R1-Qwen-7B}):
\newline
The United Nations plays a significant role in shaping the global stage, particularly in areas such as international law, security, and cooperation. Its influence can be both positive and challenging, depending on the country and context. \myhl{249}{237}{91}{For many nations, the UN provides a framework for addressing global issues, such as climate change, human rights, and humanitarian crises.} It also facilitates peacekeeping missions and promotes collective security, which can enhance stability in certain regions. However, the UN's effectiveness is sometimes hindered by slow decision-making, political disagreements among member states, and the difficulty of implementing its resolutions. Overall, the UN's impact is mixed, with its benefits and challenges varying widely across different countries.
\end{tcolorbox}
\relsize{1}
\caption{Visualized attributions for a random example from Global Opinions QA for \texttt{DeepSeek-R1-Qwen-7B} with tokens as sources.}
\label{fig:global_opinions_r1_qwen_7b_dtype=bfloat16_token_1}
\end{figure*}

\begin{figure*}[t]
\centering\definecolor{forestgreen}{RGB}{34, 139, 34}
\relsize{-1}
\begin{tcolorbox}[colback=white]
\textbf{Query}:
In your view, how often do the following things occur in this country's elections?TV news favors the governing party
\begin{tcolorbox}[colback=white, colframe=forestgreen, title=\ours (general)]
\textbf{Thoughts}:
 \dots\myhl{32}{192}{237}{ I}\myhl{32}{192}{237}{'m}\myhl{32}{192}{237}{ not}\myhl{32}{192}{237}{ sure}\myhl{32}{192}{237}{ how}\myhl{32}{192}{237}{ much}\myhl{32}{192}{237}{ the}\myhl{32}{192}{237}{ media}\myhl{32}{192}{237}{ actually}\myhl{32}{192}{237}{ sw}\myhl{32}{192}{237}{ays}\myhl{32}{192}{237}{ public}\myhl{32}{192}{237}{ opinion}\myhl{32}{192}{237}{ or}\myhl{32}{192}{237}{ if}\myhl{32}{192}{237}{ it}\myhl{32}{192}{237}{'s}\myhl{32}{192}{237}{ just}\myhl{32}{192}{237}{ a}\myhl{32}{192}{237}{ perception}\myhl{32}{192}{237}{.}\myhl{183}{235}{249}{ Maybe}\myhl{183}{235}{249}{ there}\myhl{183}{235}{249}{'s}\myhl{183}{235}{249}{ some}\myhl{183}{235}{249}{ bias}\myhl{183}{235}{249}{,}\myhl{183}{235}{249}{ but}\myhl{183}{235}{249}{ is}\myhl{183}{235}{249}{ it}\myhl{183}{235}{249}{ significant}\myhl{183}{235}{249}{?} \dots\myhl{189}{236}{249}{ I}\myhl{189}{236}{249}{ think}\myhl{189}{236}{249}{ in}\myhl{189}{236}{249}{ many}\myhl{189}{236}{249}{ countries}\myhl{189}{236}{249}{,}\myhl{189}{236}{249}{ media}\myhl{189}{236}{249}{ outlets}\myhl{189}{236}{249}{ have}\myhl{189}{236}{249}{ clear}\myhl{189}{236}{249}{ st}\myhl{189}{236}{249}{ances}\myhl{189}{236}{249}{ on}\myhl{189}{236}{249}{ political}\myhl{189}{236}{249}{ issues}\myhl{189}{236}{249}{,}\myhl{189}{236}{249}{ so}\myhl{189}{236}{249}{ it}\myhl{189}{236}{249}{'s}\myhl{189}{236}{249}{ possible}\myhl{189}{236}{249}{ they}\myhl{189}{236}{249}{ lean}\myhl{189}{236}{249}{ towards}\myhl{189}{236}{249}{ whichever}\myhl{189}{236}{249}{ party}\myhl{189}{236}{249}{ is}\myhl{189}{236}{249}{ in}\myhl{189}{236}{249}{ power}\myhl{189}{236}{249}{.} \dots\myhl{167}{230}{247}{ If}\myhl{167}{230}{247}{ the}\myhl{167}{230}{247}{ governing}\myhl{167}{230}{247}{ party}\myhl{167}{230}{247}{ has}\myhl{167}{230}{247}{ more}\myhl{167}{230}{247}{ funding}\myhl{167}{230}{247}{,}\myhl{167}{230}{247}{ they}\myhl{167}{230}{247}{ might}\myhl{167}{230}{247}{ get}\myhl{167}{230}{247}{ more}\myhl{167}{230}{247}{ air}\myhl{167}{230}{247}{time}\myhl{167}{230}{247}{.} \dots\myhl{192}{237}{250}{ However}\myhl{192}{237}{250}{,}\myhl{192}{237}{250}{ in}\myhl{192}{237}{250}{ some}\myhl{192}{237}{250}{ cases}\myhl{192}{237}{250}{,}\myhl{192}{237}{250}{ especially}\myhl{192}{237}{250}{ in}\myhl{192}{237}{250}{ less}\myhl{192}{237}{250}{ competitive}\myhl{192}{237}{250}{ elections}\myhl{192}{237}{250}{,}\myhl{192}{237}{250}{ the}\myhl{192}{237}{250}{ media}\myhl{192}{237}{250}{ might}\myhl{192}{237}{250}{ focus}\myhl{192}{237}{250}{ more}\myhl{192}{237}{250}{ on}\myhl{192}{237}{250}{ the}\myhl{192}{237}{250}{ ruling}\myhl{192}{237}{250}{ party}\myhl{192}{237}{250}{.}\myhl{176}{233}{248}{Another}\myhl{176}{233}{248}{ angle}\myhl{176}{233}{248}{ is}\myhl{176}{233}{248}{ the}\myhl{176}{233}{248}{ impact}\myhl{176}{233}{248}{ of}\myhl{176}{233}{248}{ recent}\myhl{176}{233}{248}{ events}\myhl{176}{233}{248}{.} \dots\myhl{0}{183}{234}{ So}\myhl{0}{183}{234}{ the}\myhl{0}{183}{234}{ frequency}\myhl{0}{183}{234}{ could}\myhl{0}{183}{234}{ vary}\myhl{0}{183}{234}{ based}\myhl{0}{183}{234}{ on}\myhl{0}{183}{234}{ the}\myhl{0}{183}{234}{ political}\myhl{0}{183}{234}{ climate}\myhl{0}{183}{234}{.}\myhl{139}{222}{245}{I}\myhl{139}{222}{245}{'m}\myhl{139}{222}{245}{ also}\myhl{139}{222}{245}{ thinking}\myhl{139}{222}{245}{ about}\myhl{139}{222}{245}{ the}\myhl{139}{222}{245}{ electoral}\myhl{139}{222}{245}{ system}\myhl{139}{222}{245}{.}\myhl{176}{232}{248}{ In}\myhl{176}{232}{248}{ some}\myhl{176}{232}{248}{ systems}\myhl{176}{232}{248}{,}\myhl{176}{232}{248}{ like}\myhl{176}{232}{248}{ first}\myhl{176}{232}{248}{-p}\myhl{176}{232}{248}{ast}\myhl{176}{232}{248}{-the}\myhl{176}{232}{248}{-post}\myhl{176}{232}{248}{,}\myhl{176}{232}{248}{ the}\myhl{176}{232}{248}{ ruling}\myhl{176}{232}{248}{ party}\myhl{176}{232}{248}{ tends}\myhl{176}{232}{248}{ to}\myhl{176}{232}{248}{ win}\myhl{176}{232}{248}{ more}\myhl{176}{232}{248}{ seats}\myhl{176}{232}{248}{,}\myhl{176}{232}{248}{ which}\myhl{176}{232}{248}{ might}\myhl{176}{232}{248}{ lead}\myhl{176}{232}{248}{ to}\myhl{176}{232}{248}{ more}\myhl{176}{232}{248}{ media}\myhl{176}{232}{248}{ coverage}\myhl{176}{232}{248}{.}\myhl{156}{227}{247}{ In}\myhl{156}{227}{247}{ other}\myhl{156}{227}{247}{ systems}\myhl{156}{227}{247}{,}\myhl{156}{227}{247}{ like}\myhl{156}{227}{247}{ proportional}\myhl{156}{227}{247}{ representation}\myhl{156}{227}{247}{,}\myhl{156}{227}{247}{ the}\myhl{156}{227}{247}{ media}\myhl{156}{227}{247}{ might}\myhl{156}{227}{247}{ cover}\myhl{156}{227}{247}{ all}\myhl{156}{227}{247}{ parties}\myhl{156}{227}{247}{ more}\myhl{156}{227}{247}{ equally}\myhl{156}{227}{247}{,}\myhl{156}{227}{247}{ but}\myhl{156}{227}{247}{ I}\myhl{156}{227}{247}{'m}\myhl{156}{227}{247}{ not}\myhl{156}{227}{247}{ sure}\myhl{156}{227}{247}{ if}\myhl{156}{227}{247}{ that}\myhl{156}{227}{247}{'s}\myhl{156}{227}{247}{ the}\myhl{156}{227}{247}{ case}\myhl{156}{227}{247}{.}\myhl{196}{238}{250}{Overall}\myhl{196}{238}{250}{,}\myhl{196}{238}{250}{ I}\myhl{196}{238}{250}{ think}\myhl{196}{238}{250}{ TV}\myhl{196}{238}{250}{ news}\myhl{196}{238}{250}{ does}\myhl{196}{238}{250}{ tend}\myhl{196}{238}{250}{ to}\myhl{196}{238}{250}{ favor}\myhl{196}{238}{250}{ the}\myhl{196}{238}{250}{ governing}\myhl{196}{238}{250}{ party}\myhl{196}{238}{250}{ more}\myhl{196}{238}{250}{ often}\myhl{196}{238}{250}{ than}\myhl{196}{238}{250}{ not}\myhl{196}{238}{250}{,}\myhl{196}{238}{250}{ especially}\myhl{196}{238}{250}{ during}\myhl{196}{238}{250}{ their}\myhl{196}{238}{250}{ election}\myhl{196}{238}{250}{ years}\myhl{196}{238}{250}{.}\myhl{27}{191}{236}{ They}\myhl{27}{191}{236}{ get}\myhl{27}{191}{236}{ more}\myhl{27}{191}{236}{ coverage}\myhl{27}{191}{236}{,}\myhl{27}{191}{236}{ better}\myhl{27}{191}{236}{ reporting}\myhl{27}{191}{236}{,}\myhl{27}{191}{236}{ and}\myhl{27}{191}{236}{ might}\myhl{27}{191}{236}{ even}\myhl{27}{191}{236}{ shape}\myhl{27}{191}{236}{ public}\myhl{27}{191}{236}{ perception}\myhl{27}{191}{236}{ through}\myhl{27}{191}{236}{ their}\myhl{27}{191}{236}{ narratives}\myhl{27}{191}{236}{.} \dots\myhl{157}{227}{247}{ It}\myhl{157}{227}{247}{ might}\myhl{157}{227}{247}{ not}\myhl{157}{227}{247}{ be}\myhl{157}{227}{247}{ every}\myhl{157}{227}{247}{ election}\myhl{157}{227}{247}{,}\myhl{157}{227}{247}{ but}\myhl{157}{227}{247}{ it}\myhl{157}{227}{247}{'s}\myhl{157}{227}{247}{ a}\myhl{157}{227}{247}{ significant}\myhl{157}{227}{247}{ trend}\myhl{157}{227}{247}{.}
\end{tcolorbox}
\begin{tcolorbox}[colback=white, , title=Average attention]
\textbf{Thoughts}:
\myhl{165}{229}{247}{Okay}\myhl{165}{229}{247}{,}\myhl{165}{229}{247}{ so}\myhl{165}{229}{247}{ I}\myhl{165}{229}{247}{ need}\myhl{165}{229}{247}{ to}\myhl{165}{229}{247}{ figure}\myhl{165}{229}{247}{ out}\myhl{165}{229}{247}{ how}\myhl{165}{229}{247}{ often}\myhl{165}{229}{247}{ TV}\myhl{165}{229}{247}{ news}\myhl{165}{229}{247}{ favors}\myhl{165}{229}{247}{ the}\myhl{165}{229}{247}{ governing}\myhl{165}{229}{247}{ party}\myhl{165}{229}{247}{ in}\myhl{165}{229}{247}{ this}\myhl{165}{229}{247}{ country}\myhl{165}{229}{247}{'s}\myhl{165}{229}{247}{ elections}\myhl{165}{229}{247}{.} \dots\myhl{192}{237}{250}{First}\myhl{192}{237}{250}{,}\myhl{192}{237}{250}{ I}\myhl{192}{237}{250}{ guess}\myhl{192}{237}{250}{ the}\myhl{192}{237}{250}{ media}\myhl{192}{237}{250}{ might}\myhl{192}{237}{250}{ lean}\myhl{192}{237}{250}{ towards}\myhl{192}{237}{250}{ the}\myhl{192}{237}{250}{ current}\myhl{192}{237}{250}{ government}\myhl{192}{237}{250}{ because}\myhl{192}{237}{250}{ they}\myhl{192}{237}{250}{'re}\myhl{192}{237}{250}{ the}\myhl{192}{237}{250}{ ones}\myhl{192}{237}{250}{ in}\myhl{192}{237}{250}{ power}\myhl{192}{237}{250}{.} \dots\myhl{184}{235}{249}{ I}\myhl{184}{235}{249}{ mean}\myhl{184}{235}{249}{,}\myhl{184}{235}{249}{ elections}\myhl{184}{235}{249}{ are}\myhl{184}{235}{249}{ cyc}\myhl{184}{235}{249}{lical}\myhl{184}{235}{249}{,}\myhl{184}{235}{249}{ so}\myhl{184}{235}{249}{ the}\myhl{184}{235}{249}{ ruling}\myhl{184}{235}{249}{ party}\myhl{184}{235}{249}{ would}\myhl{184}{235}{249}{ naturally}\myhl{184}{235}{249}{ have}\myhl{184}{235}{249}{ an}\myhl{184}{235}{249}{ advantage}\myhl{184}{235}{249}{ over}\myhl{184}{235}{249}{ time}\myhl{184}{235}{249}{.} \dots\myhl{177}{233}{248}{I}\myhl{177}{233}{248}{ remember}\myhl{177}{233}{248}{ hearing}\myhl{177}{233}{248}{ that}\myhl{177}{233}{248}{ sometimes}\myhl{177}{233}{248}{ the}\myhl{177}{233}{248}{ opposition}\myhl{177}{233}{248}{ gets}\myhl{177}{233}{248}{ more}\myhl{177}{233}{248}{ coverage}\myhl{177}{233}{248}{,}\myhl{177}{233}{248}{ especially}\myhl{177}{233}{248}{ when}\myhl{177}{233}{248}{ the}\myhl{177}{233}{248}{ ruling}\myhl{177}{233}{248}{ party}\myhl{177}{233}{248}{ is}\myhl{177}{233}{248}{ doing}\myhl{177}{233}{248}{ poorly}\myhl{177}{233}{248}{.}\myhl{191}{237}{249}{ But}\myhl{191}{237}{249}{ during}\myhl{191}{237}{249}{ their}\myhl{191}{237}{249}{ own}\myhl{191}{237}{249}{ party}\myhl{191}{237}{249}{'s}\myhl{191}{237}{249}{ election}\myhl{191}{237}{249}{,}\myhl{191}{237}{249}{ maybe}\myhl{191}{237}{249}{ the}\myhl{191}{237}{249}{ news}\myhl{191}{237}{249}{ outlets}\myhl{191}{237}{249}{ give}\myhl{191}{237}{249}{ them}\myhl{191}{237}{249}{ more}\myhl{191}{237}{249}{ attention}\myhl{191}{237}{249}{.}\myhl{99}{211}{242}{ I}\myhl{99}{211}{242}{'m}\myhl{99}{211}{242}{ not}\myhl{99}{211}{242}{ sure}\myhl{99}{211}{242}{ how}\myhl{99}{211}{242}{ much}\myhl{99}{211}{242}{ the}\myhl{99}{211}{242}{ media}\myhl{99}{211}{242}{ actually}\myhl{99}{211}{242}{ sw}\myhl{99}{211}{242}{ays}\myhl{99}{211}{242}{ public}\myhl{99}{211}{242}{ opinion}\myhl{99}{211}{242}{ or}\myhl{99}{211}{242}{ if}\myhl{99}{211}{242}{ it}\myhl{99}{211}{242}{'s}\myhl{99}{211}{242}{ just}\myhl{99}{211}{242}{ a}\myhl{99}{211}{242}{ perception}\myhl{99}{211}{242}{.}\myhl{186}{235}{249}{ Maybe}\myhl{186}{235}{249}{ there}\myhl{186}{235}{249}{'s}\myhl{186}{235}{249}{ some}\myhl{186}{235}{249}{ bias}\myhl{186}{235}{249}{,}\myhl{186}{235}{249}{ but}\myhl{186}{235}{249}{ is}\myhl{186}{235}{249}{ it}\myhl{186}{235}{249}{ significant}\myhl{186}{235}{249}{?}\myhl{190}{237}{249}{Also}\myhl{190}{237}{249}{,}\myhl{190}{237}{249}{ the}\myhl{190}{237}{249}{ media}\myhl{190}{237}{249}{ might}\myhl{190}{237}{249}{ have}\myhl{190}{237}{249}{ editorial}\myhl{190}{237}{249}{ policies}\myhl{190}{237}{249}{ that}\myhl{190}{237}{249}{ favor}\myhl{190}{237}{249}{ the}\myhl{190}{237}{249}{ government}\myhl{190}{237}{249}{.} \dots\myhl{138}{222}{245}{ I}\myhl{138}{222}{245}{ think}\myhl{138}{222}{245}{ in}\myhl{138}{222}{245}{ many}\myhl{138}{222}{245}{ countries}\myhl{138}{222}{245}{,}\myhl{138}{222}{245}{ media}\myhl{138}{222}{245}{ outlets}\myhl{138}{222}{245}{ have}\myhl{138}{222}{245}{ clear}\myhl{138}{222}{245}{ st}\myhl{138}{222}{245}{ances}\myhl{138}{222}{245}{ on}\myhl{138}{222}{245}{ political}\myhl{138}{222}{245}{ issues}\myhl{138}{222}{245}{,}\myhl{138}{222}{245}{ so}\myhl{138}{222}{245}{ it}\myhl{138}{222}{245}{'s}\myhl{138}{222}{245}{ possible}\myhl{138}{222}{245}{ they}\myhl{138}{222}{245}{ lean}\myhl{138}{222}{245}{ towards}\myhl{138}{222}{245}{ whichever}\myhl{138}{222}{245}{ party}\myhl{138}{222}{245}{ is}\myhl{138}{222}{245}{ in}\myhl{138}{222}{245}{ power}\myhl{138}{222}{245}{.} \dots\myhl{192}{237}{250}{ Does}\myhl{192}{237}{250}{ it}\myhl{192}{237}{250}{ happen}\myhl{192}{237}{250}{ every}\myhl{192}{237}{250}{ election}\myhl{192}{237}{250}{,}\myhl{192}{237}{250}{ or}\myhl{192}{237}{250}{ only}\myhl{192}{237}{250}{ in}\myhl{192}{237}{250}{ certain}\myhl{192}{237}{250}{ circumstances}\myhl{192}{237}{250}{?}\myhl{152}{226}{246}{I}\myhl{152}{226}{246}{ should}\myhl{152}{226}{246}{ also}\myhl{152}{226}{246}{ consider}\myhl{152}{226}{246}{ the}\myhl{152}{226}{246}{ role}\myhl{152}{226}{246}{ of}\myhl{152}{226}{246}{ campaign}\myhl{152}{226}{246}{ finance}\myhl{152}{226}{246}{ and}\myhl{152}{226}{246}{ advertising}\myhl{152}{226}{246}{.}\myhl{171}{231}{248}{ If}\myhl{171}{231}{248}{ the}\myhl{171}{231}{248}{ governing}\myhl{171}{231}{248}{ party}\myhl{171}{231}{248}{ has}\myhl{171}{231}{248}{ more}\myhl{171}{231}{248}{ funding}\myhl{171}{231}{248}{,}\myhl{171}{231}{248}{ they}\myhl{171}{231}{248}{ might}\myhl{171}{231}{248}{ get}\myhl{171}{231}{248}{ more}\myhl{171}{231}{248}{ air}\myhl{171}{231}{248}{time}\myhl{171}{231}{248}{.}\myhl{145}{224}{246}{ But}\myhl{145}{224}{246}{ in}\myhl{145}{224}{246}{ countries}\myhl{145}{224}{246}{ with}\myhl{145}{224}{246}{ free}\myhl{145}{224}{246}{ media}\myhl{145}{224}{246}{,}\myhl{145}{224}{246}{ maybe}\myhl{145}{224}{246}{ the}\myhl{145}{224}{246}{ coverage}\myhl{145}{224}{246}{ is}\myhl{145}{224}{246}{ more}\myhl{145}{224}{246}{ balanced}\myhl{145}{224}{246}{.}\myhl{124}{218}{244}{ However}\myhl{124}{218}{244}{,}\myhl{124}{218}{244}{ in}\myhl{124}{218}{244}{ some}\myhl{124}{218}{244}{ cases}\myhl{124}{218}{244}{,}\myhl{124}{218}{244}{ especially}\myhl{124}{218}{244}{ in}\myhl{124}{218}{244}{ less}\myhl{124}{218}{244}{ competitive}\myhl{124}{218}{244}{ elections}\myhl{124}{218}{244}{,}\myhl{124}{218}{244}{ the}\myhl{124}{218}{244}{ media}\myhl{124}{218}{244}{ might}\myhl{124}{218}{244}{ focus}\myhl{124}{218}{244}{ more}\myhl{124}{218}{244}{ on}\myhl{124}{218}{244}{ the}\myhl{124}{218}{244}{ ruling}\myhl{124}{218}{244}{ party}\myhl{124}{218}{244}{.}\myhl{163}{229}{247}{Another}\myhl{163}{229}{247}{ angle}\myhl{163}{229}{247}{ is}\myhl{163}{229}{247}{ the}\myhl{163}{229}{247}{ impact}\myhl{163}{229}{247}{ of}\myhl{163}{229}{247}{ recent}\myhl{163}{229}{247}{ events}\myhl{163}{229}{247}{.}\myhl{84}{207}{241}{ If}\myhl{84}{207}{241}{ the}\myhl{84}{207}{241}{ ruling}\myhl{84}{207}{241}{ party}\myhl{84}{207}{241}{ is}\myhl{84}{207}{241}{ facing}\myhl{84}{207}{241}{ scandals}\myhl{84}{207}{241}{ or}\myhl{84}{207}{241}{ crises}\myhl{84}{207}{241}{,}\myhl{84}{207}{241}{ the}\myhl{84}{207}{241}{ media}\myhl{84}{207}{241}{ might}\myhl{84}{207}{241}{ down}\myhl{84}{207}{241}{play}\myhl{84}{207}{241}{ opposition}\myhl{84}{207}{241}{ issues}\myhl{84}{207}{241}{ more}\myhl{84}{207}{241}{,}\myhl{84}{207}{241}{ giving}\myhl{84}{207}{241}{ the}\myhl{84}{207}{241}{ party}\myhl{84}{207}{241}{ more}\myhl{84}{207}{241}{ coverage}\myhl{84}{207}{241}{.}\myhl{194}{237}{250}{ Conversely}\myhl{194}{237}{250}{,}\myhl{194}{237}{250}{ when}\myhl{194}{237}{250}{ the}\myhl{194}{237}{250}{ opposition}\myhl{194}{237}{250}{ is}\myhl{194}{237}{250}{ strong}\myhl{194}{237}{250}{,}\myhl{194}{237}{250}{ they}\myhl{194}{237}{250}{ might}\myhl{194}{237}{250}{ get}\myhl{194}{237}{250}{ more}\myhl{194}{237}{250}{ attention}\myhl{194}{237}{250}{.}\myhl{82}{206}{241}{ So}\myhl{82}{206}{241}{ the}\myhl{82}{206}{241}{ frequency}\myhl{82}{206}{241}{ could}\myhl{82}{206}{241}{ vary}\myhl{82}{206}{241}{ based}\myhl{82}{206}{241}{ on}\myhl{82}{206}{241}{ the}\myhl{82}{206}{241}{ political}\myhl{82}{206}{241}{ climate}\myhl{82}{206}{241}{.}\myhl{146}{224}{246}{I}\myhl{146}{224}{246}{'m}\myhl{146}{224}{246}{ also}\myhl{146}{224}{246}{ thinking}\myhl{146}{224}{246}{ about}\myhl{146}{224}{246}{ the}\myhl{146}{224}{246}{ electoral}\myhl{146}{224}{246}{ system}\myhl{146}{224}{246}{.}\myhl{100}{211}{242}{ In}\myhl{100}{211}{242}{ some}\myhl{100}{211}{242}{ systems}\myhl{100}{211}{242}{,}\myhl{100}{211}{242}{ like}\myhl{100}{211}{242}{ first}\myhl{100}{211}{242}{-p}\myhl{100}{211}{242}{ast}\myhl{100}{211}{242}{-the}\myhl{100}{211}{242}{-post}\myhl{100}{211}{242}{,}\myhl{100}{211}{242}{ the}\myhl{100}{211}{242}{ ruling}\myhl{100}{211}{242}{ party}\myhl{100}{211}{242}{ tends}\myhl{100}{211}{242}{ to}\myhl{100}{211}{242}{ win}\myhl{100}{211}{242}{ more}\myhl{100}{211}{242}{ seats}\myhl{100}{211}{242}{,}\myhl{100}{211}{242}{ which}\myhl{100}{211}{242}{ might}\myhl{100}{211}{242}{ lead}\myhl{100}{211}{242}{ to}\myhl{100}{211}{242}{ more}\myhl{100}{211}{242}{ media}\myhl{100}{211}{242}{ coverage}\myhl{100}{211}{242}{.}\myhl{88}{208}{241}{ In}\myhl{88}{208}{241}{ other}\myhl{88}{208}{241}{ systems}\myhl{88}{208}{241}{,}\myhl{88}{208}{241}{ like}\myhl{88}{208}{241}{ proportional}\myhl{88}{208}{241}{ representation}\myhl{88}{208}{241}{,}\myhl{88}{208}{241}{ the}\myhl{88}{208}{241}{ media}\myhl{88}{208}{241}{ might}\myhl{88}{208}{241}{ cover}\myhl{88}{208}{241}{ all}\myhl{88}{208}{241}{ parties}\myhl{88}{208}{241}{ more}\myhl{88}{208}{241}{ equally}\myhl{88}{208}{241}{,}\myhl{88}{208}{241}{ but}\myhl{88}{208}{241}{ I}\myhl{88}{208}{241}{'m}\myhl{88}{208}{241}{ not}\myhl{88}{208}{241}{ sure}\myhl{88}{208}{241}{ if}\myhl{88}{208}{241}{ that}\myhl{88}{208}{241}{'s}\myhl{88}{208}{241}{ the}\myhl{88}{208}{241}{ case}\myhl{88}{208}{241}{.}\myhl{40}{194}{237}{Overall}\myhl{40}{194}{237}{,}\myhl{40}{194}{237}{ I}\myhl{40}{194}{237}{ think}\myhl{40}{194}{237}{ TV}\myhl{40}{194}{237}{ news}\myhl{40}{194}{237}{ does}\myhl{40}{194}{237}{ tend}\myhl{40}{194}{237}{ to}\myhl{40}{194}{237}{ favor}\myhl{40}{194}{237}{ the}\myhl{40}{194}{237}{ governing}\myhl{40}{194}{237}{ party}\myhl{40}{194}{237}{ more}\myhl{40}{194}{237}{ often}\myhl{40}{194}{237}{ than}\myhl{40}{194}{237}{ not}\myhl{40}{194}{237}{,}\myhl{40}{194}{237}{ especially}\myhl{40}{194}{237}{ during}\myhl{40}{194}{237}{ their}\myhl{40}{194}{237}{ election}\myhl{40}{194}{237}{ years}\myhl{40}{194}{237}{.}\myhl{0}{183}{234}{ They}\myhl{0}{183}{234}{ get}\myhl{0}{183}{234}{ more}\myhl{0}{183}{234}{ coverage}\myhl{0}{183}{234}{,}\myhl{0}{183}{234}{ better}\myhl{0}{183}{234}{ reporting}\myhl{0}{183}{234}{,}\myhl{0}{183}{234}{ and}\myhl{0}{183}{234}{ might}\myhl{0}{183}{234}{ even}\myhl{0}{183}{234}{ shape}\myhl{0}{183}{234}{ public}\myhl{0}{183}{234}{ perception}\myhl{0}{183}{234}{ through}\myhl{0}{183}{234}{ their}\myhl{0}{183}{234}{ narratives}\myhl{0}{183}{234}{.}\myhl{18}{188}{236}{ But}\myhl{18}{188}{236}{ the}\myhl{18}{188}{236}{ exact}\myhl{18}{188}{236}{ frequency}\myhl{18}{188}{236}{ isn}\myhl{18}{188}{236}{'t}\myhl{18}{188}{236}{ clear}\myhl{18}{188}{236}{ to}\myhl{18}{188}{236}{ me}\myhl{18}{188}{236}{.}\myhl{103}{212}{242}{ It}\myhl{103}{212}{242}{ might}\myhl{103}{212}{242}{ not}\myhl{103}{212}{242}{ be}\myhl{103}{212}{242}{ every}\myhl{103}{212}{242}{ election}\myhl{103}{212}{242}{,}\myhl{103}{212}{242}{ but}\myhl{103}{212}{242}{ it}\myhl{103}{212}{242}{'s}\myhl{103}{212}{242}{ a}\myhl{103}{212}{242}{ significant}\myhl{103}{212}{242}{ trend}\myhl{103}{212}{242}{.}
\end{tcolorbox}
\begin{tcolorbox}[colback=white, , title=ESM (256 samples)]
\textbf{Thoughts}:
 \dots\myhl{152}{226}{246}{ I}\myhl{152}{226}{246}{'m}\myhl{152}{226}{246}{ not}\myhl{152}{226}{246}{ sure}\myhl{152}{226}{246}{ how}\myhl{152}{226}{246}{ much}\myhl{152}{226}{246}{ the}\myhl{152}{226}{246}{ media}\myhl{152}{226}{246}{ actually}\myhl{152}{226}{246}{ sw}\myhl{152}{226}{246}{ays}\myhl{152}{226}{246}{ public}\myhl{152}{226}{246}{ opinion}\myhl{152}{226}{246}{ or}\myhl{152}{226}{246}{ if}\myhl{152}{226}{246}{ it}\myhl{152}{226}{246}{'s}\myhl{152}{226}{246}{ just}\myhl{152}{226}{246}{ a}\myhl{152}{226}{246}{ perception}\myhl{152}{226}{246}{.} \dots\myhl{167}{230}{248}{I}\myhl{167}{230}{248}{ should}\myhl{167}{230}{248}{ also}\myhl{167}{230}{248}{ consider}\myhl{167}{230}{248}{ the}\myhl{167}{230}{248}{ role}\myhl{167}{230}{248}{ of}\myhl{167}{230}{248}{ campaign}\myhl{167}{230}{248}{ finance}\myhl{167}{230}{248}{ and}\myhl{167}{230}{248}{ advertising}\myhl{167}{230}{248}{.}\myhl{0}{183}{234}{ If}\myhl{0}{183}{234}{ the}\myhl{0}{183}{234}{ governing}\myhl{0}{183}{234}{ party}\myhl{0}{183}{234}{ has}\myhl{0}{183}{234}{ more}\myhl{0}{183}{234}{ funding}\myhl{0}{183}{234}{,}\myhl{0}{183}{234}{ they}\myhl{0}{183}{234}{ might}\myhl{0}{183}{234}{ get}\myhl{0}{183}{234}{ more}\myhl{0}{183}{234}{ air}\myhl{0}{183}{234}{time}\myhl{0}{183}{234}{.} \dots\myhl{175}{232}{248}{Another}\myhl{175}{232}{248}{ angle}\myhl{175}{232}{248}{ is}\myhl{175}{232}{248}{ the}\myhl{175}{232}{248}{ impact}\myhl{175}{232}{248}{ of}\myhl{175}{232}{248}{ recent}\myhl{175}{232}{248}{ events}\myhl{175}{232}{248}{.}\myhl{93}{209}{242}{ If}\myhl{93}{209}{242}{ the}\myhl{93}{209}{242}{ ruling}\myhl{93}{209}{242}{ party}\myhl{93}{209}{242}{ is}\myhl{93}{209}{242}{ facing}\myhl{93}{209}{242}{ scandals}\myhl{93}{209}{242}{ or}\myhl{93}{209}{242}{ crises}\myhl{93}{209}{242}{,}\myhl{93}{209}{242}{ the}\myhl{93}{209}{242}{ media}\myhl{93}{209}{242}{ might}\myhl{93}{209}{242}{ down}\myhl{93}{209}{242}{play}\myhl{93}{209}{242}{ opposition}\myhl{93}{209}{242}{ issues}\myhl{93}{209}{242}{ more}\myhl{93}{209}{242}{,}\myhl{93}{209}{242}{ giving}\myhl{93}{209}{242}{ the}\myhl{93}{209}{242}{ party}\myhl{93}{209}{242}{ more}\myhl{93}{209}{242}{ coverage}\myhl{93}{209}{242}{.} \dots\myhl{171}{231}{248}{I}\myhl{171}{231}{248}{'m}\myhl{171}{231}{248}{ also}\myhl{171}{231}{248}{ thinking}\myhl{171}{231}{248}{ about}\myhl{171}{231}{248}{ the}\myhl{171}{231}{248}{ electoral}\myhl{171}{231}{248}{ system}\myhl{171}{231}{248}{.}\myhl{168}{230}{248}{ In}\myhl{168}{230}{248}{ some}\myhl{168}{230}{248}{ systems}\myhl{168}{230}{248}{,}\myhl{168}{230}{248}{ like}\myhl{168}{230}{248}{ first}\myhl{168}{230}{248}{-p}\myhl{168}{230}{248}{ast}\myhl{168}{230}{248}{-the}\myhl{168}{230}{248}{-post}\myhl{168}{230}{248}{,}\myhl{168}{230}{248}{ the}\myhl{168}{230}{248}{ ruling}\myhl{168}{230}{248}{ party}\myhl{168}{230}{248}{ tends}\myhl{168}{230}{248}{ to}\myhl{168}{230}{248}{ win}\myhl{168}{230}{248}{ more}\myhl{168}{230}{248}{ seats}\myhl{168}{230}{248}{,}\myhl{168}{230}{248}{ which}\myhl{168}{230}{248}{ might}\myhl{168}{230}{248}{ lead}\myhl{168}{230}{248}{ to}\myhl{168}{230}{248}{ more}\myhl{168}{230}{248}{ media}\myhl{168}{230}{248}{ coverage}\myhl{168}{230}{248}{.}\myhl{182}{234}{249}{ In}\myhl{182}{234}{249}{ other}\myhl{182}{234}{249}{ systems}\myhl{182}{234}{249}{,}\myhl{182}{234}{249}{ like}\myhl{182}{234}{249}{ proportional}\myhl{182}{234}{249}{ representation}\myhl{182}{234}{249}{,}\myhl{182}{234}{249}{ the}\myhl{182}{234}{249}{ media}\myhl{182}{234}{249}{ might}\myhl{182}{234}{249}{ cover}\myhl{182}{234}{249}{ all}\myhl{182}{234}{249}{ parties}\myhl{182}{234}{249}{ more}\myhl{182}{234}{249}{ equally}\myhl{182}{234}{249}{,}\myhl{182}{234}{249}{ but}\myhl{182}{234}{249}{ I}\myhl{182}{234}{249}{'m}\myhl{182}{234}{249}{ not}\myhl{182}{234}{249}{ sure}\myhl{182}{234}{249}{ if}\myhl{182}{234}{249}{ that}\myhl{182}{234}{249}{'s}\myhl{182}{234}{249}{ the}\myhl{182}{234}{249}{ case}\myhl{182}{234}{249}{.} \dots\myhl{188}{236}{249}{ They}\myhl{188}{236}{249}{ get}\myhl{188}{236}{249}{ more}\myhl{188}{236}{249}{ coverage}\myhl{188}{236}{249}{,}\myhl{188}{236}{249}{ better}\myhl{188}{236}{249}{ reporting}\myhl{188}{236}{249}{,}\myhl{188}{236}{249}{ and}\myhl{188}{236}{249}{ might}\myhl{188}{236}{249}{ even}\myhl{188}{236}{249}{ shape}\myhl{188}{236}{249}{ public}\myhl{188}{236}{249}{ perception}\myhl{188}{236}{249}{ through}\myhl{188}{236}{249}{ their}\myhl{188}{236}{249}{ narratives}\myhl{188}{236}{249}{.} \dots
\end{tcolorbox}
\textbf{Response} (from \texttt{DeepSeek-R1-Qwen-7B}):
\newline
In this country, TV news often favors the governing party, particularly during their election years. This tendency is evident as the media tends to receive more coverage, better reporting, and a positive narrative that aligns with the party's image. The governing party's influence is amplified through extensive election coverage, favorable reporting on their policies, and sometimes editorial stances that reflect their administration's interests. While the exact frequency isn't uniform, the media's focus on the ruling party is a significant trend, often more pronounced during their election campaigns. \myhl{249}{237}{91}{This bias can shape public perception and media-driven opinion, though the extent can vary based on political events, funding, and electoral systems.}
\end{tcolorbox}
\relsize{1}
\caption{Visualized attributions for a random example from Global Opinions QA for \texttt{DeepSeek-R1-Qwen-7B} with sentences as sources.}
\label{fig:global_opinions_r1_qwen_7b_dtype=bfloat16_sentence_0}
\end{figure*}

\begin{figure*}[t]
\centering\definecolor{forestgreen}{RGB}{34, 139, 34}
\relsize{-1}
\begin{tcolorbox}[colback=white]
\textbf{Query}:
In general, do you think the Israeli government favors or opposes democracy in the Middle East? 
\begin{tcolorbox}[colback=white, colframe=forestgreen, title=\ours (general)]
\textbf{Thoughts}:
 \dots\myhl{202}{240}{250}{ This}\myhl{202}{240}{250}{ makes}\myhl{202}{240}{250}{ me}\myhl{202}{240}{250}{ think}\myhl{202}{240}{250}{ that}\myhl{202}{240}{250}{ while}\myhl{202}{240}{250}{ Israel}\myhl{202}{240}{250}{ maintains}\myhl{202}{240}{250}{ its}\myhl{202}{240}{250}{ own}\myhl{202}{240}{250}{ democratic}\myhl{202}{240}{250}{ system}\myhl{202}{240}{250}{,}\myhl{202}{240}{250}{ their}\myhl{202}{240}{250}{ foreign}\myhl{202}{240}{250}{ policy}\myhl{202}{240}{250}{ and}\myhl{202}{240}{250}{ stance}\myhl{202}{240}{250}{ on}\myhl{202}{240}{250}{ issues}\myhl{202}{240}{250}{ like}\myhl{202}{240}{250}{ human}\myhl{202}{240}{250}{ rights}\myhl{202}{240}{250}{ and}\myhl{202}{240}{250}{ territorial}\myhl{202}{240}{250}{ sovereignty}\myhl{202}{240}{250}{ might}\myhl{202}{240}{250}{ not}\myhl{202}{240}{250}{ align}\myhl{202}{240}{250}{ with}\myhl{202}{240}{250}{ the}\myhl{202}{240}{250}{ democratic}\myhl{202}{240}{250}{ values}\myhl{202}{240}{250}{ of}\myhl{202}{240}{250}{ other}\myhl{202}{240}{250}{ Middle}\myhl{202}{240}{250}{ Eastern}\myhl{202}{240}{250}{ countries}\myhl{202}{240}{250}{.}\myhl{147}{225}{246}{Moreover}\myhl{147}{225}{246}{,}\myhl{147}{225}{246}{ I}\myhl{147}{225}{246}{ think}\myhl{147}{225}{246}{ about}\myhl{147}{225}{246}{ the}\myhl{147}{225}{246}{ concept}\myhl{147}{225}{246}{ of}\myhl{147}{225}{246}{ a}\myhl{147}{225}{246}{ two}\myhl{147}{225}{246}{-state}\myhl{147}{225}{246}{ solution}\myhl{147}{225}{246}{.}\myhl{0}{183}{234}{ Many}\myhl{0}{183}{234}{ Middle}\myhl{0}{183}{234}{ Eastern}\myhl{0}{183}{234}{ democr}\myhl{0}{183}{234}{acies}\myhl{0}{183}{234}{ believe}\myhl{0}{183}{234}{ that}\myhl{0}{183}{234}{ Israel}\myhl{0}{183}{234}{ should}\myhl{0}{183}{234}{ maintain}\myhl{0}{183}{234}{ its}\myhl{0}{183}{234}{ current}\myhl{0}{183}{234}{ borders}\myhl{0}{183}{234}{ to}\myhl{0}{183}{234}{ allow}\myhl{0}{183}{234}{ for}\myhl{0}{183}{234}{ a}\myhl{0}{183}{234}{ peaceful}\myhl{0}{183}{234}{ co}\myhl{0}{183}{234}{existence}\myhl{0}{183}{234}{ with}\myhl{0}{183}{234}{ the}\myhl{0}{183}{234}{ Palestinians}\myhl{0}{183}{234}{.}\myhl{183}{234}{249}{ This}\myhl{183}{234}{249}{ perspective}\myhl{183}{234}{249}{ might}\myhl{183}{234}{249}{ be}\myhl{183}{234}{249}{ more}\myhl{183}{234}{249}{ in}\myhl{183}{234}{249}{ line}\myhl{183}{234}{249}{ with}\myhl{183}{234}{249}{ the}\myhl{183}{234}{249}{ democratic}\myhl{183}{234}{249}{ values}\myhl{183}{234}{249}{ of}\myhl{183}{234}{249}{ those}\myhl{183}{234}{249}{ countries}\myhl{183}{234}{249}{,}\myhl{183}{234}{249}{ as}\myhl{183}{234}{249}{ opposed}\myhl{183}{234}{249}{ to}\myhl{183}{234}{249}{ Israel}\myhl{183}{234}{249}{'s}\myhl{183}{234}{249}{ own}\myhl{183}{234}{249}{ approach}\myhl{183}{234}{249}{,}\myhl{183}{234}{249}{ which}\myhl{183}{234}{249}{ might}\myhl{183}{234}{249}{ prioritize}\myhl{183}{234}{249}{ security}\myhl{183}{234}{249}{ and}\myhl{183}{234}{249}{ territorial}\myhl{183}{234}{249}{ integrity}\myhl{183}{234}{249}{ over}\myhl{183}{234}{249}{ a}\myhl{183}{234}{249}{ multi}\myhl{183}{234}{249}{-state}\myhl{183}{234}{249}{ solution}\myhl{183}{234}{249}{.}\myhl{202}{240}{250}{So}\myhl{202}{240}{250}{,}\myhl{202}{240}{250}{ putting}\myhl{202}{240}{250}{ this}\myhl{202}{240}{250}{ together}\myhl{202}{240}{250}{,}\myhl{202}{240}{250}{ it}\myhl{202}{240}{250}{ seems}\myhl{202}{240}{250}{ like}\myhl{202}{240}{250}{ the}\myhl{202}{240}{250}{ Israeli}\myhl{202}{240}{250}{ government}\myhl{202}{240}{250}{ supports}\myhl{202}{240}{250}{ its}\myhl{202}{240}{250}{ own}\myhl{202}{240}{250}{ democracy}\myhl{202}{240}{250}{ but}\myhl{202}{240}{250}{ may}\myhl{202}{240}{250}{ oppose}\myhl{202}{240}{250}{ or}\myhl{202}{240}{250}{ be}\myhl{202}{240}{250}{ neutral}\myhl{202}{240}{250}{ towards}\myhl{202}{240}{250}{ the}\myhl{202}{240}{250}{ democratic}\myhl{202}{240}{250}{ approaches}\myhl{202}{240}{250}{ of}\myhl{202}{240}{250}{ other}\myhl{202}{240}{250}{ Middle}\myhl{202}{240}{250}{ Eastern}\myhl{202}{240}{250}{ countries}\myhl{202}{240}{250}{,}\myhl{202}{240}{250}{ especially}\myhl{202}{240}{250}{ regarding}\myhl{202}{240}{250}{ issues}\myhl{202}{240}{250}{ like}\myhl{202}{240}{250}{ Palestinian}\myhl{202}{240}{250}{ rights}\myhl{202}{240}{250}{ and}\myhl{202}{240}{250}{ the}\myhl{202}{240}{250}{ structure}\myhl{202}{240}{250}{ of}\myhl{202}{240}{250}{ governance}\myhl{202}{240}{250}{ in}\myhl{202}{240}{250}{ the}\myhl{202}{240}{250}{ region}\myhl{202}{240}{250}{.} \dots
\end{tcolorbox}
\begin{tcolorbox}[colback=white, , title=Average attention]
\textbf{Thoughts}:
 \dots\myhl{175}{232}{248}{ I}\myhl{175}{232}{248}{'ve}\myhl{175}{232}{248}{ heard}\myhl{175}{232}{248}{ that}\myhl{175}{232}{248}{ some}\myhl{175}{232}{248}{ Middle}\myhl{175}{232}{248}{ Eastern}\myhl{175}{232}{248}{ countries}\myhl{175}{232}{248}{ and}\myhl{175}{232}{248}{ international}\myhl{175}{232}{248}{ organizations}\myhl{175}{232}{248}{ criticize}\myhl{175}{232}{248}{ Israel}\myhl{175}{232}{248}{ for}\myhl{175}{232}{248}{ not}\myhl{175}{232}{248}{ fully}\myhl{175}{232}{248}{ respecting}\myhl{175}{232}{248}{ democratic}\myhl{175}{232}{248}{ principles}\myhl{175}{232}{248}{,}\myhl{175}{232}{248}{ especially}\myhl{175}{232}{248}{ regarding}\myhl{175}{232}{248}{ the}\myhl{175}{232}{248}{ rights}\myhl{175}{232}{248}{ of}\myhl{175}{232}{248}{ Palestinian}\myhl{175}{232}{248}{ people}\myhl{175}{232}{248}{.}\myhl{175}{232}{248}{I}\myhl{175}{232}{248}{ also}\myhl{175}{232}{248}{ recall}\myhl{175}{232}{248}{ that}\myhl{175}{232}{248}{ the}\myhl{175}{232}{248}{ Israeli}\myhl{175}{232}{248}{ government}\myhl{175}{232}{248}{ has}\myhl{175}{232}{248}{ been}\myhl{175}{232}{248}{ involved}\myhl{175}{232}{248}{ in}\myhl{175}{232}{248}{ peace}\myhl{175}{232}{248}{ negotiations}\myhl{175}{232}{248}{ with}\myhl{175}{232}{248}{ the}\myhl{175}{232}{248}{ Palestinians}\myhl{175}{232}{248}{,}\myhl{175}{232}{248}{ but}\myhl{175}{232}{248}{ there}\myhl{175}{232}{248}{'s}\myhl{175}{232}{248}{ a}\myhl{175}{232}{248}{ lot}\myhl{175}{232}{248}{ of}\myhl{175}{232}{248}{ opposition}\myhl{175}{232}{248}{ from}\myhl{175}{232}{248}{ both}\myhl{175}{232}{248}{ sides}\myhl{175}{232}{248}{.}\myhl{128}{219}{244}{ This}\myhl{128}{219}{244}{ makes}\myhl{128}{219}{244}{ me}\myhl{128}{219}{244}{ think}\myhl{128}{219}{244}{ that}\myhl{128}{219}{244}{ while}\myhl{128}{219}{244}{ Israel}\myhl{128}{219}{244}{ maintains}\myhl{128}{219}{244}{ its}\myhl{128}{219}{244}{ own}\myhl{128}{219}{244}{ democratic}\myhl{128}{219}{244}{ system}\myhl{128}{219}{244}{,}\myhl{128}{219}{244}{ their}\myhl{128}{219}{244}{ foreign}\myhl{128}{219}{244}{ policy}\myhl{128}{219}{244}{ and}\myhl{128}{219}{244}{ stance}\myhl{128}{219}{244}{ on}\myhl{128}{219}{244}{ issues}\myhl{128}{219}{244}{ like}\myhl{128}{219}{244}{ human}\myhl{128}{219}{244}{ rights}\myhl{128}{219}{244}{ and}\myhl{128}{219}{244}{ territorial}\myhl{128}{219}{244}{ sovereignty}\myhl{128}{219}{244}{ might}\myhl{128}{219}{244}{ not}\myhl{128}{219}{244}{ align}\myhl{128}{219}{244}{ with}\myhl{128}{219}{244}{ the}\myhl{128}{219}{244}{ democratic}\myhl{128}{219}{244}{ values}\myhl{128}{219}{244}{ of}\myhl{128}{219}{244}{ other}\myhl{128}{219}{244}{ Middle}\myhl{128}{219}{244}{ Eastern}\myhl{128}{219}{244}{ countries}\myhl{128}{219}{244}{.}\myhl{171}{231}{248}{Moreover}\myhl{171}{231}{248}{,}\myhl{171}{231}{248}{ I}\myhl{171}{231}{248}{ think}\myhl{171}{231}{248}{ about}\myhl{171}{231}{248}{ the}\myhl{171}{231}{248}{ concept}\myhl{171}{231}{248}{ of}\myhl{171}{231}{248}{ a}\myhl{171}{231}{248}{ two}\myhl{171}{231}{248}{-state}\myhl{171}{231}{248}{ solution}\myhl{171}{231}{248}{.}\myhl{0}{183}{234}{ Many}\myhl{0}{183}{234}{ Middle}\myhl{0}{183}{234}{ Eastern}\myhl{0}{183}{234}{ democr}\myhl{0}{183}{234}{acies}\myhl{0}{183}{234}{ believe}\myhl{0}{183}{234}{ that}\myhl{0}{183}{234}{ Israel}\myhl{0}{183}{234}{ should}\myhl{0}{183}{234}{ maintain}\myhl{0}{183}{234}{ its}\myhl{0}{183}{234}{ current}\myhl{0}{183}{234}{ borders}\myhl{0}{183}{234}{ to}\myhl{0}{183}{234}{ allow}\myhl{0}{183}{234}{ for}\myhl{0}{183}{234}{ a}\myhl{0}{183}{234}{ peaceful}\myhl{0}{183}{234}{ co}\myhl{0}{183}{234}{existence}\myhl{0}{183}{234}{ with}\myhl{0}{183}{234}{ the}\myhl{0}{183}{234}{ Palestinians}\myhl{0}{183}{234}{.}\myhl{59}{200}{239}{ This}\myhl{59}{200}{239}{ perspective}\myhl{59}{200}{239}{ might}\myhl{59}{200}{239}{ be}\myhl{59}{200}{239}{ more}\myhl{59}{200}{239}{ in}\myhl{59}{200}{239}{ line}\myhl{59}{200}{239}{ with}\myhl{59}{200}{239}{ the}\myhl{59}{200}{239}{ democratic}\myhl{59}{200}{239}{ values}\myhl{59}{200}{239}{ of}\myhl{59}{200}{239}{ those}\myhl{59}{200}{239}{ countries}\myhl{59}{200}{239}{,}\myhl{59}{200}{239}{ as}\myhl{59}{200}{239}{ opposed}\myhl{59}{200}{239}{ to}\myhl{59}{200}{239}{ Israel}\myhl{59}{200}{239}{'s}\myhl{59}{200}{239}{ own}\myhl{59}{200}{239}{ approach}\myhl{59}{200}{239}{,}\myhl{59}{200}{239}{ which}\myhl{59}{200}{239}{ might}\myhl{59}{200}{239}{ prioritize}\myhl{59}{200}{239}{ security}\myhl{59}{200}{239}{ and}\myhl{59}{200}{239}{ territorial}\myhl{59}{200}{239}{ integrity}\myhl{59}{200}{239}{ over}\myhl{59}{200}{239}{ a}\myhl{59}{200}{239}{ multi}\myhl{59}{200}{239}{-state}\myhl{59}{200}{239}{ solution}\myhl{59}{200}{239}{.}\myhl{36}{193}{237}{So}\myhl{36}{193}{237}{,}\myhl{36}{193}{237}{ putting}\myhl{36}{193}{237}{ this}\myhl{36}{193}{237}{ together}\myhl{36}{193}{237}{,}\myhl{36}{193}{237}{ it}\myhl{36}{193}{237}{ seems}\myhl{36}{193}{237}{ like}\myhl{36}{193}{237}{ the}\myhl{36}{193}{237}{ Israeli}\myhl{36}{193}{237}{ government}\myhl{36}{193}{237}{ supports}\myhl{36}{193}{237}{ its}\myhl{36}{193}{237}{ own}\myhl{36}{193}{237}{ democracy}\myhl{36}{193}{237}{ but}\myhl{36}{193}{237}{ may}\myhl{36}{193}{237}{ oppose}\myhl{36}{193}{237}{ or}\myhl{36}{193}{237}{ be}\myhl{36}{193}{237}{ neutral}\myhl{36}{193}{237}{ towards}\myhl{36}{193}{237}{ the}\myhl{36}{193}{237}{ democratic}\myhl{36}{193}{237}{ approaches}\myhl{36}{193}{237}{ of}\myhl{36}{193}{237}{ other}\myhl{36}{193}{237}{ Middle}\myhl{36}{193}{237}{ Eastern}\myhl{36}{193}{237}{ countries}\myhl{36}{193}{237}{,}\myhl{36}{193}{237}{ especially}\myhl{36}{193}{237}{ regarding}\myhl{36}{193}{237}{ issues}\myhl{36}{193}{237}{ like}\myhl{36}{193}{237}{ Palestinian}\myhl{36}{193}{237}{ rights}\myhl{36}{193}{237}{ and}\myhl{36}{193}{237}{ the}\myhl{36}{193}{237}{ structure}\myhl{36}{193}{237}{ of}\myhl{36}{193}{237}{ governance}\myhl{36}{193}{237}{ in}\myhl{36}{193}{237}{ the}\myhl{36}{193}{237}{ region}\myhl{36}{193}{237}{.}\myhl{165}{229}{247}{ They}\myhl{165}{229}{247}{ might}\myhl{165}{229}{247}{ prioritize}\myhl{165}{229}{247}{ their}\myhl{165}{229}{247}{ national}\myhl{165}{229}{247}{ interests}\myhl{165}{229}{247}{ and}\myhl{165}{229}{247}{ security}\myhl{165}{229}{247}{ over}\myhl{165}{229}{247}{ the}\myhl{165}{229}{247}{ democratic}\myhl{165}{229}{247}{ principles}\myhl{165}{229}{247}{ of}\myhl{165}{229}{247}{ other}\myhl{165}{229}{247}{ nations}\myhl{165}{229}{247}{.}
\end{tcolorbox}
\begin{tcolorbox}[colback=white, , title=ESM (256 samples)]
\textbf{Thoughts}:
 \dots\myhl{201}{240}{250}{Moreover}\myhl{201}{240}{250}{,}\myhl{201}{240}{250}{ I}\myhl{201}{240}{250}{ think}\myhl{201}{240}{250}{ about}\myhl{201}{240}{250}{ the}\myhl{201}{240}{250}{ concept}\myhl{201}{240}{250}{ of}\myhl{201}{240}{250}{ a}\myhl{201}{240}{250}{ two}\myhl{201}{240}{250}{-state}\myhl{201}{240}{250}{ solution}\myhl{201}{240}{250}{.}\myhl{0}{183}{234}{ Many}\myhl{0}{183}{234}{ Middle}\myhl{0}{183}{234}{ Eastern}\myhl{0}{183}{234}{ democr}\myhl{0}{183}{234}{acies}\myhl{0}{183}{234}{ believe}\myhl{0}{183}{234}{ that}\myhl{0}{183}{234}{ Israel}\myhl{0}{183}{234}{ should}\myhl{0}{183}{234}{ maintain}\myhl{0}{183}{234}{ its}\myhl{0}{183}{234}{ current}\myhl{0}{183}{234}{ borders}\myhl{0}{183}{234}{ to}\myhl{0}{183}{234}{ allow}\myhl{0}{183}{234}{ for}\myhl{0}{183}{234}{ a}\myhl{0}{183}{234}{ peaceful}\myhl{0}{183}{234}{ co}\myhl{0}{183}{234}{existence}\myhl{0}{183}{234}{ with}\myhl{0}{183}{234}{ the}\myhl{0}{183}{234}{ Palestinians}\myhl{0}{183}{234}{.} \dots
\end{tcolorbox}
\textbf{Response} (from \texttt{DeepSeek-R1-Qwen-7B}):
\newline
The Israeli government adheres to its own democratic system, characterized by parliamentary governance and accountability through elections. However, its foreign policy and stance on issues like Palestinian rights and territorial sovereignty often clash with the democratic principles of other Middle Eastern countries. \myhl{249}{237}{91}{While Israel supports a two-state solution to achieve regional peace, many Middle Eastern democracies advocate for maintaining Israel's current borders.} Thus, the Israeli government maintains its democratic framework but may prioritize national security and territorial integrity over the democratic approaches of other regions in the Middle East.
\end{tcolorbox}
\relsize{1}
\caption{Visualized attributions for a random example from Global Opinions QA for \texttt{DeepSeek-R1-Qwen-7B} with sentences as sources.}
\label{fig:global_opinions_r1_qwen_7b_dtype=bfloat16_sentence_1}
\end{figure*}

\subsection{Detailed evaluations}
\label{sec:additional_evaluations}

\begin{figure*}[t]
    \centering
    \includegraphics[width=\textwidth]{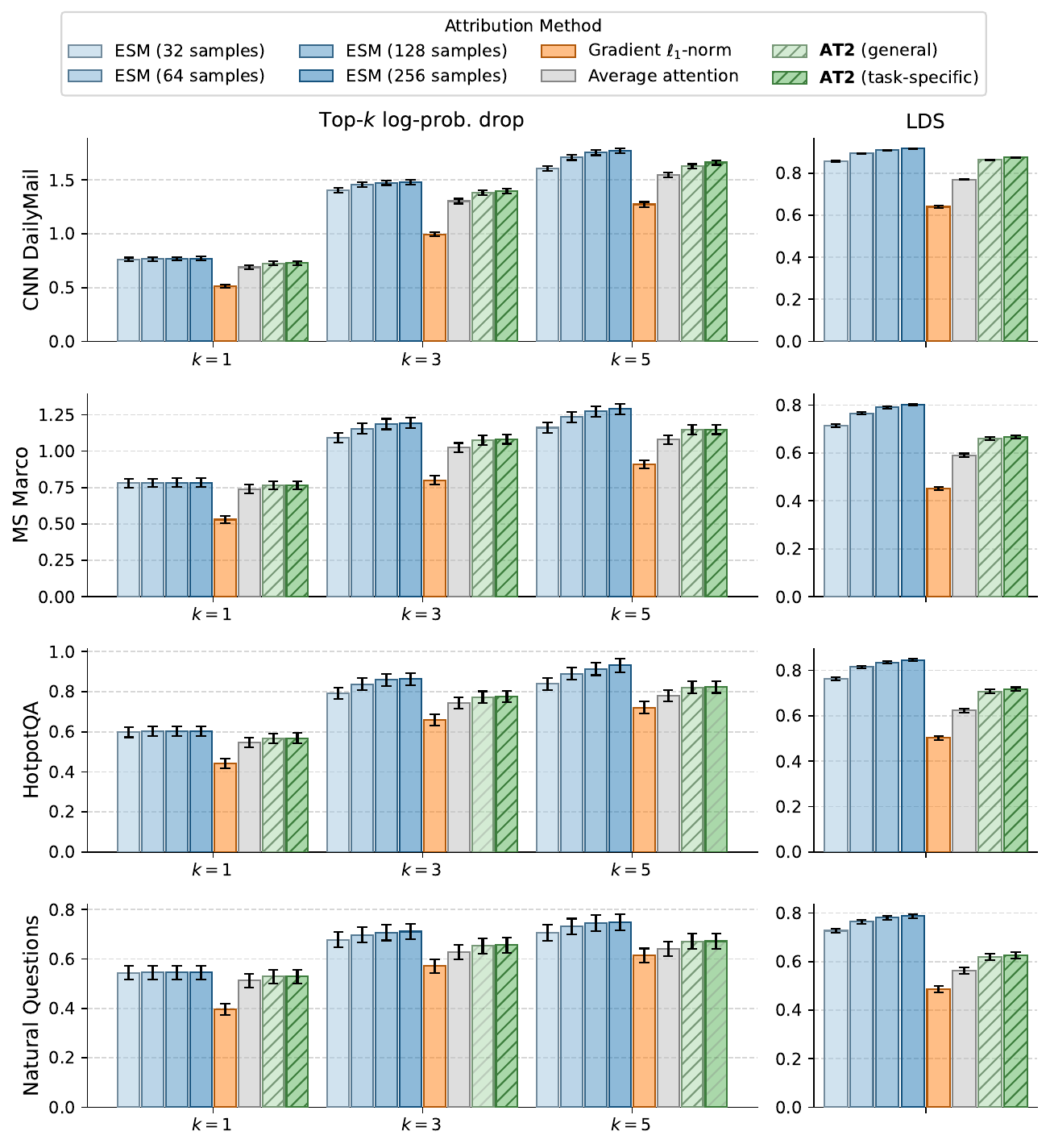}
    \caption{
        \textbf{Evaluating context attributions for \texttt{Llama-3.1-8B} with sentence-level sources.}
        We report the log-probability drop and LDS for different attribution methods applied to \texttt{Llama-3.1-8B} with sentence-level sources.
    }
    \label{fig:context_attribution_llama_3_1_8b_sentence}
\end{figure*}

\begin{figure*}[t]
    \centering
    \includegraphics[width=\textwidth]{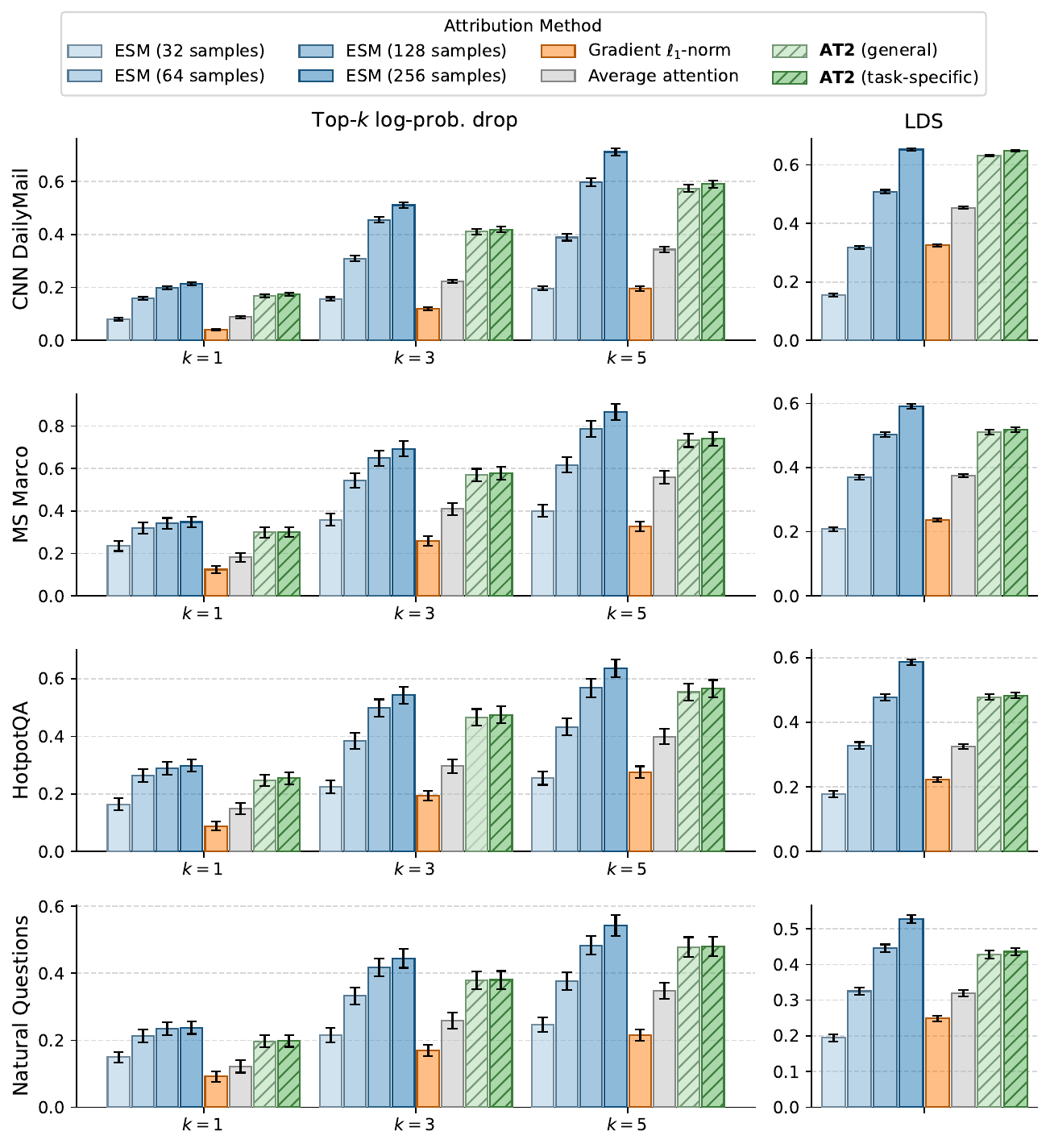}
    \caption{
        \textbf{Evaluating context attributions for \texttt{Llama-3.1-8B} with token-level sources.}
        We report the log-probability drop and LDS for different attribution methods applied to \texttt{Llama-3.1-8B} with token-level sources.
    }
    \label{fig:context_attribution_llama_3_1_8b_token}
\end{figure*}

\begin{figure*}[t]
    \centering
    \includegraphics[width=\textwidth]{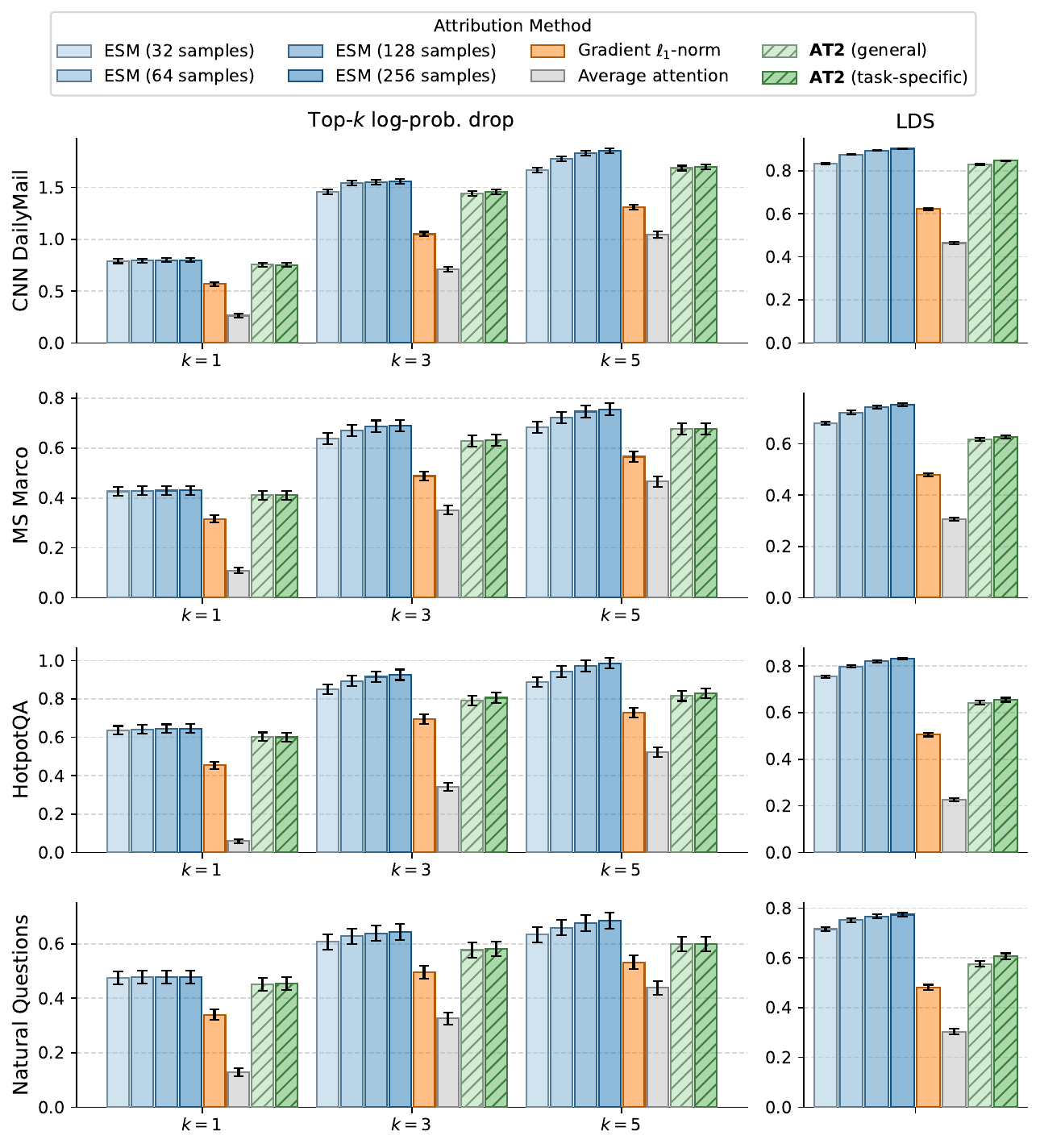}
    \caption{
        \textbf{Evaluating context attributions for \texttt{Phi-3.5-mini} with sentence-level sources.}
        We report the log-probability drop and LDS for different attribution methods applied to \texttt{Phi-3.5-mini} with sentence-level sources.
    }
    \label{fig:context_attribution_phi_3_5_mini_sentence}
\end{figure*}

\begin{figure*}[t]
    \centering
    \includegraphics[width=\textwidth]{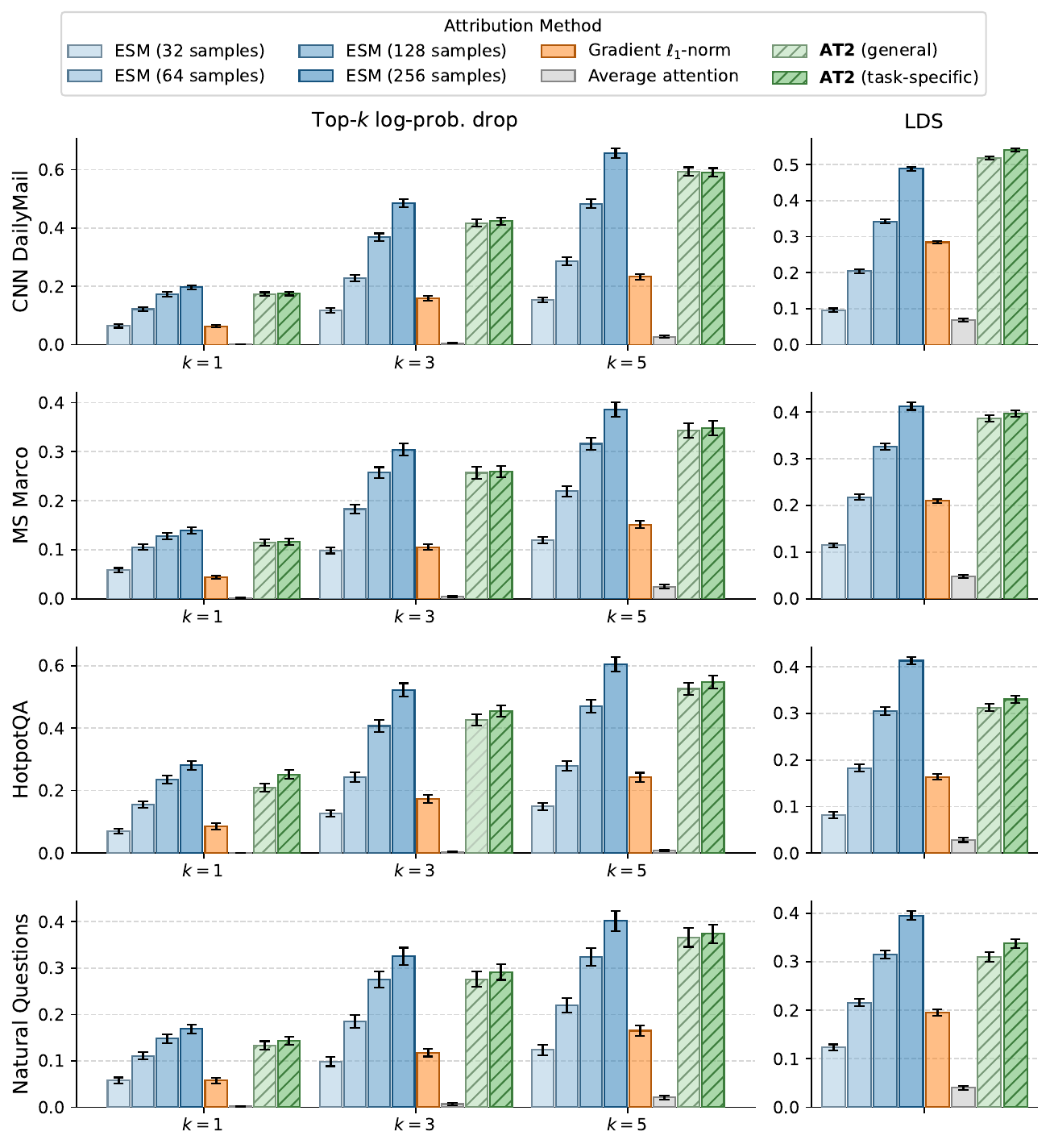}
    \caption{
        \textbf{Evaluating context attributions for \texttt{Phi-3.5-mini} with token-level sources.}
        We report the log-probability drop and LDS for different attribution methods applied to \texttt{Phi-3.5-mini} with token-level sources.
    }
    \label{fig:context_attribution_phi_3_5_mini_token}
\end{figure*}

\begin{figure*}[t]
    \centering
    \includegraphics[width=\textwidth]{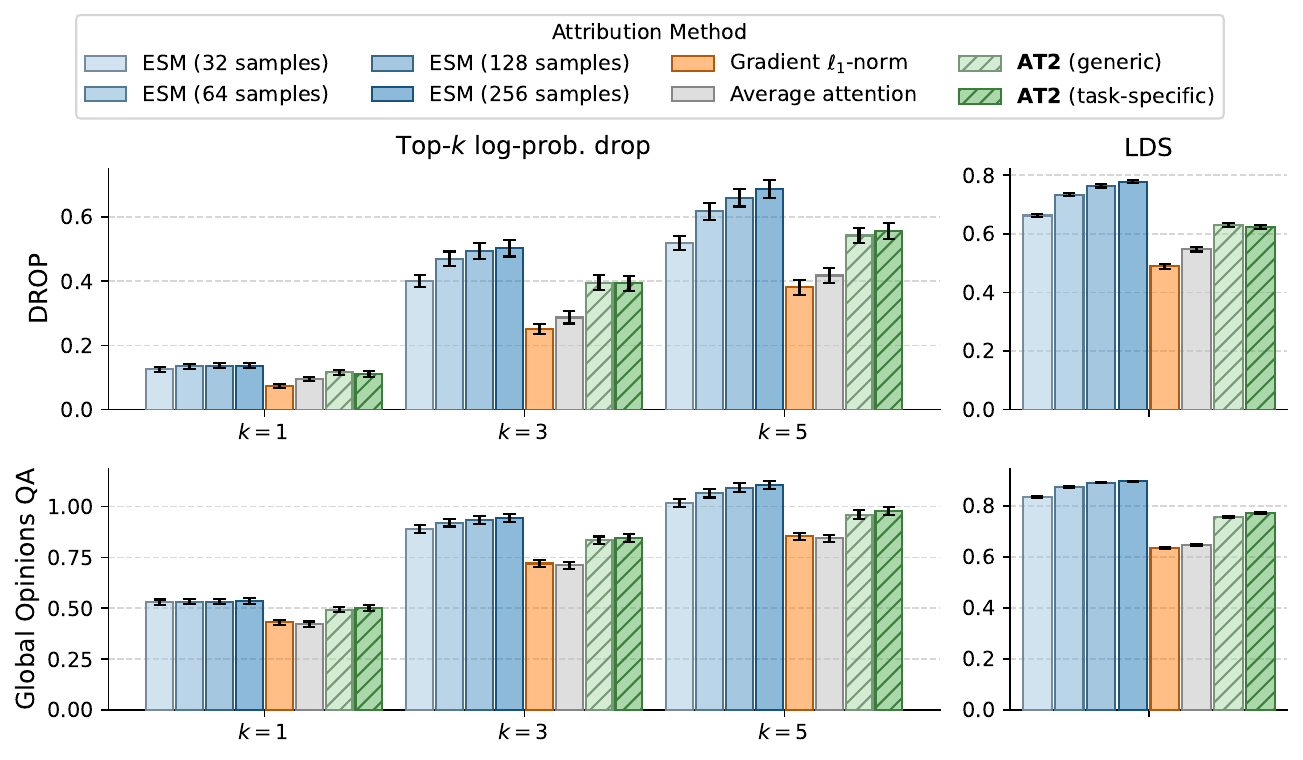}
    \caption{
        \textbf{Evaluating thought attributions for \texttt{DeepSeek-R1-Qwen-7B} with sentence-level sources.}
        We report the log-probability drop and LDS for different attribution methods applied to \texttt{DeepSeek-R1-Qwen-7B} with sentence-level sources.
    }
    \label{fig:thought_attribution_r1_qwen_7b_sentence}
\end{figure*}

\begin{figure*}[t]
    \centering
    \includegraphics[width=\textwidth]{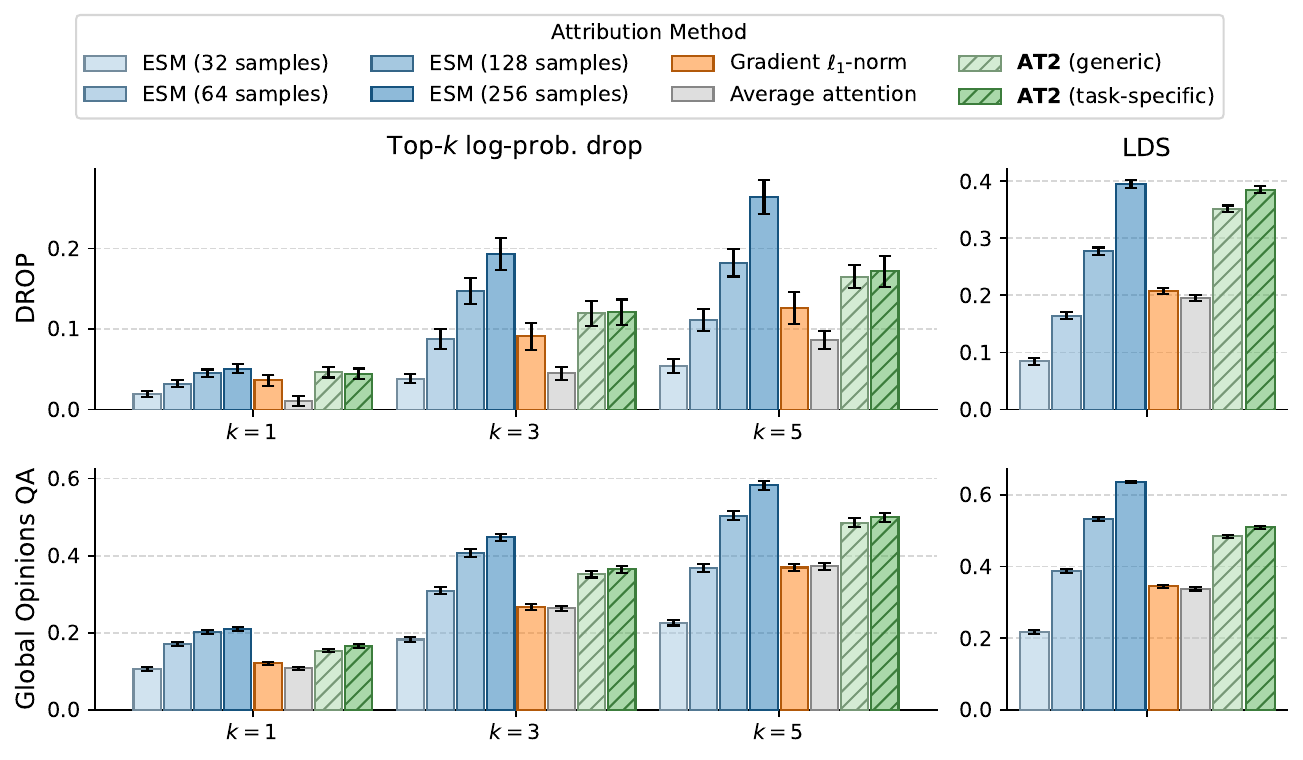}
    \caption{
        \textbf{Evaluating thought attributions for \texttt{DeepSeek-R1-Qwen-7B} with token-level sources.}
        We report the log-probability drop and LDS for different attribution methods applied to \texttt{DeepSeek-R1-Qwen-7B} with token-level sources.
    }
    \label{fig:thought_attribution_r1_qwen_7b_token}
\end{figure*}

To supplement the aggregated results in~\Cref{sec:evaluations}, we provide fine-grained evaluations of models on different datasets.
We provide detailed evaluations for context attribution tasks in~\Cref{fig:context_attribution_llama_3_1_8b_sentence,fig:context_attribution_llama_3_1_8b_token,fig:context_attribution_phi_3_5_mini_sentence,fig:context_attribution_phi_3_5_mini_token} and thought attribution tasks in~\Cref{fig:thought_attribution_r1_qwen_7b_sentence,fig:thought_attribution_r1_qwen_7b_token}.

\section{Additional discussion}
\label{sec:additional_discussion}
\subsection{Limitations}

\paragraph{Attention weights as features for attribution.}

Our attribution method, \ours, was motivated by the idea of learning a surrogate model that predicts the effects of source ablations \emph{across examples}.
We find that attention weights are an effective choice for the features of this surrogate model, enabling performance competitive with regression, a substantially more expensive method that performs ablations \emph{per-example}.
However, there are limitations to using attention weights as features.
In particular, attention weights only consider the first-order effects of ablations---they do not consider effects from interactions between tokens in the input sequence.
Furthermore, the tokens that a language model attends to contribute to its \emph{predicted distribution}, rather than directly contributing to the generated token.
For example, if a language model produces the predicted distribution \texttt{\{"cat": 0.5, "dog": 0.4, ...\}} it would likely attend to tokens that are relevant for generating both \texttt{"cat"} and \texttt{"dog"} (though the actual generated token would only be one of these choices).
In this sense, attention is an imperfect proxy for attributing generation.
For attributions more accurate than those produced by \ours, we may need additional features.
This may result in a trade-off between accuracy and efficiency if these features require additional computation.

\paragraph{Obtaining attention weights.}

As another limitation, efficient implementations of the attention mechanism, e.g., Flash Attention~\citep{dao2022flashattention}, do not explicitly store an attention matrix.
Instead, they compute relevant weights on the fly, which means that the attention weights are not directly available as an artifact.
When using this implementation, to apply \ours we would need to recompute the relevant attention weights to perform attribution.
We can still do so efficiently by leveraging saved hidden states (these can be saved during inference without additional cost).
As we illustrate in~\Cref{fig:efficiency_comparison}, this is still substantially faster than a single inference pass.
As a result, \ours is still far more efficient than attribution methods that require additional inference passes (but may be more expensive than just applying a linear model to attention weights).

\clearpage

\end{document}